\documentclass{article}

\usepackage[main, final]{neurips_2026}

\usepackage{amsmath,amssymb,amsthm,mathtools}

\usepackage[hyperfootnotes=false]{hyperref}
\usepackage{booktabs}
\usepackage{cleveref}
\usepackage{url}
\usepackage{caption}
\usepackage{subcaption}
\usepackage{rotating}
\usepackage{xcolor}
\usepackage{tikz}
\usepackage{tikz-3dplot}
\usepackage{tabularx}
\usepackage[para]{threeparttable}
\usetikzlibrary{positioning, shapes.geometric, arrows.meta, calc, matrix, fit, backgrounds}
\usepackage{adjustbox}
\usepackage{multirow}
\usepackage{graphicx}
\usepackage{float}
\usepackage{colortbl}
\usepackage{wrapfig}
\usepackage{enumitem}
\usepackage{tcolorbox}
\usepackage{placeins}

\usepackage[table]{xcolor} %
\usepackage{siunitx}       %

\definecolor{impacthigh}{HTML}{FADBD8}   %
\definecolor{impactmedium}{HTML}{FCF3CF} %

\usepackage{soul}
\usepackage{threeparttable}
\usepackage[linesnumbered,ruled,vlined]{algorithm2e}

\usepackage[textwidth=1.3in,textsize=tiny,disable=true]{todonotes}

\usepackage[utf8]{inputenc} %
\usepackage[T1]{fontenc}    %
\usepackage{hyperref}       %
\usepackage{url}            %
\usepackage{booktabs}       %
\usepackage{amsfonts}       %
\usepackage{nicefrac}       %
\usepackage{microtype}      %
\usepackage{xcolor}         %
\usepackage{booktabs}
\usepackage{fontawesome5}

\usepackage{pifont}
\newcommand{\cmark}{\ding{51}}      %
\newcommand{\xmark}{\ding{55}}      %

\usepackage{amsmath,amsfonts,bm}

\usepackage[symbol]{footmisc} %

\makeatletter
\renewcommand\@fnsymbol[1]{%
  \ifcase#1\or\dagger\or\ddagger\or\mathsection\or\mathparagraph\or\|\or\#\fi}
\makeatother

\definecolor{SolarizedBase3}{HTML}{FDF6E3} %
\definecolor{SolarizedBase1}{HTML}{93A1A1} %
\definecolor{SolarizedGreen}{HTML}{859900} %
\definecolor{SolarizedCyan}{HTML}{2AA198} %
\definecolor{SolarizedViolet}{HTML}{6C71C4} %
\definecolor{SolarizedBase00}{HTML}{657B83} %
\definecolor{SolarizedOrange}{HTML}{CB4B16} %
\definecolor{SolarizedRed}{HTML}{DC322F}   %
\definecolor{LineNoGray}{rgb}{0.65,0.65,0.65} %
\usepackage{listings}

\lstdefinestyle{mystyle}{
    backgroundcolor=\color{white}, %
    commentstyle=\color{SolarizedBase1}\itshape,
    keywordstyle=\color{SolarizedGreen}\bfseries,
    numberstyle=\tiny\color{LineNoGray}, %
    stringstyle=\color{SolarizedCyan},
    basicstyle=\scriptsize\ttfamily, %
    identifierstyle=\color{SolarizedBase00}, %
    breakatwhitespace=false,
    breaklines=true,
    captionpos=b,
    keepspaces=true,
    numbers=left,
    numbersep=4pt, %
    showspaces=false,
    showstringspaces=false,
    showtabs=false,
    tabsize=3, %
    frame=tb, %
    framerule=0.4pt, %
    rulecolor=\color{black!20}, %
    aboveskip=1em, %
    belowskip=1em, %
    emph={ %
        GatedDeltaProduct, FusedRMSNormSwishGate, RMSNorm, ShortConvolution,
        Cache, Unpack, nn, torch, F, einops, rearrange, math,
        chunk_delta_rule, chunk_gated_delta_rule, __init__, forward,
        ModuleList, Parameter, Module, Linear, SiLU, Identity,
        Optional, Tuple, Dict, TYPE_CHECKING,
        range, isinstance, super, getattr, setattr, len, all, int, float, bool, str
    },
    emphstyle=\color{SolarizedOrange}\bfseries, %
    emph={[2]self, True, False, None}, %
    emphstyle={[2]\color{SolarizedRed}}, %
    morekeywords={ %
        output_attentions, use_cache, past_key_values, attention_mask, hidden_states,
        layer_idx, conv_state, recurrent_state, offsets, cu_seqlens
    },
    keywordstyle={[2]\color{SolarizedViolet}}, %
    keywordsprefix=@, %
    keywordstyle={[3]\color{SolarizedViolet}}, %
}

\def\eqref#1{equation~\ref{#1}}

\def\1{\bm{1}}

\def\vk{{\bm{k}}}

\def\vv{{\bm{v}}}

\def\vx{{\bm{x}}}
\def\vy{{\bm{y}}}

\def\mA{{\bm{A}}}
\def\mB{{\bm{B}}}

\def\mH{{\bm{H}}}
\def\mI{{\bm{I}}}

\def\mU{{\bm{U}}}
\def\mV{{\bm{V}}}
\def\mW{{\bm{W}}}

\DeclareMathAlphabet{\mathsfit}{\encodingdefault}{\sfdefault}{m}{sl}
\SetMathAlphabet{\mathsfit}{bold}{\encodingdefault}{\sfdefault}{bx}{n}

\newcommand{\methodname}{TempoPFN}

\title{\methodname{}: Synthetic Pre-training of Linear RNNs\\ for Zero-shot Time Series Forecasting}

\author{Vladyslav Moroshan$^{*\diamondsuit}$, Julien Siems$^{*\diamondsuit}$, Arber Zela$^{\heartsuit}$, Timur Carstensen$^{\diamondsuit \clubsuit}$, \textbf{Frank Hutter$^{\text{\faTable} \clubsuit \diamondsuit}$} \\
University of Freiburg$^{\diamondsuit}$,~~ELLIS Institute Tübingen$^{\clubsuit}$,~~Prior Labs$^{\text{\faTable}}$,~~EPFL$^{\heartsuit}$ \\
{\small $^*$Equal contribution}
\quad\quad
{\small \texttt{vlad.moroshan@gmail.com}}
\quad
{\small \texttt{juliensiems@gmail.com}}
}

\begin{document}

\maketitle

\begin{abstract} Foundation models for zero-shot time series forecasting face challenges in efficient long-horizon prediction and reproducibility, with existing synthetic-only approaches underperforming on challenging benchmarks. This paper presents TempoPFN, a univariate time series foundation model based on linear Recurrent Neural Networks (RNNs) pre-trained exclusively on synthetic data. The model uses a GatedDeltaProduct architecture with state-weaving for fully parallelizable training across sequence lengths, eliminating the need for windowing or summarization techniques while maintaining robust temporal state-tracking. Our comprehensive synthetic data pipeline unifies diverse generators, including stochastic differential equations, Gaussian processes, and audio synthesis, with novel augmentations. In zero-shot evaluations on the Gift-Eval, fev-bench and Chronos-ZS benchmarks, TempoPFN achieves top-tier competitive performance, surpassing the majority of models trained on real-world data, while leveraging fully parallelizable training and inference. We open-source our complete data generation pipeline and training code, providing a reproducible foundation for future research.
\end{abstract}
\begin{table}[b]
\vspace{-3mm}
\small
  \caption{Contributions of TempoPFN: the first fully open-source time series forecasting foundation model with competitive performance; with fully synthetic pretraining and fast training and inference.}
  \label{table:contributions}
  \vspace{1mm}
  \centering
  \begin{tabular}{lccccc}
    \toprule
    Criterion & TiRex & TabPFN-TS & Mamba4Cast & Chronos & TempoPFN \\
    \midrule
    Fully open-source data pipeline   & \textcolor{red}{\xmark} & \textcolor{red}{\xmark}   & \textcolor{green}{\cmark} & \textcolor{red}{\xmark}   & \textcolor{green}{\cmark} \\
    Open-source training code   & \textcolor{red}{\xmark} & \textcolor{red}{\xmark}   & \textcolor{green}{\cmark} & \textcolor{green}{\cmark}   & \textcolor{green}{\cmark} \\
    Competitive with SOTA performance     & \textcolor{green}{\cmark} & \textcolor{green}{\cmark}   & \textcolor{red}{\xmark}  &  \textcolor{green}{\cmark}   & \textcolor{green}{\cmark} \\
    Fast training and inference     & \textcolor{green}{\cmark} & \textcolor{red}{\xmark}   & (\cmark) & (\cmark)   & \textcolor{green}{\cmark} \\
    Purely synthetic pretraining       & \textcolor{red}{\xmark} & (\cmark) & \textcolor{green}{\cmark} & \textcolor{red}{\xmark}     & \textcolor{green}{\cmark} \\
    \bottomrule
  \end{tabular}
\end{table}

\section{Introduction}\label{sec:intro}

Recent advances in large language models have inspired foundation models for time series forecasting that enable zero-shot predictions across diverse datasets without fine-tuning~\citep{ansari2024chronos,das2024decoder,woo2024unified,auer2025tirex}. By treating historical observations as input context, these models democratize forecasting for non-experts and excel in data-scarce domains. 

However, current approaches face critical limitations. Transformer-based models struggle with long-horizon forecasting due to quadratic complexity and error accumulation~\citep{zeng2023transformers}. While non-linear RNNs like those in TiRex~\citep{auer2025tirex} maintain temporal state, they require sequential processing that limits scalability. Although some recent models attempt synthetic-only pre-training including ForecastPFN~\citep{dooley2023forecastpfn}, CauKer~\citep{cauker}, and Mamba4Cast~\citep{mamba4cast} none reported state-of-the-art performance on the Gift-Eval benchmark. TabPFN-TS~\citep{hoo2024the}, which adapts a tabular foundation model to time series, achieves strong Gift-Eval performance but does not release its synthetic pre-training data, limiting reproducibility and extensibility. 

\begin{figure}[t]
    \vspace{-12mm}  
    \centering
    \adjustbox{width=0.9\linewidth}{        \input{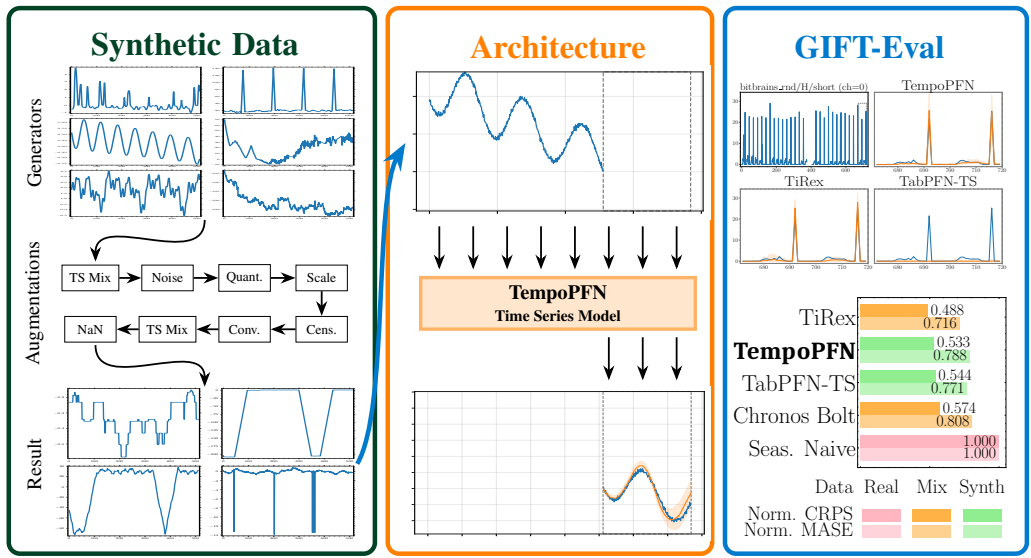}
    }
    \vspace{-1mm}
    \caption{(Left) Synthetic Data Generation pipeline containing a mix of novel and existing time-series generators are augmented with a diverse set of augmentations to produce the time-series used for training.
    (Middle) The \methodname{} architecture produces coherent quantile predictions for all future time-stamps in parallel. (Right) \methodname{} obtains competitive performance on Gift-Eval despite being trained only on synthetic time-series.}
    \label{fig:ts_pipeline}
\end{figure}

We introduce \textbf{\methodname{}} (see \Cref{table:contributions} and \Cref{fig:ts_pipeline}), a time series forecasting foundation model using \emph{linear RNNs with GatedDeltaProduct recurrence}~\citep{siems2025deltaproduct} for parallelizable training and inference across the sequence length. We adopt the Prior-Data Fitted Network (PFN) framework~\citep{muller2022transformers}, treating zero-shot forecasting as Bayesian inference approximated via in-context learning on a diverse synthetic prior (see Appendix~\ref{app:pfn_background} for a detailed background). Unlike TiRex~\citep{auer2025tirex} which argued that non-linear RNNs like sLSTM are necessary for time-series forecasting due to their state-tracking capabilities, we find that linear RNNs based on the GatedDeltaProduct recurrence are sufficient, in line with recent research demonstrating how linear RNNs can perform state-tracking~\citep{grazzi-iclr25a}. As detailed in Appendix~\ref{app:gated_deltaproduct}, DeltaProduct applies orthogonal rotations via multiple online gradient steps, enabling superior state-tracking compared to diagonal SSMs. Our synthetic data pipeline unifies diverse generators with novel augmentations, ensuring exclusive synthetic pre-training to prevent benchmark leakage. Unlike TabPFN-TS, we open-source our complete data generation pipeline and training code as a basis for future research (available at \url{https://github.com/automl/TempoPFN}).

In summary, our contributions are: 
\begin{itemize}[leftmargin=2.5mm]
    \item The \textbf{\methodname{}} architecture is, to our knowledge, the first univariate time series foundation model based on \emph{linear RNNs with GatedDeltaProduct recurrence}. Our architecture and input representation allows the prediction of all future time-stamps in parallel, producing coherent quantile forecasts, without patching or windowing heuristics. We further propose a state-weaving mechanism for linear RNNs that facilitates bidirectional information flow across horizons without overhead.
    \item We design a \textbf{synthetic data pipeline} combining existing and novel synthetic generators with a cascade of augmentations, ensuring diverse temporal structures without relying on real-world data, thereby eliminating benchmark leakage and mitigating privacy concerns associated with training on real-world data. We release a fully open-source synthetic data generation pipeline for time series forecasting that achieves competitive performance on Gift-Eval, and show via a backbone swap that this prior is the primary driver of accuracy. 
    \item Compared to nonlinear RNNs and transformer time series foundation models, \methodname{} achieves \textbf{top-tier competitive zero-shot performance on Gift-Eval, \texttt{fev-bench} and Chronos-ZS}, surpassing the vast majority of models trained on real-world data. This result is achieved without any non-linearity in the recurrence, demonstrating that linear RNNs as a scalable and powerful alternative to non-linear RNNs and transformers for time series foundation models. 
\end{itemize}

\section{Background and Related Work}
\vspace{-2mm}
\textbf{Time Series Forecasting.} Time series forecasting aims to predict future values $y_{T+1:T+H}$ from historical observations $y_{1:T}$. Classical methods such as ARIMA \citep{box1968some} and exponential smoothing \citep{hyndman2008forecasting} produce point estimates, while probabilistic forecasting models the predictive distribution $p(y_{T+1:T+h}\mid y_{1:T})$. Deep learning has expanded this toolkit with transformers \citep{vaswani2017attention} and modern recurrent architectures \citep{beck2024xlstm, gu2023mamba}. A key recent development is \emph{zero-shot forecasting}, in which models pre-trained on diverse corpora can predict unseen time series without fine-tuning, mirroring cross-domain generalization in NLP and vision. Most successful approaches employ transformers: Chronos \citep{ansari2024chronos}, TimesFM \citep{das2024decoder}, and MOIRAI \citep{woo2024unified} use patching, frequency-specific projections, and masked modeling to handle heterogeneous data. MOIRAI-MOE \citep{liu2024moirai} adds a sparse mixture of experts for token-level specialization and robustness. Among true zero-shot models, MOIRAI currently achieves state-of-the-art performance on Gift-Eval while avoiding benchmark overlap.

\textbf{Prior-data Fitted Networks (PFNs) and Synthetic Data.} PFNs~\citep{muller2022transformers} represent a paradigm shift from solving a single task to learning a \emph{universal inference algorithm} by training a neural network to approximate the posterior predictive distribution induced by a prior over datasets. By minimizing the expected negative log-likelihood across datasets sampled from this prior (see Appendix \ref{app:pfn_background} for more details), PFNs enable \emph{in-context learning} as fast approximate Bayesian inference: instead of storing solutions to tasks, the model’s weights encode an inference algorithm that conditions on a small context (e.g., a time-series history) and predicts future values in a single forward pass~\citep{muller2022transformers, hollmann2023tabpfn}. As PFN performance is determined by the expressiveness of the prior, models are typically trained on large synthetic corpora. For instance, TabPFN~\citep{hollmann2023tabpfn} uses structural causal models for tabular data, ForecastPFN~\citep{dooley2023forecastpfn} employs trend–seasonality priors for time series, TimePFN~\citep{taga2025timepfn} extends this to multivariate settings with Gaussian process kernels, and TabPFN-TS~\citep{hoo2024the} adapts PFNs to time series via TabPFNv2~\citep{hollmann2025accurate}. In this paper, the prior is our synthetic temporal-dynamics pipeline (see Section~\ref{subsec:synthetic_data}). Training on this synthetic prior enables \methodname{} to perform zero-shot probabilistic forecasting on unseen time series.

\textbf{Linear RNNs and State-Space Models.}
Recent work has revisited recurrent architectures for long-horizon forecasting. TiRex~\citep{auer2025tirex} uses xLSTM~\citep{beck2024xlstm} pre-trained on synthetic Gaussian processes, Chronos datasets, and selected Gift-Eval subsets, combined with augmentations such as amplitude modulation, censoring, and spike injection. In contrast, \methodname{} leverages linear RNNs with GatedDeltaProduct~\citep{siems2025deltaproduct} mechanisms and negative eigenvalues~\citep{grazzi-iclr25a}, enabling fully parallelizable training without patching or summarization, while relying solely on synthetic pretraining data to avoid any leakage. Linear RNNs have regained interest due to their efficient parallelization. While non-linear RNNs are not easily parallelizable~\citep{NEURIPS2024_0b2b199f}, linear RNNs admit chunk-wise parallelization~\citep{yang-icml24a} or associative scans~\citep{gu2023mamba, martin2018parallelizing}. Formally, they map input sequences $\vx_{1:t}\!\in\!\mathbb{R}^l$ to outputs $\hat{\vy}_{1:t}\!\in\!\mathbb{R}^p$ via
$\mH_i = \mA(\vx_i)\mH_{i-1} + \mB(\vx_i), \qquad  
\hat{\vy}_i = \mathrm{dec}(\mH_i,\vx_i), \quad i=1,\ldots,t$
where $\mA$ parameterizes state transitions, $\mB$ state inputs, and $\mathrm{dec}$ the output. Variants such as Mamba~\citep{dao-icml24a}, GLA~\citep{yang-icml24a}, and mLSTM~\citep{beck2024xlstm} use diagonal transitions, while more expressive models relax this constraint, including DeltaNet~\citep{schlag-icml21a, irie2023practical, yang-neurips24a}, TTT-Linear~\citep{sun-arxiv24a}, RWKV-7~\citep{peng2025rwkv7gooseexpressivedynamic}, B'MOJO~\citep{zancato-neurips24a}, and Titans~\citep{behrouz2024titans}.

\section{\methodname{}}\label{sec:bullet_time}
\vspace{-1mm}
\subsection{Architecture}\label{sec:architecture}
\begin{wrapfigure}[26]{r}{0.275\textwidth}
    \centering
    \vspace{-4.5mm}
    \adjustbox{width=1.0\linewidth, trim={2 0 0 0}, clip=true}{
    \begin{tikzpicture}[
    block/.style={rectangle, draw=black, thick,
                    minimum width=3.5cm, minimum height=0.6cm, align=center},
    arrow/.style={-Stealth, thick},
    arrow_inner/.style={-Stealth, black!50},
    state_arrow/.style={-Stealth, thick, blue},
    plus_circle/.style={circle, draw, inner sep=1pt, font=\large, blue},
    prediction_box/.style={rectangle, draw=red!80, fill=red!20, thick,
                           minimum width=0.7cm, minimum height=0.6cm, anchor=center},
    node distance=0.8cm and 1cm
]

\matrix (timeline) [
    matrix of nodes,
    nodes={minimum height=0.6cm, minimum width=0.8cm, anchor=center},
    column sep=0.1cm,
    row sep=0.05cm,
    nodes in empty cells,
] at (0,0.5cm)
{
    |[rectangle, draw=teal!80, fill=teal!20, thick]| $y_0$ &
    |[rectangle, draw=teal!80, fill=teal!20, thick]| $y_1$ &
    |[rectangle, draw=teal!80, fill=teal!20, thick]| {\tiny NaN} &
    |[rectangle, draw=teal!80, fill=teal!20, thick]| $y_3$ &
    |[rectangle, draw=red!80, fill=red!20, thick]| ? &
    |[rectangle, draw=red!80, fill=red!20, thick]| ? \\
    |[rectangle, draw=teal!80, fill=white!20, thick]| $t_0$  & |[rectangle, draw=teal!80, fill=white!20, thick]| $t_1$ & |[rectangle, draw=teal!80, fill=white!20, thick]| $t_2$ & |[rectangle, draw=teal!80, fill=white!20, thick]| $t_3$ & |[rectangle, draw=red!80, fill=white!20, thick]| $t_4$ & |[rectangle, draw=red!80, fill=white!20, thick]| $t_5$ \\
};

\node[below=0.5cm of $(timeline-2-1)!0.5!(timeline-2-4)$] {\textbf{HISTORY}};
\node[below=0.5cm of $(timeline-2-5)!0.5!(timeline-2-6)$] {\textbf{FUTURE}};

\node[block] (delta1) [above=0.7cm of timeline] {Gated DeltaProduct};
\node[block] (mlp1)   [above=0.4cm of delta1] {MLP}; %
\node[block] (delta2) [above=of mlp1] {Gated DeltaProduct};
\node[block] (mlp2)   [above=0.4cm of delta2] {MLP}; %
\node[block] (delta3) [above=of mlp2] {Gated DeltaProduct};
\node[block] (mlp3)   [above=0.4cm of delta3] {MLP}; %

\node[block, minimum width=1.3cm] (output) at ($(mlp3.north -| timeline-2-4)!0.375!(mlp3.north -| timeline-2-6)$) [above=0.5cm] {Linear};

\colorlet{block1color}{cyan!15}
\colorlet{block2color}{green!15}
\colorlet{block3color}{orange!15}

\begin{scope}[on background layer]
    \node [fit=(delta1) (mlp1), fill=block1color, draw=cyan!50!black, thick, rounded corners=5pt, inner sep=0.25cm] {};
    \node [fit=(delta2) (mlp2), fill=block2color, draw=green!50!black, thick, rounded corners=5pt, inner sep=0.25cm] {};
    \node [fit=(delta3) (mlp3), fill=block3color, draw=orange!50!black, thick, rounded corners=5pt, inner sep=0.25cm] {};
\end{scope}

\foreach \i in {1,...,6} {
    \coordinate (dest) at ($(delta1.south west)!{(\i-0.5)/6}!(delta1.south east)$);
    \draw[arrow] (timeline-1-\i.north) -- (dest);
}

\foreach \source/\dest in {delta1/mlp1, mlp1/delta2, delta2/mlp2, mlp2/delta3, delta3/mlp3} {
    \foreach \i in {1,...,6} {
        \draw[arrow_inner] ($(\source.north west)!{(\i-0.5)/6}!(\source.north east)$) -- ($(\dest.south west)!{(\i-0.5)/6}!(\dest.south east)$);
    }
}

\foreach \i in {5,6} {
    \coordinate (startarrow) at ($(mlp3.north west)!{(\i-0.5)/6}!(mlp3.north east)$);
    \draw[arrow] (startarrow) -- (startarrow |- output.south);
}

\node[plus_circle] (add2) [left=0.5cm of delta2] {$+$};
\node[plus_circle] (add3) [left=0.5cm of delta3] {$+$};

\node[align=center, left=0.33cm of delta1, blue] (h0_1) {$\mH_0^0$};
\node[align=center, above=0.35cm of add2, blue] (h0_2) {$\mH_0^1$};
\node[align=center, above=0.35cm of add3, blue] (h0_3) {$\mH_0^2$};

\draw[state_arrow] (h0_1.east) -- (delta1.west);
\draw[state_arrow] (h0_2.south) -- (add2.north);
\draw[state_arrow] (h0_3.south) -- (add3.north);

\draw[state_arrow] (add2.east) -- (delta2.west);
\node[blue, right=0.35cm of delta1] (h_out_1) {$\mH_5^0$};
\draw[state_arrow] (delta1.east) -- (h_out_1.west);
\draw[state_arrow] (h_out_1.north) to[out=100, in=-90, looseness=0.8] (add2.south);

\draw[state_arrow] (add3.east) -- (delta3.west);
\node[blue, right=0.35cm of delta2] (h_out_2) {$\mH_5^1$};
\draw[state_arrow] (delta2.east) -- (h_out_2.west);
\draw[state_arrow] (h_out_2.north) to[out=100, in=-90, looseness=0.8] (add3.south);

\coordinate (y4_start) at ($(output.north west)!0.225!(output.north east)$);
\coordinate (y5_start) at ($(output.north west)!0.85!(output.north east)$);

\node[prediction_box] (y4) [above=0.5cm of y4_start] {$\hat{y}_4$};
\node[prediction_box] (y5) [above=0.5cm of y5_start] {$\hat{y}_5$};

\draw[arrow] (y4_start) -- (y4);
\draw[arrow] (y5_start) -- (y5);

\node[left=0.45cm of $(y4)$] {\textbf{PREDICTION}};

\end{tikzpicture}
    }
\vspace{-13pt} 
\caption{The \methodname{} architecture (3 blocks) uses stacked GatedDeltaProduct blocks, learnable initial states $\mH_0^i$ and state-weaving.}\label{fig:architecture_diagram}
\end{wrapfigure}
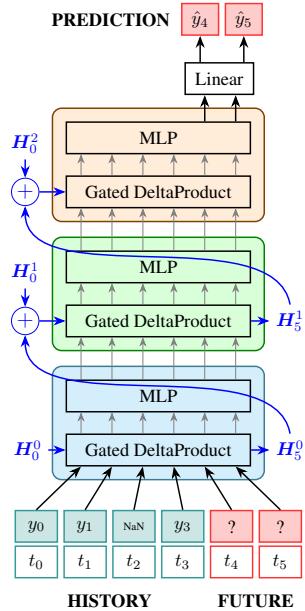

\vspace{-1mm}
The \methodname{} architecture is designed to forecast univariate time series across a full prediction horizon in a single forward pass, as illustrated in~\Cref{fig:architecture_diagram}. It consists of four main stages: input representation, backbone, non-causality through state weaving, and prediction.

\textbf{Input representation.}
\methodname{} uses an input representation in which history (time steps + values) and future (time steps) are concatenated into one token sequence enabling communication between future time steps for coherent predictions. In contrast to TiRex, which presummarizes time steps into windows of size 32, \methodname{} directly operates on the individual time steps. Each time step $t_i$ is encoded using \texttt{GluonTS}~\citep{gluonts} time features (e.g., seasonality indicators, day-of-week, or index-based encodings) that are linearly projected into the embedding dimension of the model. For historical steps, observed values $y_i$ are projected via a linear layer, while missing values are handled by a learnable \texttt{NaN} embedding. The historical embedding is obtained by additively combining the value and the time-feature embeddings. For future time steps, only the time-feature embedding is used.

\textbf{Backbone.}
The core of \methodname{} is a stack of 10 encoder layers, each based on the \emph{Gated DeltaProduct block}~\citep{siems2025deltaproduct} from the \texttt{flash-linear-attention} library~\citep{fla}, originally derived from the LLaMA architecture~\citep{touvron2023llama}. Each block consists of three components: 
(1) \emph{token mixing} through a Gated DeltaProduct recurrence with short one-dimensional convolutions (kernel size 16–32), 
(2) \emph{pre-normalization} applied before the recurrent unit to stabilize training, and 
(3) a gated MLP for channel-wise feature transformation. This design combines the parallelization advantages of linear recurrences with the expressivity of lightweight convolutional and feedforward operations. 
DeltaProduct, which generalizes DeltaNet’s non-diagonal transitions by expressing $\mA(\vx_i)$ as a product of $n_h$ generalized Householder transformations, enabling a rank-$n_h$ transformation of the matrix-valued hidden state $\mA(\vx_i) = \prod_{j=1}^{n_h} \bigl(\mI - \beta_{i,j}\,\vk_{i,j}\vk_{i,j}^\top\bigr)$. For each token $\vx_i$, the model generates $n_h$ normalized keys $\vk_{i,j} = \psi(\mW_j \vx_i)/\|\psi(\mW_j \vx_i)\|_2$, values $\vv_{i,j} = \mV_j \vx_i$, and coefficients $\beta_{i,j} = \phi(\mU_j \vx_i)$ using learnable matrices $\mW_j,\mV_j,\mU_j$, SiLU activation $\psi$~\citep{hendrycks2016gaussian}, and a sigmoid-based gating function $\phi$. \citet{siems2025deltaproduct} found that increasing $n_h$ leads to significantly improved length-extrapolation, language modeling, and state-tracking on permutation tasks, all capabilities which are equally desirable for time-series forecasting.

\textbf{Non-causality via state weaving.} 
Whereas DeltaProduct was originally developed for autoregressive language modeling, forecasting across a full horizon does not require causal masking. To exploit this property, we introduce \emph{state weaving}. Specifically, the final hidden state of each layer $\mH_t^i$ is added to the learnable initial state of the next layer $\mH_0^{i+1}$. This mechanism enables bidirectional information flow across the entire sequence length without additional parameters or computational overhead through explicit bidirectionality~\citep{NEURIPS2024_c7f795dc,afzal2025linear}, hence allowing future time-steps to attend to the entire history and future context, preventing the information bottleneck typical of causal RNNs during the prediction phase.

\textbf{Prediction.} 
At the output stage, embeddings corresponding to the forecast horizon are extracted from the final encoder block. These embeddings are passed through a linear projection head that outputs multiple \textit{quantiles} of the predictive distribution, enabling probabilistic forecasting (see also Appendix~\ref{app:training-details} for details on the quantile loss used for training). Overall, this design allows to directly predict all future values given a history using a single forward pass.

\subsection{Synthetic Data Generation}
\label{subsec:synthetic_data}
\vspace{-2mm}
To train our time series foundation model, we generated a large and diverse dataset using 10 different synthetic generators. This approach combines established data generation techniques with novel methods to capture a wide spectrum of temporal patterns and behaviors. For a more comprehensive description of each generator, refer to Appendix~\ref{app:synthetic_details}.

\textbf{Existing Generators.}
We adapted several established generators from prior work to ensure comprehensive coverage of common temporal patterns. The \textbf{ForecastPFN} generator \citep{dooley2023forecastpfn} composes multiplicative trend and seasonality components, combining linear and exponential growth terms with sinusoidal harmonics. The generator includes Weibull-distributed noise and augmentations such as time warping, magnitude scaling, and spike injection, with filtering mechanisms to avoid extreme values. \textbf{KernelSynth}, following \citet{ansari2024chronos}, samples univariate time series from Gaussian process priors with composite kernels. Base kernels include periodic (ExpSineSquared), stationary (RBF, RationalQuadratic), and noise (WhiteKernel) components, combined through addition or multiplication to yield smooth yet varied trajectories. We extended this approach with a broader \textbf{Gaussian Process} generator, as in \citet{mamba4cast} that randomly combines kernels with greater functional diversity, producing wider ranges of stationary and nonstationary patterns. The \textbf{CauKer} generator \citep{cauker} introduces causal dependencies by sampling from structural causal models (SCMs). Each node represents a Gaussian process with composite kernels and stochastic mean functions, while edges in a random DAG apply nonlinear transformations. We generate 21-channel multivariate series, treating each channel as an independent univariate signal to capture diverse, interdependent dynamics.

\textbf{Novel Generators.}
We developed several new generators to fill gaps in existing approaches and capture specific temporal behaviors. \textbf{Sawtooth} creates ramp-like patterns with upward or downward slopes, enhanced with small linear trends and low-amplitude seasonal components to avoid overly idealized signals. \textbf{Step Function} produces piecewise constant series with configurable changepoints, step sizes, and drift, using Gaussian smoothing at boundaries along with added noise, seasonality, and anomalies. For anomaly-rich data, we created two specialized generators. \textbf{Anomaly} produces baseline signals with periodic or clustered spikes, varying in magnitude regimes (constant, trending, cyclical, or correlated random) and timing patterns. \textbf{Spikes} emphasizes event-driven behavior by placing sharp spikes on flat baselines, with configurable shapes (V, inverted V, or plateau variants) arranged in bursty or evenly spread patterns. The \textbf{Sine Wave} generator provides clean oscillatory patterns with configurable period, amplitude, phase, and noise, offering fundamental periodic signals for learning basic oscillatory structures. To capture highly complex, real-world dynamics, we introduce \textbf{Audio-Inspired Generators} that use procedural audio synthesis techniques implemented with Pyo~\citep{pyo}. These generators model phenomena such as \textbf{Stochastic Rhythms} for event data, \textbf{Financial Volatility} with market shocks and clustering, \textbf{Network Topology} with traffic bursts and congestion, and \textbf{Multi-Scale Fractals} for self-similar patterns. Our most sophisticated contribution is the \textbf{stochastic differential equation} (\textbf{SDE}) generator, a flexible synthetic data generator based on a regime-switching, time-inhomogeneous Ornstein–Uhlenbeck (OU) process. The OU process follows the SDE $dy_t = \theta(t,r_t)\,\bigl(\mu(t,r_t) - y_t \bigr)\,dt + \sigma(t,r_t)\, dW_t$ 
where $\theta(t,r_t)$ is the mean reversion speed, $\mu(t,r_t)$ the time-varying mean, and $\sigma(t,r_t)$ the volatility. Each parameter depends on both time $t$ and a latent regime $r_t \in \{0,1\}$ that evolves as a Markov chain. This framework enables parameters to shift abruptly across regimes while drifting smoothly over time through polynomial, sinusoidal, logistic, or piecewise-linear trends. Seasonal patterns are injected additively into both mean and volatility components, with amplitudes subject to gradual growth or decay. For enhanced realism, we optionally replace standard Brownian motion with fractional Brownian motion, introducing long-memory dynamics through the Hurst exponent $H \in (0,1)$. Each simulated series undergoes global rescaling and shifting before additive Gaussian measurement noise is applied. This construction produces highly diverse temporal structures, capturing regime shifts, non-stationarity, periodicity, and measurement noise within a principled stochastic framework.

\subsection{Data Augmentations}
\begin{figure}[t]
\centering
\adjustbox{width=0.9\linewidth}{
\begin{tikzpicture}[
    node distance=0.3cm and 0.8cm,
    font=\scriptsize,
    process/.style={draw, thick, rectangle, rounded corners, minimum height=1.5em, text centered, fill=cyan!15, text width=3cm},
    category_node/.style={draw, thick, rectangle, rounded corners, inner sep=0.3em, fill=cyan!15, text centered},
    image/.style={inner sep=0pt}, %
    op/.style={text width=3.2cm, align=center},
    arrow/.style={-Stealth, thick, color=blue!60},
    prob_label/.style={midway, above, font=\tiny, color=black, fill=white, inner sep=1pt, yshift=-1pt}
]
\node[process] (source) {Source Sampling \& Scaling};
\node[process, below=of source] (mixup1) {TS-Mixup (Early)};
\node[category_node, below=of mixup1, anchor=north, scale=0.8, transform shape] (cat_box) {
    \begin{tabular}{@{}l@{\hspace{0.1em}}l@{}}
    
    \begin{tabular}[t]{l} %
        \textbf{Category: Invariances} \\
        \hspace{0.3em}\textit{- Time Reversal} \\
        \hspace{0.3em}\textit{- Sign Inversion} \\[0.3em] %
        
        \textbf{Category: Structure} \\
        \hspace{0.3em}\textit{- Regime Change} \\
        \hspace{0.3em}\textit{- Shock \& Recovery} \\[0.3em]
        
        \textbf{Category: Artifacts} \\
        \hspace{0.3em}\textit{- Resampling Artifacts} \\
    \end{tabular} &
    
    \begin{tabular}[t]{l} %
        \textbf{Category: Seasonality} \\
        \hspace{0.3em}\textit{- Amplitude Modulation} \\
        \hspace{0.3em}\textit{- Calendar Effects} \\[0.3em]
        
        \textbf{Category: Analytic} \\
        \hspace{0.3em}\textit{- Differential Operators} \\
        \hspace{0.3em}\textit{- Integration} \\[0.3em]
        
        \textbf{Category: Discrete} \\
        \hspace{0.3em}\textit{- Censoring} \\
        \hspace{0.3em}\textit{- Quantization} \\
    \end{tabular}
    
    \end{tabular}
};
\node[process, below=of cat_box] (randconv) {Random Conv. Filter};
\node[process, below=of randconv] (mixup2) {TS-Mixup (Late)};
\node[process, below=of mixup2] (finishing) {Noise \& Scaling};
\node[process, below=of finishing, fill=orange!20] (selection) {Selection \& Storage};
\node[process, below=of selection, fill=green!20] (nanaug) {NaN Patterns (Online)};

\draw[arrow] (source) -- (mixup1);
\draw[arrow] (mixup1) -- (cat_box);
\draw[arrow] (cat_box) -- (randconv);
\draw[arrow] (randconv) -- (mixup2);
\draw[arrow] (mixup2) -- (finishing);
\draw[arrow] (finishing) -- (selection);
\draw[arrow] (selection) -- (nanaug);

\node[image, left=1.2cm of $(source.west)!0.5!(nanaug.west)$] (unaugmented)
    {\includegraphics[width=5.7cm]{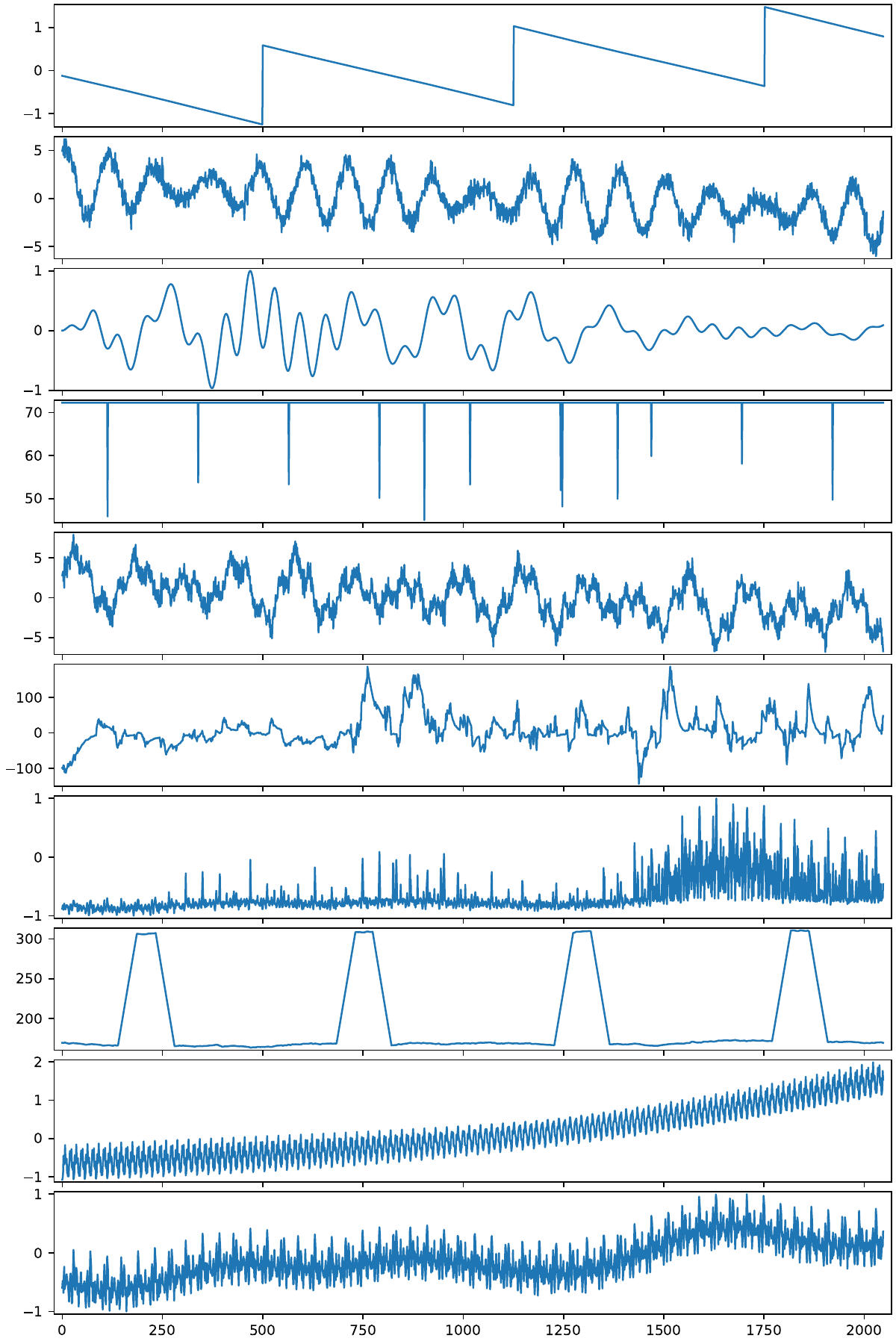}};
\node[image, right=1.2cm of $(source.east)!0.5!(nanaug.east)$] (augmented)
    {\includegraphics[width=5.7cm]{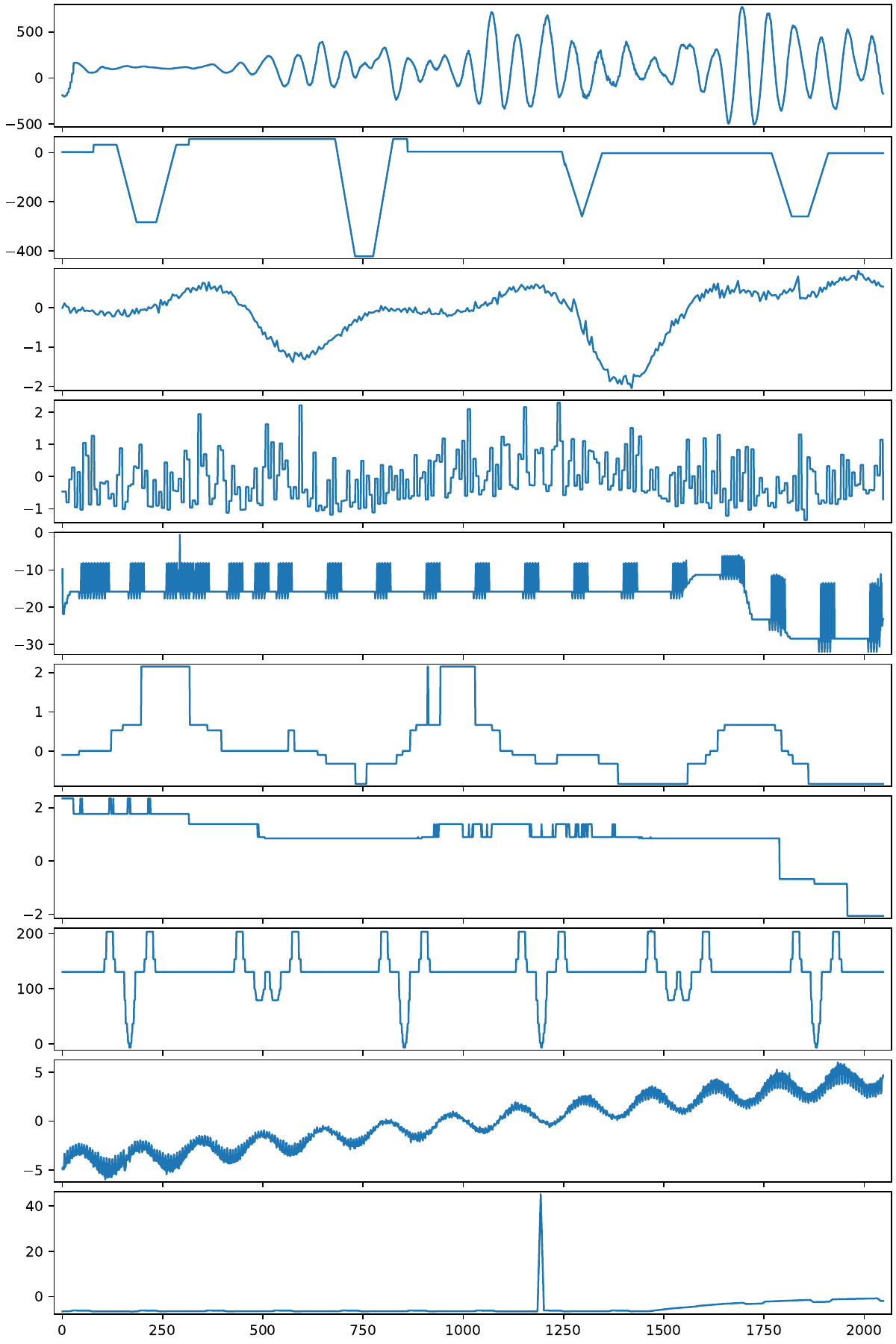}};

\draw[arrow] (unaugmented.east) .. controls +(40:0.5cm) and +(180:1cm) .. (source.west);
\draw[arrow] (nanaug.east) .. controls +(0:1.2cm) and +(180:1.2cm) .. (augmented.west);

\end{tikzpicture}
}
\vspace{-1mm}
\caption{Synthetic data augmentation pipeline. Synthetic time-series undergo probabilistic transformations across six categorical groups (Invariances, Structure, Seasonality, Analytic, Discrete, Artifacts), with optional TS-Mixup combinations and stochastic convolution filtering. Competitive selection based on change scores ensures meaningful augmentations. Example outputs demonstrate the diversity of generated temporal patterns.}
\label{fig:augmentations}
\vspace{-1em}
\end{figure}
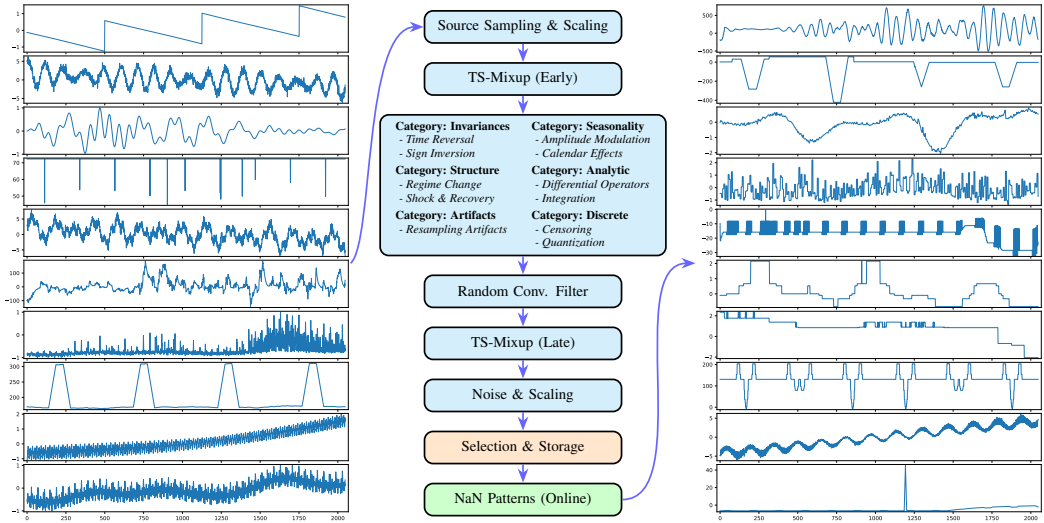

\vspace{-2mm}
In addition to diverse synthetic time-series generators, our pipeline (\Cref{fig:augmentations}) also contains a mix of existing and novel augmentations to mix, transform, and distort the time-series for greater diversity. 

\textbf{Augmentation Pipeline.}
The offline pipeline applies transformations in a structured sequence. Base series undergo optional normalization (80\% probability) using random scalers (Robust, MinMax, Median, or Mean). Early-stage TS-Mixup \citep{darlow2023tsmix} creates convex combinations of multiple source series with probability $p=0.5$. The core augmentation step samples 2-5 distinct transformation categories with weighted probabilities: Invariances (0.6), Structure (0.6), Seasonality (0.5), Signal Processing (0.4), Discrete Effects (0.6), and Measurement Artifacts (0.3). From each selected category, one specific transformation is randomly chosen and applied in fixed global order. Optional stochastic convolution filtering (probability $p=0.3$) applies 1-3 random 1D convolutions with randomized parameters. Late-stage TS-Mixup provides additional combination opportunities, followed by finishing transformations, including minor global scaling and low-magnitude Gaussian noise injection.
In the following, we provide details on the augmentations we implemented.

\textbf{Transformation Categories.}
\textbf{Invariance transformations} promote robustness through temporal reversal ($\mathbf{x} \rightarrow \mathbf{x}_{T:1}$) and sign inversion ($\mathbf{x} \rightarrow -\mathbf{x}$), preserving temporal dependencies while testing directional conventions. \textbf{Structural modifications} inject non-stationarity via regime changes with piecewise affine transforms across random change-points, and shock-recovery dynamics using exponential decay impulses $I(t) = A e^{-(t-t_0)/\tau}$ with randomized parameters. 
\textbf{Seasonal effects} simulate real-world periodicities through calendar injections that apply multiplicative factors for weekend dips, month-end spikes, and holiday-like impulses using timestamp metadata. Amplitude modulation applies localized scaling to random segments, simulating time-varying volatility. \textbf{Signal processing} transformations include Gaussian smoothing followed by finite-difference operators (Sobel, Laplacian, higher-order derivatives up to 4th order) and numerical integration, with outputs rescaled to preserve original value ranges. Random convolution layers with highly randomized parameters \citep{dempster2020rocket} provide additional signal transformation capabilities. 
\textbf{Measurement artifacts} introduce realistic data collection imperfections: censoring clips values at random quantiles (similarly used by TiRex~\citep{auer2025tirex}), non-uniform quantization maps values to discrete levels using quasi-random Sobol sequences, and resampling artifacts downsample and upsample series with various interpolation methods.  
\textbf{Combination strategies.} We implement TS-Mixup~\citep{ansari2024chronos} to generate novel series through convex combinations of 2-10 source series, with mixing weights sampled from Dirichlet distributions and extend it with time-dependent mixing using smooth simplex path interpolation.

\begin{figure}[t]
    \centering
    \begin{subfigure}{0.4\textwidth}
        \centering
        \includegraphics[width=\linewidth, trim={0 0 0 0}, clip]{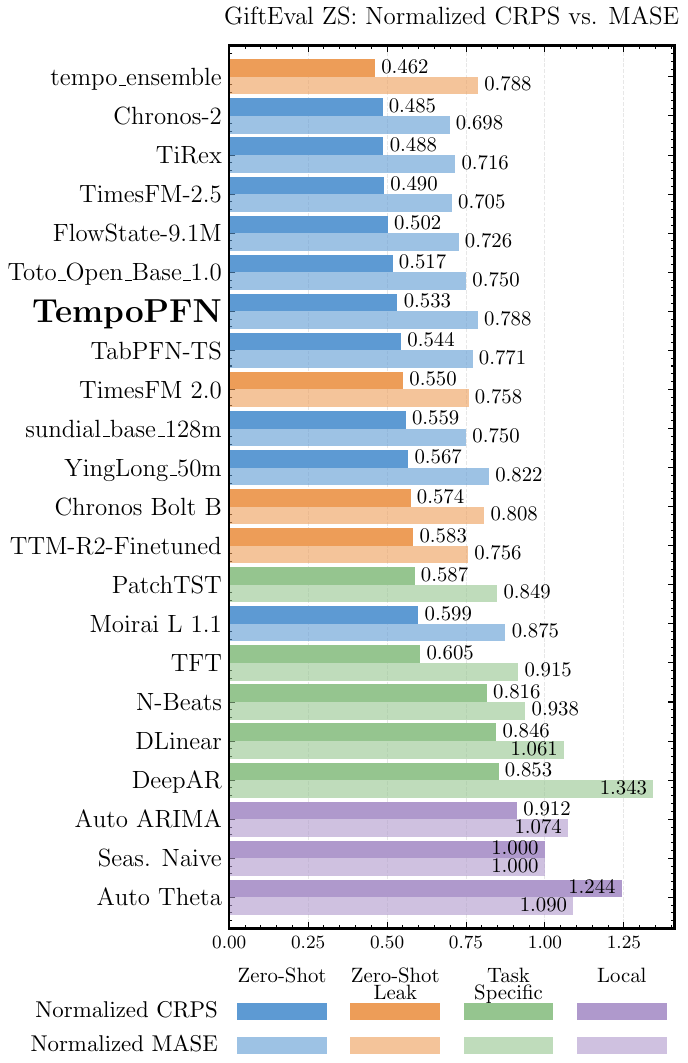} 
        \label{fig:normalized_scores}
    \end{subfigure}%
    \hspace{5mm}
    \begin{subfigure}{0.4\textwidth}
        \centering
        \includegraphics[width=\linewidth, trim={0 0 0 0}, clip]{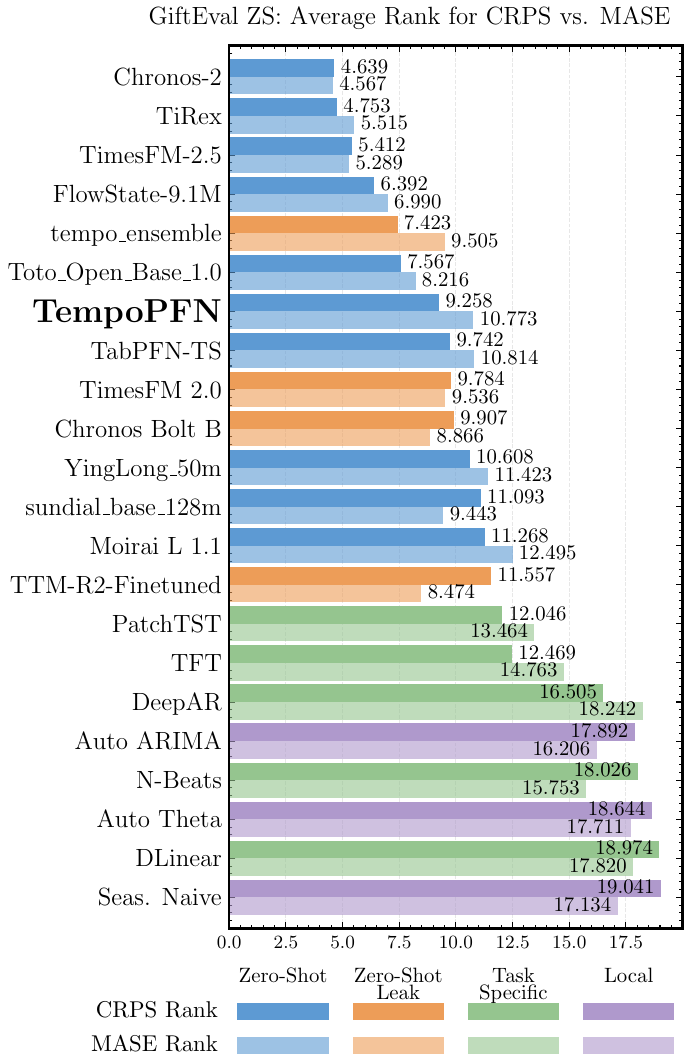}  
        \label{fig:average_ranks}
    \end{subfigure}
    \vspace{-5mm}
    \caption{Comparison of \methodname{} (3M iterations with a per-GPU batch size of 40) against other models on GiftEval benchmark. We compute both normalized and average ranks for CRPS and MASE. Colors represent the class of time series model.}
    \label{fig:performance_comparison}
\end{figure}

\section{Experiments}
\vspace{-2mm}
We ran experiments on three standard zero-shot forecasting benchmarks: Gift-Eval~\cite{aksu2024gifteval}, Chronos-ZS~\citep{ansari2024chronos}, and \texttt{fev-bench}~\citep{shchur2025fev}.

\textbf{Pretraining Setup.} \methodname{}'s pretraining is conducted \textbf{exclusively on synthetic data}, ensuring no exposure to real-world benchmarks prior to evaluation. The training corpus consists of approximately 10 million time series from our generators, each with a maximum length of 2048. We train our main model with the AdamW optimizer~\citep{loshchilov-iclr19a} to minimize the quantile loss, for a total of 3 million iterations and a batch size of 40. We selected a 40M model (10 layers, 4 heads, 512 embedding dimension) for its strong performance and comparability to TiRex~\citep{auer2025tirex}. To ensure robustness across sequence lengths, we randomly sample both the context length and historical window size during training. Complete training details are in \Cref{app:training-details}.

\subsection{Results on Gift-Eval}
\label{subsec:gifteval_results}

\textbf{Quantitative Results.}  
We evaluate \methodname{} on the Gift-Eval benchmark~\citep{aksu2024gifteval}, a comprehensive zero-shot forecasting suite covering 23 diverse real-world datasets across domains and horizons. \methodname{} surpasses probabilistic performance of TabPFN-TS, the strongest synthetic-only baseline, with an overall CRPS of 0.533 (vs. 0.544). Its point-forecast accuracy is competitive, though slightly lower, with an overall MASE of 0.788 (vs. 0.771). Remarkably, despite relying solely on synthetic training data, \methodname{} matches or exceeds several leading models trained on real-data, including Chronos Bolt B (0.574/0.808), TimesFM 2.0 (0.550/0.758), and YingLong 50m (0.567/0.822), ranking 7th overall in CRPS. Figure~\ref{fig:performance_comparison} summarizes these quantitative comparisons against state-of-the-art baselines.  

\begin{figure}
    \centering
    \includegraphics[width=0.9\linewidth, trim={0 23 0 0}, clip=true]{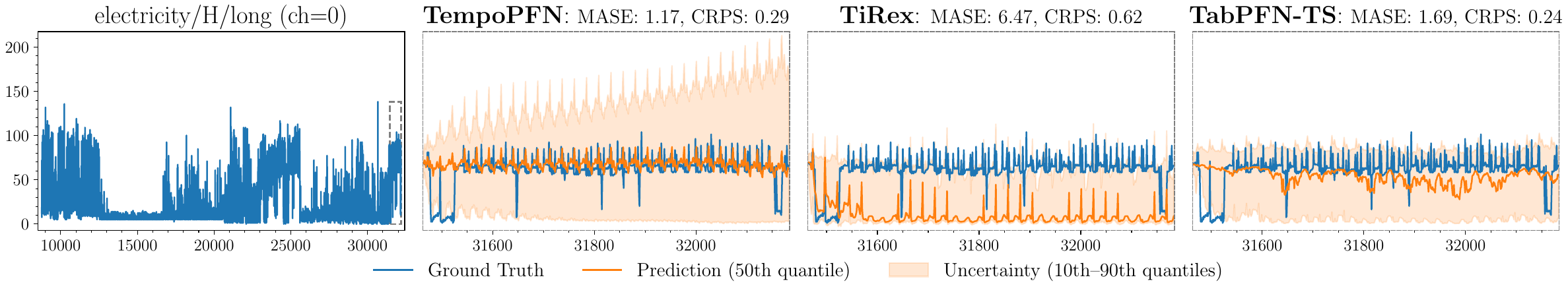}
    \includegraphics[width=0.9\linewidth]{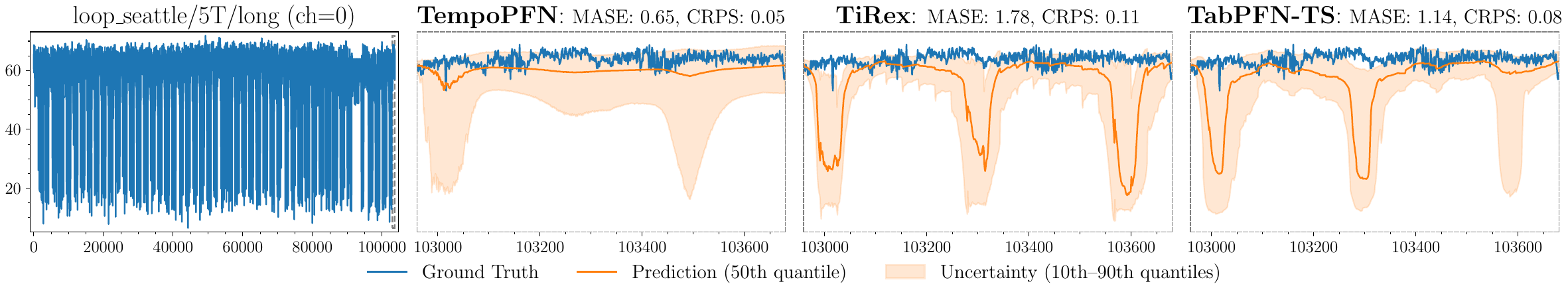}
    \vspace{-3mm}
    \caption{Qualitative comparison between \methodname{}, TiRex and TabPFN-TS on two series from the GIFT-Eval Benchmark. (Left) Total context with prediction window in dashed grey box. (Right) Predictions between TempoPFN, TiRex, and TabPFN-TS.}
    \label{fig:qualitative_results_excerpt}
\end{figure}

\begin{wrapfigure}[12]{R}{0.32\linewidth}
    \vspace{-4mm}
    \centering
    \includegraphics[width=0.95\linewidth]{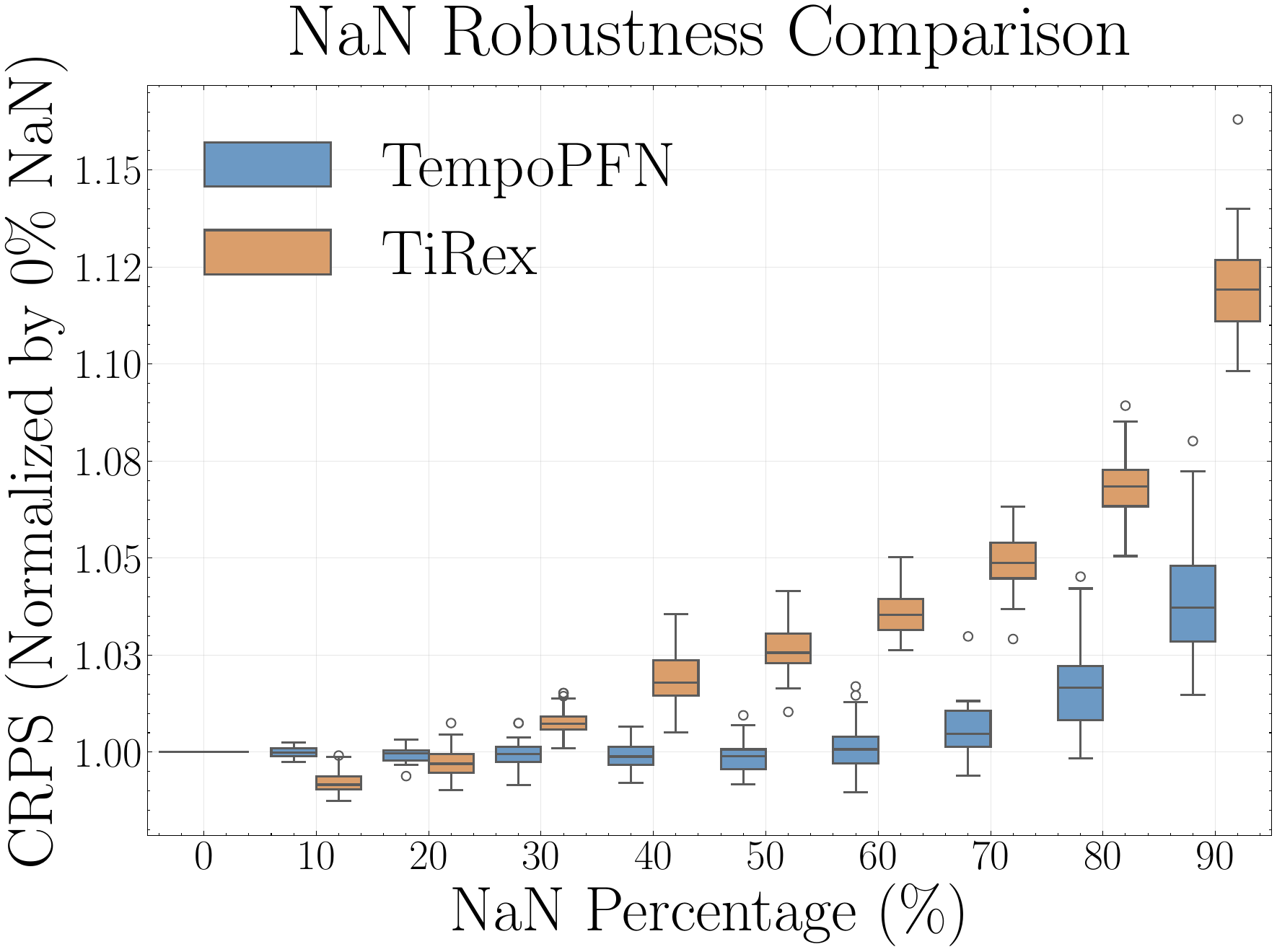}
    \vspace{-1mm}
    \caption{Normalized CRPS of \methodname{} and TiRex as a function of missing values (NaN) in the data.}
    \label{fig:nans}
\end{wrapfigure}
\textbf{Qualitative Results.}  
\Cref{fig:qualitative_results_excerpt} shows forecasting results on representative Gift-Eval series with varying temporal patterns (see~\Cref{appfig:qualitative} for additional results). All models capture key trends and seasonality, but \methodname{} produces coherent predictive distributions without artifacts. Compared to TabPFN-TS, \methodname{} generates smoother uncertainty bounds while maintaining competitive point forecasts. This is likely because TabPFN-TS predicts all future time steps in isolation, whereas our architecture allows future time steps to communicate. In many longer predictions made by TiRex (e.g. bizitobs\_service in \Cref{appfig:qualitative}), we find high-frequency artifacts in the prediction of the quantiles, which we hypothesize to be a result of the windowing done by TiRex which compresses the time series into chunks of size 32 before applying the model and later up-projects them back to the original resolution. Since \methodname{} requires no windowing, we did not notice similar artifacts.

\textbf{Robustness to NaNs.} We now compare the robustness of \methodname{} and TiRex towards missing values (NaN) in the data. Figure \ref{fig:nans} shows that both models exhibit a degradation in performance as the percentage of NaNs increases, however, TempoPFN is significantly more robust. While the normalized CRPS score (relative to \methodname{}'s CRPS at 0\% NaNs) for both models rises with more NaNs, TiRex's performance deteriorates more rapidly, with its median CRPS increasing by over 11\% when 90\% of the data is missing. In contrast, TempoPFN's median error increases by only about 4\% under the same conditions, showcasing its superior stability and resilience when faced with incomplete time series data.

\textbf{Per-dataset analysis.} The per-dataset results (\Cref{tab:normalized_crps,tab:normalized_mase,tab:crps_rank,tab:mase_rank} in Appendix~\ref{app:quantitative_results}) show that \methodname{} excels on event-driven, bursty series such as \texttt{bitbrains\_fast\_storage}, \texttt{bizitobs\_application}, and \texttt{covid\_deaths}, the target of our Spike, Anomaly, and Step generators. It underperforms on high-frequency series with multiple seasonalities and long horizons, e.g., \texttt{electricity/15T/long} (CRPS 1.010 vs.\ 0.656 for TiRex), \texttt{m\_dense/H/long} (0.438 vs.\ 0.291), and \texttt{solar/H}. This points to richer multi-seasonal generators as a natural extension of our prior, and shows that the prior is not tuned to the seasonal patterns of Gift-Eval.

\subsection{Results on \texttt{fev-bench}}
\label{subsec:fev_results}
\vspace{-2mm}
We also evaluate on the recent \texttt{fev-bench}~\citep{shchur2025fev}, which standardizes evaluation across 100 diverse forecasting tasks, as well as rigorously tracks data leakage and failure rates, making this benchmark more challenging than Gift-Eval and Chronos-ZS. \texttt{fev-bench} evaluates the model performance using two aggregate scores derived from the pairwise error matrices: (1) \textit{Win Rate (\%)}, representing the fraction of model pairs and tasks where the model achieves a lower error than the competitor; and (2) \textit{Skill Score (\%)}, a robust measure of relative error reduction compared to the Seasonal Naive baseline.

\textbf{Leaderboard Results.} Table~\ref{tab:fev_bench_results_mase} presents the leaderboard results for MASE, where \methodname{} is ranked fifth, evaluated with a sign-flip and context-length ensemble at a maximum context of 3072. We can see the same rank in Table~\ref{tab:fev_bench_results_sql} of Appendix~\ref{app:chronos_fevbench_results} for the \textit{Scaled Quantile Loss} (SQL), which captures calibration quality by evaluating the quality of the entire predictive distribution at each time step. In both metrics, our model achieves a higher win rate and rank than the other leading synthetic-only baseline, TabPFN-TS (Rank 8 in both), while TabPFN-TS retains a slightly higher skill score (SQL 39.6 vs.\ 38.3; MASE 27.6 vs.\ 25.7) at a roughly 15$\times$ higher median runtime. To also visualize relative strengths in probabilistic forecasting, we show in Figure~\ref{fig:pairwise_heatmaps_sql} in the appendix the head-to-head Win Rates and Skill Scores.

\textbf{Runtime.} The reported median runtime of 20.67\,s corresponds to the full sign-flip and context-length ensemble; a single forward pass takes about 8\,s. The remaining gap to TiRex (1.40\,s) stems from windowing: TiRex summarizes the input into windows of 32 time steps, so its sLSTM processes $32\times$ fewer tokens, whereas \methodname{} processes each of up to 3072 time steps individually. The wall-clock gap (${\approx}5.7\times$) is much smaller than this token ratio, since the GatedDeltaProduct recurrence is computed chunk-wise in parallel while the sLSTM recurrence is strictly sequential. In exchange, \methodname{} avoids the windowing artifacts discussed in \Cref{subsec:gifteval_results} and keeps horizon-independent latency and $\mathcal{O}(d)$ streaming memory (Appendix~\ref{app:complexity_analysis}).

\begin{table}[b]
\centering
\caption{\texttt{fev-bench} leaderboard based on MASE. Models are ranked by Win Rate and Skill Score. \methodname{} is evaluated with a sign-flip and context-length ensemble at a maximum context of 3072.}
\label{tab:fev_bench_results_mase}
\resizebox{\linewidth}{!}{%
\begin{tabular}{llccccccc}
\toprule
Rank & Model & Avg. Win Rate (\%) & Skill Score (\%) & Median Runtime (s) & Leakage (\%) & Failed Tasks (\%) & Organization & Zero-shot \\
\midrule
1 & Chronos-2 & 87.7 & 35.5 & 2.69 & 0 & 0 & AWS & \cmark \\
2 & TiRex & 76.5 & 30.0 & 1.40 & 1 & 0 & NX-AI & \cmark \\
3 & TimesFM 2.5 & 74.8 & 30.2 & 16.92 & 10 & 0 & Google & \cmark \\
4 & Toto 1.0 & 66.6 & 28.2 & 90.74 & 8 & 0 & Datadog & \cmark \\
\textbf{5} & \textbf{TempoPFN} & \textbf{62.4} & \textbf{25.7} & \textbf{20.67} & \textbf{0} & \textbf{0} & \textbf{Uni Freiburg / ELLIS} & \textbf{\cmark} \\
6 & Moirai 2.0 & 60.4 & 27.3 & 2.46 & 28 & 0 & Salesforce & \cmark \\
7 & Chronos-Bolt & 59.9 & 26.5 & 1.04 & 0 & 0 & AWS & \cmark \\
8 & TabPFN-TS & 58.0 & 27.6 & 305.51 & 0 & 2 & Prior Labs & \cmark \\
9 & Sundial-Base & 52.0 & 24.7 & 35.62 & 1 & 0 & Tsinghua University & \cmark \\
10 & Stat. Ensemble & 47.0 & 15.7 & 690.55 & 0 & 11 & — & \xmark \\
11 & AutoARIMA & 35.4 & 11.2 & 186.84 & 0 & 10 & — & \xmark \\
12 & AutoTheta & 33.5 & 11.0 & 9.27 & 0 & 0 & — & \xmark \\
13 & AutoETS & 32.5 & 2.3 & 16.96 & 0 & 3 & — & \xmark \\
14 & Seasonal Naive & 20.1 & 0.0 & 2.32 & 0 & 0 & — & \xmark \\
\bottomrule
\end{tabular}
}
\end{table}

\subsection{Architecture-Agnostic Ablation}
\label{subsec:backbone_vs_prior}
\vspace{-2mm}
A natural question is whether \methodname{}'s remaining gap to the top-ranked zero-shot models (Chronos-2, TiRex, TimesFM 2.5, Toto) stems from its linear-RNN backbone or from the synthetic prior. To disentangle these two factors, we pre-train \texttt{chronos-2-small}~\citep{chronos2small_hf}, a 28M-parameter Transformer with a fundamentally different inductive bias from \methodname{}, \emph{from scratch} on the exact same univariate synthetic corpus and with the same 500k-iteration training budget as our other ablations (Table~\ref{tab:leave_one_out_ablations}). All other choices — context sampling, augmentation pipeline, optimizer, and quantile loss — are held fixed. As shown in \Cref{tab:ablation_chronos2_small}, the difference between the backbones is negligible (0.7\% relative CRPS), whereas removing a single generator changes CRPS by 2.4\% (GP) to 26\% (SDE) (Table~\ref{tab:leave_one_out_ablations}). \methodname{} is better on short and medium horizons, the Transformer on long horizons. Under our prior, the synthetic data distribution is thus the primary driver of accuracy, while the linear-RNN backbone makes \methodname{} parallelizable, robust to missing values, and free of windowing artifacts.

\begin{table}[t]
\centering
\caption{
    Backbone swap. Both models are pre-trained \emph{from scratch} on the same univariate synthetic corpus for the matched 500k-iteration ablation budget (as in Table~\ref{tab:leave_one_out_ablations}). \methodname{} uses our GatedDeltaProduct linear-RNN backbone (40M parameters); \texttt{chronos-2-small}~\citep{chronos2small_hf} is a 28M-parameter Transformer. Lower is better.
}
\label{tab:ablation_chronos2_small}
\small
\setlength{\tabcolsep}{3.5pt}
\resizebox{0.9\linewidth}{!}{
\begin{tabular}{lcccccccc}
\toprule
& \multicolumn{2}{c}{\textbf{Gift-ZS Overall}} & \multicolumn{2}{c}{\textbf{Gift-ZS Short}} & \multicolumn{2}{c}{\textbf{Gift-ZS Medium}} & \multicolumn{2}{c}{\textbf{Gift-ZS Long}} \\
\cmidrule(lr){2-3} \cmidrule(lr){4-5} \cmidrule(lr){6-7} \cmidrule(lr){8-9}
\textbf{Model} & \textbf{CRPS} & \textbf{MASE} & \textbf{CRPS} & \textbf{MASE} & \textbf{CRPS} & \textbf{MASE} & \textbf{CRPS} & \textbf{MASE} \\
\midrule
\methodname{} (Linear RNN, 40M)             & \textbf{0.577} & \textbf{0.842} & \textbf{0.563} & \textbf{0.763} & \textbf{0.566} & \textbf{0.900} & 0.631 & 1.019 \\
\texttt{chronos-2-small} (Transformer, 28M) & 0.581 & 0.849 & 0.578 & 0.784 & 0.571 & 0.913 & \textbf{0.619} & \textbf{1.001} \\
\midrule
seasonal\_naive                          & 1.000 & 1.000 & 1.000 & 1.000 & 1.000 & 1.000 & 1.000 & 1.000 \\
\bottomrule
\end{tabular}
}
\vspace{-3mm}
\end{table}

\subsection{The Importance of \methodname{}'s Components}
\label{subsec:ablations}
\vspace{-2mm}
We conduct several ablation studies on TempoPFN's building blocks.
Due to significant pretraining costs, not all ablation experiments were conducted on the full 3M-iteration training schedule. 

\textbf{Ablating the synthetic time series generators.} To assess the individual contribution of each synthetic data source, we conducted an ablation study by retraining our model while excluding one synthetic time series generator at a time. As detailed in Table~\ref{tab:leave_one_out_ablations}, the results reveal a clear hierarchy of importance, consistently observed across short, medium, and long-term forecasting horizons, with every generator proving beneficial for high performance. The highest impact data generator is our proposed SDE generator; its removal caused the most severe performance degradation, increasing the overall CRPS by 26\% from 0.577 to 0.729. This highlights the importance of exposing the model to time series with mean-reverting and noisy, continuous-time dynamics. Significant, albeit smaller, performance losses were also observed upon removing generators responsible for complex seasonality (Cauker), abrupt changes (Step), and transient events (Spike), underscoring the necessity of a diverse pre-training corpus that captures a wide array of structural and stochastic patterns. In Table~\ref{tab:single_prior_ablations} in the appendix, we also compare our base model trained using all generators with models trained using a single generator at a time. The full mixture outperforms the best single generator (CauKer, 0.600) and every leave-one-out variant, showing that the generators complement each other. The generator set is not closed: a new generator should improve performance if it adds dynamics not yet covered, such as the multi-seasonal structure discussed in \Cref{subsec:gifteval_results}.

\begin{table*}[t]
\caption{
    Ablation study of synthetic data priors using a leave-one-out methodology (500k iterations). The 'Ablation' column indicates the single prior excluded from the training mixture. Performance is measured by CRPS and MASE (lower is better). Rows are colored to indicate the performance impact on the overall CRPS when a prior is removed: \colorbox{impacthigh}{High Impact} ($> 25\%$ increase) and \colorbox{impactmedium}{Medium Impact} ($> 10\%$ increase). \small\textbf{N} = Novel prior (our contribution), \small\textbf{A} = Adapted from open-source.
}
\vspace{-2mm}
\label{tab:leave_one_out_ablations}
\centering
\sisetup{detect-weight=true, detect-family=true} %
\setlength{\tabcolsep}{6pt} %
\small %
\resizebox{0.9\textwidth}{!}{
\begin{tabular}{
    l %
    c %
    S[table-format=1.3, table-number-alignment=center]
    S[table-format=1.3, table-number-alignment=center]
    S[table-format=1.3, table-number-alignment=center]
    S[table-format=1.3, table-number-alignment=center]
    S[table-format=1.3, table-number-alignment=center]
    S[table-format=1.3, table-number-alignment=center]
    S[table-format=1.3, table-number-alignment=center]
    S[table-format=1.3, table-number-alignment=center]
}
\toprule
& & \multicolumn{2}{c}{\textbf{Gift-ZS Overall}} & \multicolumn{2}{c}{\textbf{Gift-ZS Short}} & \multicolumn{2}{c}{\textbf{Gift-ZS Medium}} & \multicolumn{2}{c}{\textbf{Gift-ZS Long}} \\
\cmidrule(lr){3-4} \cmidrule(lr){5-6} \cmidrule(lr){7-8} \cmidrule(lr){9-10}
\textbf{Ablation} & \textbf{Source} & {\textbf{CRPS}} & {\textbf{MASE}} & {\textbf{CRPS}} & {\textbf{MASE}} & {\textbf{CRPS}} & {\textbf{MASE}} & {\textbf{CRPS}} & {\textbf{MASE}} \\
\midrule
Base Model            & -- & 0.577 & 0.842 & 0.563 & 0.763 & 0.566 & 0.900 & 0.631 & 1.019 \\
\midrule %
- GP                  & \textbf{A} & 0.591 & 0.830 & 0.576 & 0.749 & 0.605 & 0.924 & 0.618 & 0.981 \\
- Kernel              & \textbf{A} & 0.611 & 0.885 & 0.589 & 0.796 & 0.637 & 0.981 & 0.648 & 1.056 \\
- ForecastPFN         & \textbf{A} & 0.617 & 0.885 & 0.588 & 0.791 & 0.643 & 0.981 & 0.674 & 1.075 \\
- Sawtooth            & \textbf{N} & 0.628 & 0.900 & 0.597 & 0.800 & 0.661 & 1.012 & 0.684 & 1.091 \\
- Sinewave            & \textbf{N} & 0.628 & 0.899 & 0.594 & 0.799 & 0.677 & 1.032 & 0.676 & 1.070 \\
- Anomaly             & \textbf{N} & 0.630 & 0.897 & 0.592 & 0.794 & 0.684 & 1.024 & 0.683 & 1.079 \\
\rowcolor{impactmedium}
- Step                & \textbf{N} & 0.640 & 0.927 & 0.605 & 0.819 & 0.686 & 1.063 & 0.689 & 1.120 \\
\rowcolor{impactmedium}
- Stochastic Rhythm   & \textbf{N} & 0.642 & 0.911 & 0.601 & 0.802 & 0.701 & 1.043 & 0.699 & 1.111 \\
\rowcolor{impactmedium}
- Spike               & \textbf{N} & 0.645 & 0.936 & 0.619 & 0.836 & 0.678 & 1.059 & 0.684 & 1.115 \\
\rowcolor{impactmedium}
- Cauker              & \textbf{A} & 0.656 & 0.928 & 0.605 & 0.810 & 0.728 & 1.084 & 0.729 & 1.132 \\
\rowcolor{impacthigh}
- SDE (OU Process)    & \textbf{N} & \bfseries 0.729 & \bfseries 1.031 & \bfseries 0.684 & \bfseries 0.916 & \bfseries 0.799 & \bfseries 1.184 & \bfseries 0.789 & \bfseries 1.225 \\
\midrule
seasonal\_naive       & -- & 1.000 & 1.000 & 1.000 & 1.000 & 1.000 & 1.000 & 1.000 & 1.000 \\
\bottomrule
\end{tabular}
}
\end{table*}

\begin{wraptable}[11]{R}{0.6\linewidth}
\vspace{-4.5mm} 
\caption{The impact of data augmentation on model performance, shown by normalized performance values (500k iterations). Individual augmentation effects and probability tuning were not exhaustively explored due to resource constraints. }
\vspace{-2mm}
\label{tab:augmentation_normalized_impact}
\centering
\small
\setlength{\tabcolsep}{3.5pt}
\resizebox{0.99\linewidth}{!}{
\begin{tabular}{lcccccccc}
\toprule
& \multicolumn{2}{c}{\textbf{Gift-ZS Overall}} & \multicolumn{2}{c}{\textbf{Gift-ZS Short}} & \multicolumn{2}{c}{\textbf{Gift-ZS Medium}} & \multicolumn{2}{c}{\textbf{Gift-ZS Long}} \\
\cmidrule(lr){2-3} \cmidrule(lr){4-5} \cmidrule(lr){6-7} \cmidrule(lr){8-9}
\textbf{Model} & \textbf{CRPS} & \textbf{MASE} & \textbf{CRPS} & \textbf{MASE} & \textbf{CRPS} & \textbf{MASE} & \textbf{CRPS} & \textbf{MASE} \\
\midrule
w/ Aug & \textbf{0.577} & \textbf{0.842} & \textbf{0.563} & \textbf{0.763} & \textbf{0.566} & \textbf{0.900} & \textbf{0.631} & \textbf{1.019} \\
w/o Aug & 0.610 & 0.875 & 0.582 & 0.783 & 0.617 & 0.963 & 0.643 & 1.059 \\
\midrule
seasonal\_naive & 1.000 & 1.000 & 1.000 & 1.000 & 1.000 & 1.000 & 1.000 & 1.000 \\
\bottomrule
\end{tabular}
}
\end{wraptable}

\textbf{Ablating the augmentation pipeline.}
To quantify the impact of augmentations, we trained models with and without the full augmentation suite (see Table~\ref{tab:augmentation_normalized_impact}). Results on the GIFT-Eval benchmark reveal consistent gains from our augmentation pipeline, yielding a ~5.4\% relative improvement in overall CRPS and a ~3.8\% improvement in overall MASE. These gains are present across all forecasting horizons, underscoring the benefit of our complex data augmentation pipeline for extrapolation to real-world data.

\textbf{Ablating architectural components.} Results on architectural ablations are provided in Table \ref{tab:scale_depth_ablations} in the Appendix. Specifically, Table \ref{tab:scale_depth_ablations} shows that disabling 'weaving' in our main model (d=512, L=10) leads to a small performance degradation, increasing the overall CRPS from 0.533 to 0.537, and negative eigenvalues change it by at most 0.002 (Table~\ref{tab:architectural_ablations_v2}). As we did not run repeated seeds, we view both as low-cost design choices with a mechanistic rationale rather than as leading accuracy drivers: weaving avoids the causal-RNN bottleneck at long horizons, and negative eigenvalues enable state tracking beyond diagonal recurrences. Clearer effects come from the number of Householder matrices ($n_h=1$ costs +2.1\% CRPS) and the positional encoding (sinusoidal costs +15.5\%).
\vspace{-2mm}
\section{Conclusion and Future Work}\label{sec:conclusion}
\vspace{-2mm}
We introduce \emph{\methodname{}}, a novel time series foundation model demonstrating that linear RNNs, specifically the \textit{GatedDeltaProduct} architecture, provide a highly efficient and scalable solution for zero-shot forecasting. By enabling parallelizable training, our model processes long sequences without patching or summarization heuristics. \methodname{} is trained exclusively on our open-source synthetic data generation pipeline, which integrates diverse generators and a complex augmentation framework. This synthetic-only approach ensures full reproducibility and eliminates data leakage concerns. On the Gift-Eval, \texttt{fev-bench} and Chronos-ZS benchmarks, \methodname{} achieves top-tier competitive performance, surpassing the vast majority of models trained on real-world data, establishing linear RNNs as a powerful and scalable alternative to prevailing architectures. Our backbone swap further shows that the synthetic prior is the main driver of accuracy, while the linear RNN makes the model parallelizable, robust to missing values, and free of windowing artifacts. A key limitation of our current work is its focus on univariate time series. Extending our synthetic generation pipeline and state-weaving architecture to the more complex multivariate case represents a primary direction for future work: our CauKer generator already samples 21-channel series from a structural causal model, and retaining these dependencies is a direct path to synthetic covariates, while multivariate inputs could be handled by 2D architectures as in Chronos-2~\citep{ansari2025chronos2}. Additionally, incorporating pre-training on diverse real-world time series datasets could further enhance forecasting accuracy and generalization. Finally, investigating the performance of Linear RNN architectures against Transformer-based models for zero-shot forecasting represents a potential direction for further research.

\clearpage
\newpage

\section*{Author contributions}
Vladyslav Moroshan led TempoPFN’s development, implemented all synthetic priors and many data augmentations, carried out all data generation, pre-training, ablations, and experiments, conducted the NaN-robustness, feature-space, and generator-throughput analyses, proposed the sign-flip and context-length ensembling, submitted TempoPFN to the benchmarks, and led the software engineering and the open-source release. Julien Siems conceived and supervised the project, led the architectural design including the state-weaving idea, contributed data augmentations, and proposed the audio prior. Arber Zela proposed the SDE prior and later assumed project supervision. Timur Carstensen provided technical support for large-scale experiments on compute clusters. Frank Hutter provided project supervision, guidance on the paper’s direction, and manuscript revisions. Moroshan, Siems, and Zela contributed to manuscript writing and the rebuttal process.

\section*{Acknowledgements and Disclosures}
\textbf{Funding.} This research was partially supported by the following sources: PNRR MUR Project PE000013 CUP J53C22003010006 ``Future Artificial Intelligence Research (FAIR)'', funded by the European Union -- NextGenerationEU, and EU Project ELSA under grant agreement No.~101070617; TAILOR, a project funded by EU Horizon 2020 research and innovation programme under GA No.~952215; the Deutsche Forschungsgemeinschaft (DFG, German Research Foundation) under grant number 417962828; the European Research Council (ERC) Consolidator Grant `Deep Learning 2.0' (grant no.~101045765). This research was partially funded by the Deutsche Forschungsgemeinschaft (DFG, German Research Foundation) under grant number 539134284, through EFRE (FEIH 2698644) and the state of Baden-Württemberg. Frank Hutter acknowledges financial support by the Hector Foundation. The authors acknowledge support from ELLIS and ELIZA. The authors gratefully acknowledge the computing time made available to them on the high-performance computers at the NHR Centers at TU Dresden and KIT. These centers are jointly supported by the Federal Ministry of Research, Technology and Space of Germany and the state governments participating in the NHR.
Funded by the European Union. Views and opinions expressed are however those of the author(s) only and do not necessarily reflect those of the European Union or the ERC. Neither the European Union nor the ERC can be held responsible for them.

\textbf{Competing interests.} The authors declare no competing interests.

\vspace{-5pt}
\begin{figure}[h]
\begin{center}
\includegraphics[width=0.18\textwidth]{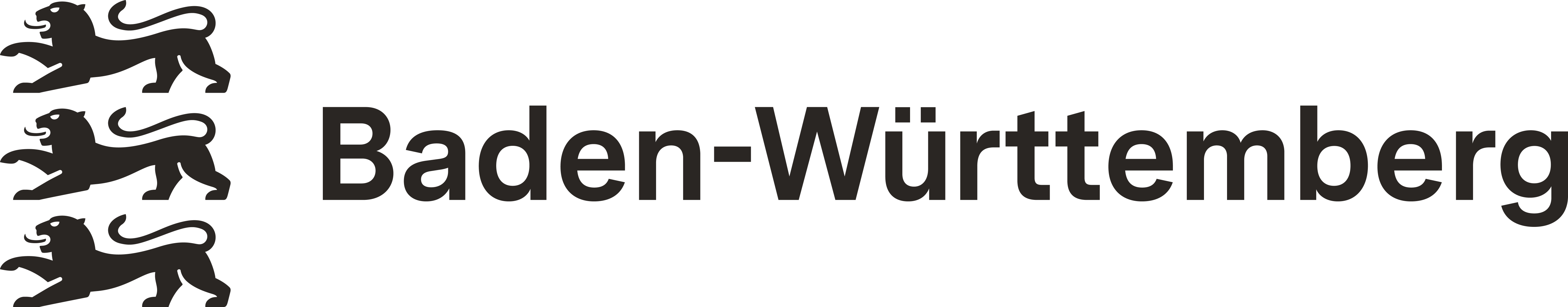} ~~~ \includegraphics[width=0.18\textwidth]{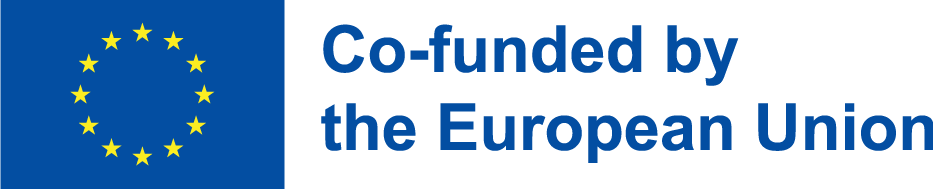} 
\end{center}
\end{figure}

\bibliography{arxiv/iclr2026_conference,bib/lib,bib/proc,bib/strings}
\bibliographystyle{iclr2026/iclr2026_conference}

\appendix

\newpage

\section{Details on the Design Principles of TempoPFN}\label{app:theory}

\subsection{Background: Prior Data Fitted Networks (PFNs)}
\label{app:pfn_background}

Prior Data Fitted Networks (PFNs)~\citep{muller2022transformers} represent a paradigm shift in machine learning, moving from learning a single fixed task to learning a \textit{universal inference algorithm}. In this section, we provide a brief overview of the PFN framework to contextualize the methodology used in \methodname{}.

\paragraph{Definition and Objective.}
A PFN is a neural network $\phi$, with parameters $\theta$, trained to approximate the posterior predictive distribution (PPD) induced by a prior distribution $P(\mathcal{D})$ over datasets. Formally, let a dataset $\mathcal{D} = \{(x_i, y_i)\}_{i=1}^N$ be drawn from a prior $P$. The goal of a PFN is to minimize the Kullback-Leibler divergence (or equivalently, the cross-entropy loss) between its output and the true posterior predictive distribution of the prior:
\begin{equation}
    \min_\theta \mathbb{E}_{\mathcal{D} \sim P} \left[ \sum_{i=1}^N -\log p_\theta(y_i \mid x_i, \mathcal{D}_{1:i-1}) \right]
\end{equation}
where $\mathcal{D}_{1:i-1}$ represents the context (history) observed so far.

\paragraph{In-Context Learning as Bayesian Inference.}
Unlike traditional supervised learning, where the model's weights $\theta$ encode the solution to a specific task (e.g., "predict sales for Company X"), a PFN's weights encode the \textit{algorithm} for solving a class of tasks defined by the prior. At inference time, the PFN performs \textit{In-Context Learning} (ICL): it takes a small dataset of context observations (the history of a time series) and produces predictions for new query points (the future) in a single forward pass. Crucially, this forward pass acts as a fast approximation of Bayesian Inference without the computational cost of MCMC or variational methods~\citep{muller2022transformers, hollmann2023tabpfn}.

\paragraph{The Role of Synthetic Data.}
The performance of a PFN is fundamentally bounded by the quality and diversity of its prior $P$. Since real-world data is often limited, biased, or private, PFNs are typically trained on vast repositories of \textit{synthetic data} generated from procedural priors. For example, TabPFN~\citep{hollmann2023tabpfn} uses Structural Causal Models (SCMs) to generate tabular data, while ForecastPFN~\citep{dooley2023forecastpfn} uses a mix of trends and seasonalities. Similarly, in \methodname{} the "Prior" is the novel synthetic data pipeline detailed in Appendix~\ref{app:synthetic_details}, which generates diverse temporal dynamics using SDEs, GPs, and asymmetric waveforms. Our "Network" is a Linear RNN, namely GatedDeltaProduct~\citep{siems2025deltaproduct}, chosen for its efficiency in handling long sequential contexts compared to the Transformers used in previous PFNs. \textit{By training the sequence model on this synthetic prior, \methodname{} learns to infer the underlying generative process of \textit{any} unseen time series given its history, enabling zero-shot probabilistic forecasting.}

\subsection{Design Principles for Synthetic Data Generation}
\label{app:design_principles}

Our selection of synthetic data generators is grounded in the principle of \textit{structural decomposition}. We posit that the manifold of real-world time series can be spanned by four fundamental dynamical properties: \textit{Smoothness}, \textit{Stochastic Volatility} (Roughness), \textit{Temporal Asymmetry}, and \textit{Discontinuities}. Existing synthetic pipelines often over-index on the first (trends/seasonality)~\citep{dooley2023forecastpfn} while neglecting the latter three. We designed a principled portfolio of generators to act as orthogonal "basis functions" for these properties, ensuring our prior distribution covers the complex dynamics found in downstream tasks.

\textbf{Smooth Dynamics (Gaussian Processes).} 
To capture non-parametric trends and local correlations, we employ Gaussian Processes (GPs). Real-world data is dominated by latent trends that evolve smoothly but unpredictably, such as demographic shifts or climate warming, which cannot be captured by rigid linear or polynomial functions. GPs with RBF or Matérn kernels serve as the standard for modeling such smooth, differentiable manifolds, providing the model with a robust prior for interpolation and extrapolation of continuous trends.

\textbf{Stochastic Volatility and Roughness (SDEs).} 
A critical deficiency in standard synthetic pipelines is the reliance on additive Gaussian noise ($\epsilon \sim \mathcal{N}(0,1)$), which implies homoscedasticity (constant variance). However, financial and physical systems are inherently \textit{heteroscedastic}, exhibiting state-dependent volatility and mean-reverting dynamics. To fill this gap, we integrate Stochastic Differential Equations (SDEs), specifically Regime-Switching Ornstein-Uhlenbeck (OU) processes. By explicitly modeling the diffusion term $\sigma(t, y_t)$, we force the model to learn to distinguish between deterministic signal drift and stochastic volatility clustering, a capability essential for accurate uncertainty quantification.

\textbf{Asymmetric Periodicity (Sawtooth Waveforms).} 
Standard sinusoidal generators rely on an assumption of \textit{time-reversal symmetry} (equal rise and fall times). Yet, many physical and economic processes are inherently asymmetric and irreversible: inventory levels deplete gradually and restock instantaneously; capacitors discharge rapidly. We selected the \textit{Sawtooth} wave as the fundamental primitive for asymmetry. Unlike Triangle waves (symmetric) or Square waves (step functions), the Sawtooth explicitly models the gradual-rise/sharp-drop dynamic. 

\textbf{Discontinuities and Structural Breaks (Spikes/Steps).} 
Finally, real-world data is rife with instantaneous regime changes: policy shifts, sensor failures, or sudden shocks. All of these violate the smoothness assumptions of GPs and SDEs. To model these structural breaks, we include explicit \textbf{Step} and \textbf{Spike} generators. Including these non-differentiable primitives ensures the model remains robust to covariate shifts and prevents the "smearing" of distinct regimes into a single average.

By composing these four distinct dynamical behaviors, \methodname{} has access to a \textit{complex prior}, allowing it to generalize in a zero-shot manner to unseen time series by identifying the governing combination of dynamics (related to the state-tracking capabilities of GatedDeltaProduct too), rather than memorizing dataset-specific statistics.

\subsection{Comparison with Existing Synthetic Strategies}
\label{app:strategy_comparison}

Freq-Synth~\citep{nochumsohn2024beyond} and TabPFN-TS~\citep{hoo2024the} are two recent methods that employ only synthetic data training as well. However, their methodologies represent fundamentally different paradigms in synthetic data generation and usage compared to \methodname{}. Freq-Synth adopts a task-specific generation strategy, hence requiring prior knowledge of the target dataset's sampling rate to generate a custom training corpus of harmonic signals (sums of sinusoids) tailored to mitigate data scarcity for that specific task. In contrast, in \methodname{}, we pretrain a single model on a fixed, comprehensive corpus designed to marginalize over diverse temporal dynamics without requiring task-specific data generation. This distinction places \methodname{} in a unique position within the broader landscape of synthetic time series data \citep{liu2025empowering}. While most existing synthetic pre-training methods (e.g., Chronos, TimesFM) rely on standard statistical components like GPs or ARMA processes, \methodname{} explicitly expands this design space by introducing novel generators for stochastic volatility (via SDEs) and temporal asymmetry (via sawtooth waves). Interestingly, TabPFN-TS relies on cross-domain adaptation, leveraging a model pre-trained on synthetic tabular data (via structural causal models) to effectively "feature-engineer" time series problems into tabular regression tasks, rather than learning temporal dynamics directly.

Parallel to this work, \citet{graf2025flowstate} introduced \textit{FlowState}, a time series foundation model that also leverages State Space Models (SSMs) for subquadratic computational efficiency. FlowState features an SSM-based encoder combined with a functional basis decoder to enable sampling-rate invariance and continuous-time modeling. While both FlowState and \methodname{} move away from Transformer-based architectures in favor of linear recurrences, our approaches differ fundamentally in their pre-training data paradigms. FlowState is pre-trained on a combination of real-world datasets (subsets of GIFT-Eval and Chronos corpora), whereas \methodname{} establishes the viability of a \textit{purely synthetic} pre-training pipeline. Furthermore, while FlowState focuses on resolution adaptation, \methodname{} focuses on maximizing zero-shot generalization through a diverse synthetic prior designed and utilizing the GatedDeltaProduct for state tracking.

\subsection{Architecture-Agnostic Ablation: the Synthetic Prior is the Bottleneck}
\label{app:chronos2_small_ablation}

The setup and results of this ablation are reported in \Cref{subsec:backbone_vs_prior} and \Cref{tab:ablation_chronos2_small}. On Gift-ZS, swapping the GatedDeltaProduct linear RNN for a Transformer of comparable capacity does \emph{not} close the gap to the leaderboard leaders: the two backbones differ by less than 3\% on every horizon. Since both models share the prior, the prior is the binding constraint. This is consistent with the PFN view (\Cref{app:pfn_background}) that the performance ceiling of an in-context learner is set by the support and fidelity of $P(\mathcal{D})$, and it positions architecture choices (linear RNN vs.\ Transformer) primarily as efficiency trade-offs rather than as accuracy levers under the current prior. Conversely, the leave-one-out study in Table~\ref{tab:leave_one_out_ablations} shows that the prior is highly responsive to additional dynamical primitives (notably SDE-based stochastic volatility), suggesting that the most impactful next step for closing the gap is enriching the synthetic generation pipeline rather than scaling or replacing the backbone.

\section{Details on Architectural Mechanisms}\label{app:architecture}

\subsection{GatedDeltaProduct Architecture and State Tracking}
\label{app:gated_deltaproduct}

To overcome the expressivity limitations of diagonal linear RNNs while retaining linear-time parallel scan computation, \methodname{} uses the \textit{GatedDeltaProduct} recurrence~\citep{siems2025deltaproduct}. Unlike diagonal SSMs such as Mamba~\citep{gu2023mamba} or RWKV~\citep{peng2023rwkv}, whose state-transition matrices are restricted to diagonal structure, GatedDeltaProduct employs a structured non-diagonal transition matrix constructed as a product of generalized Householder matrix updates. This yields a more expressive class of linear operators while keeping both training and inference efficient.

\paragraph{Recurrence Mechanism.}
Each layer maintains a matrix-valued hidden state $\mH_t \in \mathbb{R}^{d \times n}$, updated via a linear recurrence:
\begin{equation}
    \mH_t = \mA_t \mH_{t-1} + \mB_t, 
    \qquad
    \vy_t = \mH_t \vx_t.
\end{equation}
Here, $x_t \in \mathbb{R}^D$ represents the input vector at the current time step $t$, where $D$ is the input dimension (e.g., number of features). $\mH_t \in \mathbb{R}^N$ is the hidden state vector, representing the compressed memory, where $N$ is the state dimension. $\mA_t \in \mathbb{R}^{N \times N}$ is the state-transition matrix, which defines how the information from the previous hidden state evolves and is stored in the new state. DeltaProduct utilizes a dense, parameterized matrix $\mA_t$, which is explicitly constrained to be near orthogonal through its initialization and parameterization scheme (a product of $n_h$ rank-1 Householder-like updates):
\begin{equation}
    \mA_t = g_t \prod_{j=1}^{n_h} \left(\mI - \beta_{t,j} \vk_{t,j} \vk_{t,j}^\top \right),
\end{equation}
where $g_t \in [0,1]$ is a forget gate, $\vk_{t,j}$ are normalized key vectors, and $\beta_{t,j}$ are step sizes predicted from the input. The orthogonality constraint ensures that the transformation applied to the hidden state, $\mH_{t-1}$, preserves its magnitude. Consequently, the hidden state $\mH_{t}$ can effectively maintain its stability and information content over extended time steps, enabling robust state tracking of long-term trends and cyclical patterns in time series forecasting. Because each factor $\mI - \beta \vk\vk^\top$ is a rank-1 update, the full matrix-vector product remains $\mathcal{O}(d n_h)$ and is fully parallelizable via a parallel prefix scan~\citep{yang2024parallelizing}. Finally, $\mB_t\in \mathbb{R}^{N \times D}$ is the input weight matrix. This matrix transforms the current input $x_t$ and integrates it into the hidden state $\mH_t$.

\paragraph{Gating.}
The gating in GatedDeltaProduct was introduced to embed essential non-linearity in the basic linear recurrence above, therefore enhancing the model's overall expressiveness and selective memory. After the main linear recurrence is performed, its output is processed along two parallel streams: 1) \textit{Main Stream}: The result of the linear recurrence, intended for the final output; 2) \textit{Gate Stream}: The same recurrent output is passed through a non-linear activation function (e.g., the SiLU or Swish function). These two streams are combined via element-wise multiplication (the gate). This operation selectively controls which parts of the recurrent output are emphasized or suppressed, mirroring the functionality of sophisticated recurrent units like the Gated Recurrent Unit (GRU) or the selectivity found in Mamba's architecture.

\paragraph{State Weaving.}
This mechanism was specifically designed for the overall multi-layer structure of the TempoPFN framework where multiple GatedDeltaProduct layers are stacked. Instead of simply discarding the final hidden state, the mechanism \textit{weaves} temporal information across the depth of the model. Specifically, the final hidden state ($\mH_{T}$) outputted by the first GatedDeltaProduct layer in the stack is passed forward to serve as the initial hidden state ($\mH_{0}$) for the subsequent GatedDeltaProduct layer. This ensures that deeper layers do not start their recurrence from a blank slate but instead \textit{build upon the aggregated temporal state representations} learned by the shallower layers. This process creates a dense flow of information across both the time dimension (via the recurrence) and the model depth (via the weaving).

\paragraph{State Tracking and Its Relevance in Time-Series Forecasting.}
A central advantage of DeltaProduct is its ability to perform \emph{state tracking}, i.e., maintaining and updating information over long sequences. Diagonal linear RNNs and SSMs (e.g., Mamba, RWKV, GLA) update each hidden dimension independently, which is efficient but severely limits expressivity: they cannot mix coordinates, cannot implement basic tracking functions such as parity or counting~\citep{grazzi-iclr25a}, and their states inevitably drift toward zero due to exponential decay. DeltaProduct avoids this failure mode through negative eigenvalues present in the structured non-diagonal transitions, where each factor $(\mI - \beta \vk\vk^\top)$ acts as a reflection or low-rank rotation that preserves geometry and prevents collapse. This capability is crucial for time-series forecasting, where tracking corresponds to maintaining \emph{trend} and \emph{level} information across long contexts. As a result, GatedDeltaProduct layers maintain trend information without attenuation, enabling stable and coherent extrapolation over extended horizons.

\color{black}
\section{Synthetic Data Implementation Details}
\label{app:synthetic_details}

\subsection{Generator Specifications}

\textbf{CauKer.} To increase the diversity and structural complexity of our training data, we used the CauKer generator from~\citep{cauker}. This method produces multivariate time series by sampling from a structural causal model (SCM) where each variable is a Gaussian process. A random Directed Acyclic Graph (DAG) defines the dependencies between nodes, with each node having a maximum number of parents. Root nodes in the DAG are sampled from a GP prior $y_i \sim \mathcal{GP}(m(t), \kappa(t, t'))$, using complex composite kernels $\kappa$ (combined via + or *) and stochastic mean functions $m(t)$ (e.g., linear $at + b$, exponential $a\exp(bt)$, or functions with anomalous impulses). Child nodes then apply nonlinear activation functions (e.g., ReLU, sigmoid, sin) to affine combinations of their parents' values, introducing intricate, non-Gaussian dependencies.

We generated multivariate series with 21 channels and  treated each channel as an independent univariate time series. This approach allows us to efficiently create a vast corpus of realistic, interdependent patterns from a single generative process, providing the model with a rich and varied learning signal that encompasses trends, periodicities, and complex nonlinear interactions.
\begin{figure}[H]
    \centering
    \includegraphics[width=\linewidth]{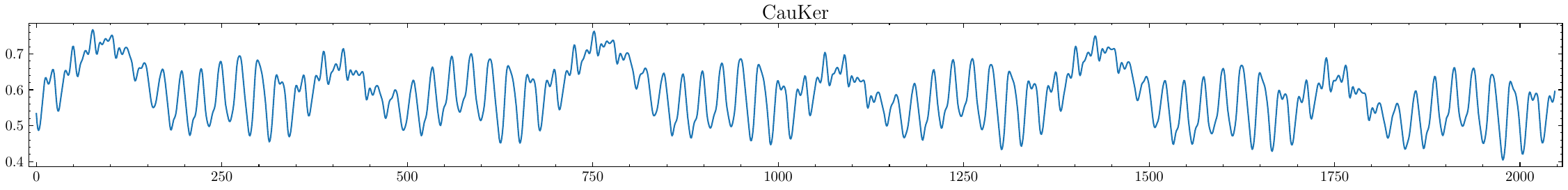}
    \caption{Example time series generated by CauKer}
    \label{fig:cauker}
\end{figure}

\textbf{KernelSynth.} The KernelSynth generator, based on Chronos~\citep{ansari2024chronos}, samples independent univariate time series from Gaussian process priors $y \sim \mathcal{GP}(0, \kappa(t, t'))$. It constructs composite kernels $\kappa$ by randomly combining base kernels (using addition or multiplication) from a large bank. This bank includes periodic kernels (ExpSineSquared($p$) with periods $p$ normalized by series length), stationary kernels (RBF, RationalQuadratic), and noise kernels (WhiteKernel). This method efficiently produces a vast array of smooth and structured series, ideal for learning fundamental temporal representations.

\begin{figure}[H]
    \centering
    \includegraphics[width=\linewidth]{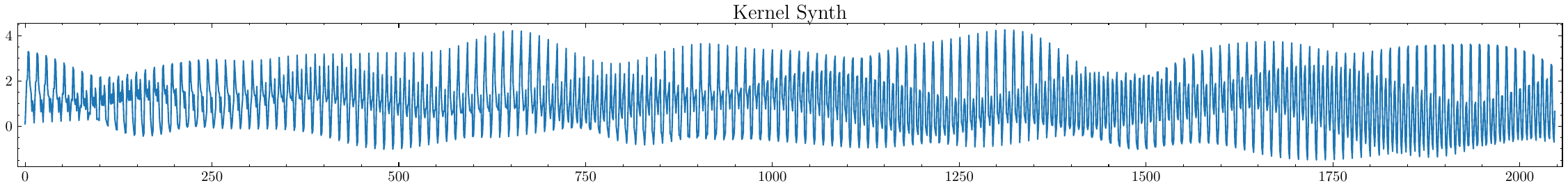}
    \caption{Example time series generated by KernelSynth}
    \label{fig:kernelsynth}
\end{figure}

\textbf{Gaussian Process.} The Gaussian Process generator, inspired by Mamba4Cast~\citep{mamba4cast}, extends the GP sampling approach with greater complexity and realism. It constructs a composite kernel by combining up to six base kernels from a weighted bank that includes Matern, linear, periodic, and polynomial kernels. The combination logic (addition or multiplication) is also chosen randomly. To generate realistic periodicities, the periods of any periodic kernels are sampled from distributions tailored to the time series' specified frequency (e.g., daily, weekly). Crucially, with a certain probability, we inject \textbf{periodic peak spikes} that are aligned with the dominant periodicity of the sampled kernel. This process creates sharp, recurring events on top of the smooth GP trajectory, yielding a wide range of both stationary and non-stationary series with complex covariance structures that mix smooth and abrupt dynamics.

\begin{figure}[H]
    \centering
    \includegraphics[width=\linewidth]{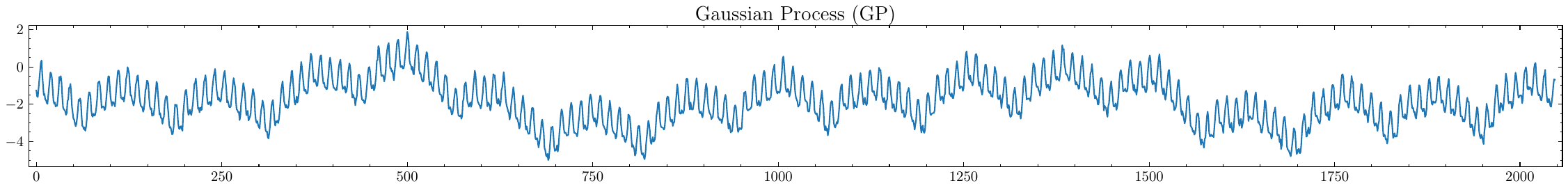}
    \caption{Example time series generated by GP}
    \label{fig:gp}
\end{figure}

\textbf{ForecastPFN.} The ForecastPFN generator, adapted from~\citet{dooley2023forecastpfn}, creates time series with configurable trends, seasonality, and noise patterns. The trend component combines linear and exponential elements multiplicatively for improved stability: 
\begin{equation}
\tau(t) = [b + s_l(t + o_l)] \times s_e^{(t + o_e)},
\end{equation}
where the exponential base $s_e$ is carefully scaled based on series length and frequency to prevent unbounded growth. The seasonality component is also multiplicative: 
\begin{equation}
s(t) = \prod_f \left(1 + s_f \sum_h \left[c_{f,h} \sin\left(\frac{2\pi h(t + o_f)}{p_f}\right) + d_{f,h} \cos\left(\frac{2\pi h(t + o_f)}{p_f}\right)\right]\right).
\end{equation}
The final series values are given by $\tau(t) \cdot s(t) \cdot (1 + n(t))$, where $n(t)$ is Weibull-distributed noise. We enhanced this foundation with a noise injection strategy inspired by~\citet{mamba4cast}, incorporating univariate augmentations like time warping, magnitude scaling, damping, and spike injection. A built-in filtering mechanism with retry logic ensures generated series avoid unrealistic spreads or extreme values, guaranteeing robust training data.
\begin{figure}[H]
    \centering
    \includegraphics[width=\linewidth]{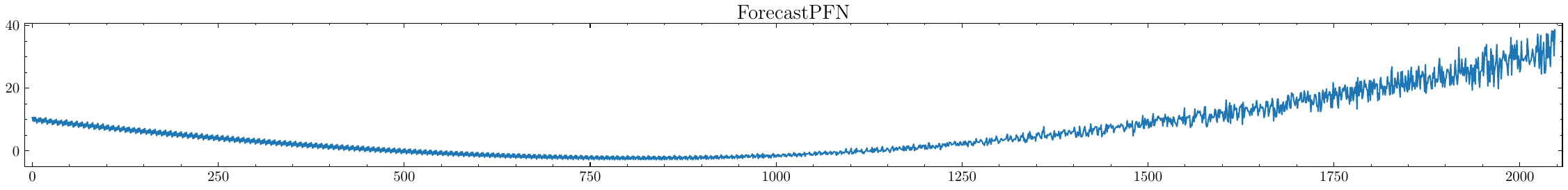}
    \caption{Example time series generated by ForecastPFN}
    \label{fig:forecastpfn}
\end{figure}

\textbf{Sawtooth.} The Sawtooth generator creates univariate series with linear ramping patterns. The core waveform is a sawtooth function: $y_t = A \cdot \text{frac}((t/P) + \phi)$ for upward ramps, or $y_t = A \cdot (1 - \text{frac}((t/P) + \phi))$ for downward ramps (direction chosen randomly). To prevent overly idealised signals, minimal linear trends ($s_l t$) and low-amplitude seasonal components ($a\sin(2\pi t/Q)$) are added. This encourages the model to learn robust representations of trend-dominated series.

\begin{figure}[H]
    \centering
    \includegraphics[width=\linewidth]{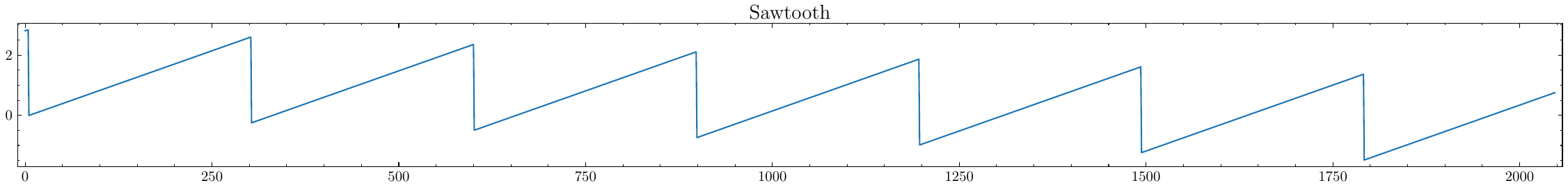}
    \caption{Example time series generated by Sawtooth Generator}
    \label{fig:sawtooth}
\end{figure}

\textbf{Step Function.} Our Step Function generator constructs complex piecewise constant series by concatenating multiple subseries. Each subseries is generated from a configurable distribution of patterns (stable, gradual trends, spikes, oscillations, random walks) with specific lengths, number of changepoints, step sizes, and drift. The combined series undergoes optional Gaussian smoothing at transitions. Finally, global components like noise, seasonality, a linear trend, and point anomalies are added, creating rich and non-stationary step-like data.

\begin{figure}[H]
    \centering
    \includegraphics[width=\linewidth]{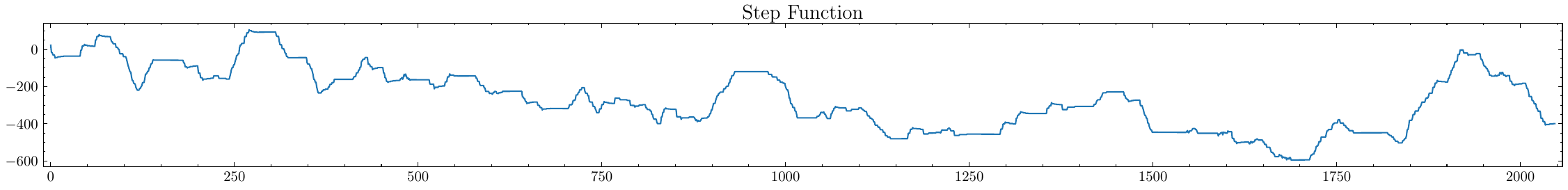}
    \caption{Example time series generated by Step Function}
    \label{fig:step_function}
\end{figure}

\textbf{Anomaly.} The Anomaly generator focuses on outlier detection by producing otherwise constant baseline signals contaminated with periodic spike anomalies. For a given series, all spikes are exclusively positive or negative. Their timing follows patterns (single, clustered, or mixed) with period variance and jitter, while their magnitudes follow defined regimes (constant, trending, cyclical, or correlated random). This provides a controlled environment for learning anomaly detection semantics.

\begin{figure}[H]
    \centering
    \includegraphics[width=\linewidth]{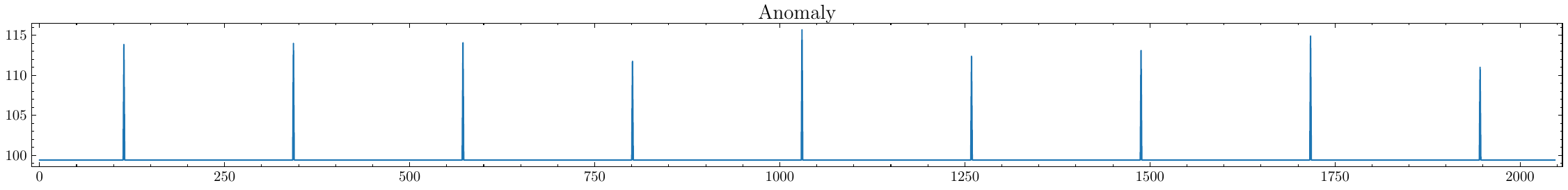}
    \caption{Example time series generated by Anomaly Generator}
    \label{fig:anomaly}
\end{figure}

\textbf{Spikes.} The Spikes generator creates series where the primary feature is the spike itself, defined on a flat baseline. Spikes have consistent per-series direction and shape (V-shaped, inverted-V, or chopped variants with plateaus). They are generated in either "burst" (clustered) or "spread" (evenly spaced with defined edge margins) modes. Colored (brown/pink) noise is added probabilistically. This generator is designed to simulate event-driven signals common in domains like healthcare or intrusion detection.

\begin{figure}[H]
    \centering
    \includegraphics[width=\linewidth]{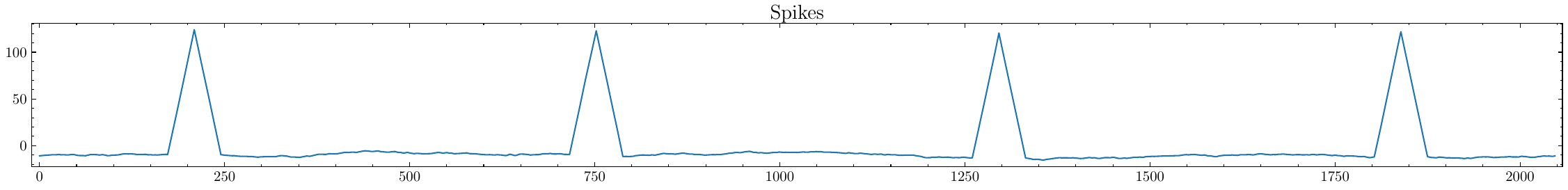}
    \caption{Example time series generated by Spike Generator}
    \label{fig:spikes}
\end{figure}

\textbf{Sine Wave.} Our Sine Wave generator produces complex and non-stationary oscillatory patterns, moving beyond simple periodic signals. It generates a time series by summing 1 to 3 sinusoidal components, each subject to modulation, and then adds a global trend and noise. The underlying model is:
$$ y_t = \sum_{i=1}^{N} A_i(t) \sin\left( \phi_i(t) \right) + (at+b) + \epsilon_t $$
Here, $A_i(t)$ represents a time-varying amplitude and $\phi_i(t)$ is a time-varying phase. This is achieved through slow \textbf{amplitude and frequency modulation}, where the amplitude and instantaneous frequency of each sine wave are themselves modulated by another low-frequency sinusoid. This technique introduces realistic drifts and warping in the periodic patterns, preventing the signal from being perfectly predictable. A final linear trend $(at+b)$ and Gaussian noise $\epsilon_t$ are added to complete the series.

\begin{figure}[H]
    \centering
    \includegraphics[width=\linewidth]{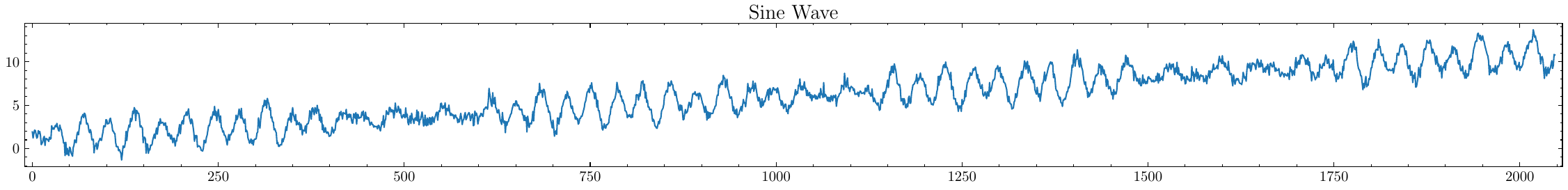}
    \caption{Example time series generated by Sine Wave Generator}
    \label{fig:sinewave}
\end{figure}

\textbf{Audio-Inspired Generators.} To generate exceptionally complex and realistic time series, we introduce a family of four novel generators based on procedural audio synthesis techniques, using the \texttt{pyo} digital signal processing library. An audio synthesis graph is constructed with various oscillators and modulators, rendered offline, and then resampled to the target time series length. This paradigm allows us to model intricate, dynamic systems.

\begin{itemize}
    \item \textbf{Stochastic Rhythm:} This generator creates multi-layered, event-driven patterns. A base tempo is set, and 3 to 5 rhythmic layers are created on top, each triggering at a random subdivision of the tempo (e.g., twice, three times, etc.). Each trigger fires a percussive envelope controlling a sine wave oscillator, resulting in a complex, polyrhythmic signal ideal for modeling data with recurring, patterned events.    

    \item \textbf{Financial Volatility:} This generator mimics financial market dynamics. It combines three components: a slow-moving LFO that acts as the market trend, a Brownian noise source whose amplitude is modulated to create \textit{volatility clustering}, and a triggered, sharp envelope that creates sudden positive or negative \textit{jumps} or shocks.

    \item \textbf{Network Topology:} This generator simulates network traffic data. The signal is a mixture of five components: a base traffic flow (slow LFO), high-frequency noise bursts representing packet traffic, periodic dips from triggered envelopes to model congestion, a high-frequency sine wave for protocol overhead, and large, sharp spikes from filtered noise to simulate DDoS-like attacks.
     
    \item \textbf{Multi-Scale Fractal:} This generator produces self-similar, fractal-like patterns. A Brownian noise source is passed through a bank of 3 to 6 parallel band-pass filters. The center frequencies of these filters are logarithmically spaced, and each successive filter has a higher attenuation. Summing the outputs creates a signal with structure at multiple time scales.
\end{itemize}

\begin{figure}[H]
    \centering      \includegraphics[width=\linewidth]{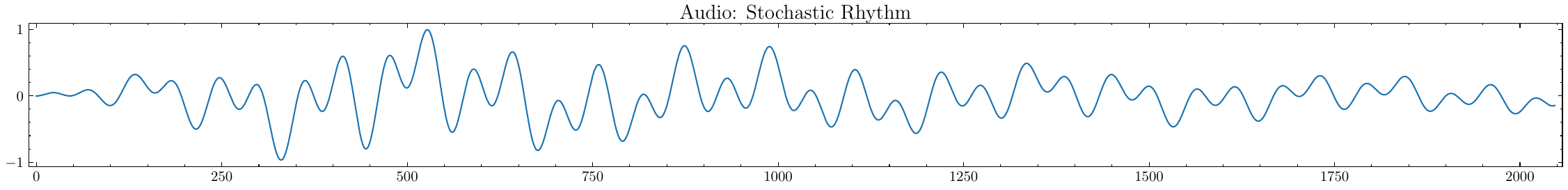}
    \caption{Example time series generated by Audio Stochastic Rhythm}
    \label{fig:audio_stochastic_rhythm}
\end{figure}

\textbf{Stochastic Differential Equations (SDEs).}
SDEs provide a principled framework for modeling continuous-time random processes. An SDE specifies the infinitesimal dynamics of a state variable $y_t$ as
\begin{equation}
dy_t = a(y_t,t) dt + b(y_t,t) dW_t,
\end{equation}
where $a(\cdot,\cdot)$ is the \textit{drift function} governing deterministic trends, $b(\cdot,\cdot)$ is the \textit{diffusion function} controlling random fluctuations, and $W_t$ is a standard Brownian motion. Unlike deterministic differential equations, solutions are random trajectories whose distribution is determined by $(a,b)$ and the distribution of initial conditions.

We adopt the Itô convention of stochastic calculus. This choice is standard in financial mathematics and machine learning because Itô integrals enjoy martingale properties. For simulation, we discretize the SDE on a time grid $\{0,\Delta t, 2\Delta t,\dots,T\}$ using the Euler–Maruyama scheme:
\begin{equation}
y_{t+\Delta t} = y_t + a(y_t,t)\Delta t + b(y_t,t)\sqrt{\Delta t}Z_t,\quad Z_t \sim \mathcal{N}(0,1).
\end{equation}
More advanced schemes such as the Milstein method can reduce bias when the diffusion term depends on $y_t$, but Euler–Maruyama suffices for our purposes.

\begin{figure}[h]
    \centering
    \includegraphics[width=\linewidth]{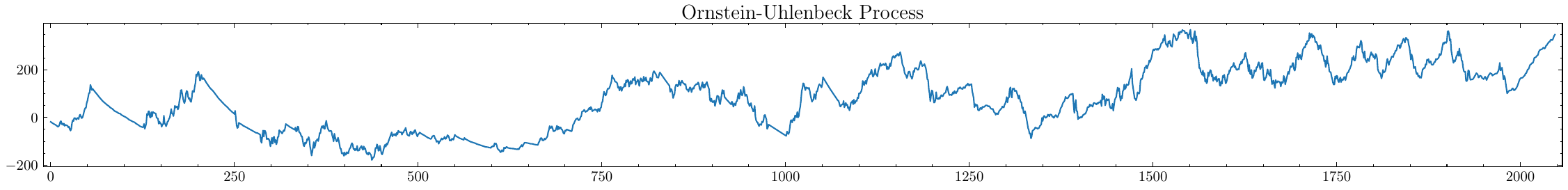}
    \caption{Example time series generated by Ornstein–Uhlenbeck process}
    \label{fig:ornstein_uhlenbeck_process}
\end{figure}

Equation $dy_t = \theta(t,r_t)\,\bigl(\mu(t,r_t) - y_t \bigr)\,dt + \sigma(t,r_t)\, dW_t$,
where $\theta(t,r_t)$ is the mean reversion speed, $\mu(t,r_t)$ the time-varying mean, and $\sigma(t,r_t)$ the volatility, defines the process. In regime $r_t \in \{0,1\}$, the drift and diffusion coefficients are parameterized as
\begin{align}
    \theta(t, r_t) &= \theta^{(r_t)} \cdot (1 + \delta_\theta(t)), \\
    \mu(t, r_t) &= \mu^{(r_t)} + \mu^{\text{trend}}(t) + \mu^{\text{season}}(t), \\
    \sigma(t, r_t) &= \sigma^{(r_t)} \cdot \bigl(1 + \sigma^{\text{trend}}(t) + \sigma^{\text{season}}(t)\bigr),
\end{align}
where $\delta_\theta(t)$, $\mu^{\text{trend}}(t)$, $\sigma^{\text{trend}}(t)$ are smooth trend functions (e.g., linear, logistic, polynomial), and $\mu^{\text{season}}(t)$, $\sigma^{\text{season}}(t)$ are sinusoidal seasonal components with possible amplitude evolution. Regime switching occurs with probabilities $p_{00}, p_{11} \in [0.85,0.999]$. The initial state is drawn from $X_0 \sim \mathcal{N}(\mu^{(r_0)}, \sigma^{(r_0)2})$, with $r_0$ chosen uniformly. Each path is subsequently transformed via a global scaling factor $s \sim U[0.1, 50.0]$, global level shift $\ell \sim U[-100,100]$, and additive Gaussian measurement noise $\epsilon_t \sim \mathcal{N}(0,\sigma_\epsilon^2)$ with $\sigma_\epsilon \in [0,0.1]$. When long memory is enabled, $W_t$ is replaced with fractional Brownian motion $B^H_t$ with Hurst exponent $H \in [0.3,0.8]$. Table~\ref{tab:rsou_params} summarizes the sampling ranges for all parameters used in the generator.

\begin{table}[h]
\centering
\small
\caption{Parameter ranges for the Regime-Switching OU generator.}
\label{tab:rsou_params}
\resizebox{0.75\textwidth}{!}{
\begin{tabular}{ll}
\toprule
\textbf{Parameter} & \textbf{Range / Distribution} \\
\midrule
Integration step size $dt$ & $0.01$ \\
Initial value $y_0$ & $\mathcal{N}(0, 2^2)$ \\
Regime 0 mean reversion $\theta^{(0)}$ & $[1.0, 5.0]$ \\
Regime 0 mean $\mu^{(0)}$ & $\mathcal{N}(-2.0, 1.0^2)$ \\
Regime 0 volatility $\sigma^{(0)}$ & $\log\mathcal{N}(\log 0.3, 0.3)$ \\
Regime 0 vol. process $(\kappa_v, \theta_v, \xi_v)$ & $[2.0, 5.0], [0.2, 0.4], [0.1, 0.3]$ \\
Regime 1 mean reversion $\theta^{(1)}$ & $[0.05, 0.5]$ \\
Regime 1 mean $\mu^{(1)}$ & $\mathcal{N}(2.0, 1.0^2)$ \\
Regime 1 volatility $\sigma^{(1)}$ & $\log\mathcal{N}(\log 1.5, 0.5)$ \\
Regime 1 vol. process $(\kappa_v, \theta_v, \xi_v)$ & $[0.5, 2.0], [0.8, 1.2], [0.3, 0.5]$ \\
Regime transition probs $p_{00}, p_{11}$ & $[0.85, 0.999]$ \\
Global level shift $\ell$ & $[-100.0, 100.0]$ \\
Global scale factor $s$ & $[0.1, 50.0]$ \\
Measurement noise std $\sigma_\epsilon$ & $[0.0, 0.1]$ \\
Hurst exponent $H$ & $[0.3, 0.8]$ \\
Seasonal components & $1$–$3$ harmonics \\
Seasonal periods & $\{7.0, 30.0, 90.0, 182.6, 365.25\}$ \\
Seasonal amplitude & $[0.5, 3.0]$ \\
Seasonal phase shift & $[0, 2\pi]$ \\
Seasonal period jitter & $\pm 5\%$ \\
Seasonal amplitude evolution & $[-0.001, 0.001]$ \\
Trend application probs & $\mu: 0.7,\ \theta: 0.2,\ \sigma: 0.3$ \\
Seasonality application probs & $\mu: 0.6,\ \sigma: 0.3$ \\
\bottomrule
\end{tabular}
}
\end{table}

\subsection{Synthetic Data Generation Throughput}
\label{app:throughput}

In this section, we present the computational efficiency and resource flexibility of our pipeline. Unlike kernel-based methods such as KernelSynth, which can be computationally intensive due to the cubic $O(T^3)$ complexity of Gaussian Processes, our approach enables high-throughput generation as shown in Table~\ref{tab:generation_throughput}.

The benchmarking was conducted on a high-performance system featuring dual AMD EPYC 9334 32-Core Processors (128 threads total) and an NVIDIA L40S GPU. Crucially, \textit{the majority of our synthetic generators run exclusively on the CPU}. The GPU is leveraged primarily for the few neural network-based prior models (e.g. Cauker).

\begin{table}[h]%
\caption{
    Profiling results for synthetic data generation throughput. \small\textbf{N} = Novel prior, \small\textbf{A} = Adapted from open-source.
}
\label{tab:generation_throughput}
\centering
\sisetup{detect-weight=true, detect-family=true}
\setlength{\tabcolsep}{8pt}
\small
\begin{tabular}{
    l %
    c %
    c %
    S[table-format=3.2] %
}
\toprule
\textbf{Generator} & \textbf{Source} & \textbf{Length} & {\textbf{Series / Sec}} \\
\midrule
Cauker & \textbf{A} & 2048 & 0.66 \\
GP & \textbf{A} & 2048 & 7.04 \\
Kernel & \textbf{A} & 2048 & 0.32 \\
ForecastPFN & \textbf{A} & 2048 & 35.49 \\
Sawtooth & \textbf{N} & 2048 & 242.95 \\
Sinewave & \textbf{N} & 2048 & 144.93 \\
Anomaly & \textbf{N} & 2048 & 174.51 \\
Step & \textbf{N} & 2048 & 106.58 \\
Stochastic Rhythm & \textbf{N} & 2048 & 33.46 \\
Spike & \textbf{N} & 2048 & 201.13 \\
SDE (OU Process) & \textbf{N} & 2048 & 13.17 \\
Offline augmentations & \textbf{N} & 2048 & 18.30 \\
\bottomrule
\end{tabular}
\end{table}

\section{Training Details and Hyperparameters}\label{app:training-details}

\textbf{Data Composition and Sampling.} The training corpus consists of approximately 10 million synthetic time series (500k--2M per generator), with batches composed of mixed samples from our generators. We apply higher weights to Cauker and augmented data to promote diversity in the training distribution.

\textbf{Dynamic Structure Construction.} Our training uses dynamic, per-sample construction of time series structures. For each training instance, we first randomly sample a total sequence length from a weighted distribution that favors longer contexts: \{128: 0.05, 256: 0.10, 512: 0.10, 1024: 0.10, 1536: 0.15, 2048: 0.50\}. When length shortening is applied, we use either cutting or subsampling with equal probability (50/50 split). Next, we perform a random history-future split, with forecast horizon lengths sampled from the range specified by the GIFT benchmark. This two-stage sampling creates highly variable training examples that simulate diverse forecasting tasks.

\textbf{Data Augmentation.} We apply several augmentation techniques during training: (1) Scaler augmentation with 0.5 probability, randomly selecting among minmax, median, or mean scalers (excluding the main robust scaler); (2) NaN augmentation that injects realistic missing data patterns into the history based on GIFT-Eval statistics.

\textbf{Training Infrastructure.} Pretraining uses PyTorch with distributed data parallelism (DDP) across 8--16 NVIDIA A100 or H100 GPUs and mixed precision (\texttt{bfloat16}), a requirement for the DeltaProduct implementation (in FLA: \url{https://github.com/fla-org/flash-linear-attention}).

\textbf{Training Protocol.}
For pretraining, we employ the AdamW optimizer~\citep{loshchilov-iclr19a} with a weight decay of 0.01 and batch size of 40. No additional regularization techniques, such as dropout or early stopping, are applied. Pretraining is conducted for 3 million iterations using a cosine annealing learning rate schedule~\citep{loshchilov-iclr17a} with a peak learning rate of $2 \times 10^{-4}$, a warmup ratio of 0.003, and a minimum learning-rate ratio of 0.01.
The model is trained using the quantile regression loss, computed independently for each output token across the set of quantile levels $\mathcal{Q} = \{q_1, q_2, \ldots, q_m\}$. In our experiments, we set $\mathcal{Q} = \{0.1, 0.2, \ldots, 0.9\}$ similarly as in TiRex and TabPFN-TS. The resulting losses are then averaged over all $h$ output tokens in a training sample. Given the true value $y_t$ at time $t$ and its predicted quantile value $\hat{y}^{(q)}_t$ for quantile level $q \in \mathcal{Q}$, the loss is defined as:
\[
L = \frac{1}{|\mathcal{Q}|\, h}
\sum_{t=1}^{h} \sum_{q \in \mathcal{Q}}
\begin{cases}
q\, (y_t - \hat{y}^{(q)}_t), & \text{if } \hat{y}^{(q)}_t \le y_t,\\[6pt]
(1 - q)\, (\hat{y}^{(q)}_t - y_t), & \text{otherwise}.
\end{cases}
\]

\textbf{Architecture Selection.} We ablated deeper models (8--16 layers) and found no consistent architectural winner. We  selected the 10 layer model with embedding dimension of 512.

\begin{table}[ht]
\centering
\caption{Hyperparameters for main \methodname{} model.}
\label{tab:training-hyperparams}
\resizebox{0.99\textwidth}{!}{
\begin{tabular}{l|l|l}
\toprule
\textbf{Category} & \textbf{Parameter} & \textbf{Value} \\
\midrule
\multirow{8}{*}{Model} & Total Parameters & 40M \\
& Embedding size (\texttt{embed\_size}) & 512 \\
& Encoder layers & 10 \\
& Number of heads (\texttt{num\_heads}) & 4 \\
& Encoder attention mode & \texttt{chunk} \\
& Short convolution kernel size & 32 \\
& State weaving & \texttt{True} \\
& Quantiles for loss & [0.1, 0.2, 0.3, 0.4, 0.5, 0.6, 0.7, 0.8, 0.9] \\
\midrule
\multirow{12}{*}{Training} & Total training series & 10M \\
& Max series length & 2048 \\
& Total training iterations & 3M \\
& Batch size (per GPU) & 40 \\
& Gradient accumulation steps & 5 \\
& Effective batch size (per GPU) & 200 \\
& Peak learning rate & $2 \times 10^{-4}$ \\
& LR scheduler & Cosine annealing \\
& Min learning rate ratio & 0.01 \\
& Warmup ratio & 0.003 \\
\midrule
\multirow{6}{*}{Optimization} & Optimizer & AdamW \\
& $\beta_1$ & 0.9 \\
& $\beta_2$ & 0.98 \\
& Weight decay & 0.01 \\
& Adam $\epsilon$ & $1 \times 10^{-6}$ \\
& Gradient clipping & 100.0 \\
\midrule
\multirow{4}{*}{Augmentations} & Length shortening & \texttt{True} (cut/subsample: 50/50) \\
& NaN augmentation & \texttt{True} \\
& Scaler augmentation prob. & 0.5 (minmax/median/mean) \\
& Batch composition & Mixed (proportions favoring augmented/Cauker) \\
\midrule
\multirow{2}{*}{Hardware} & GPUs & 8--16 $\times$ A100/H100 \\
& Precision & \texttt{bfloat16} \\
\bottomrule
\end{tabular}
}
\end{table}

\FloatBarrier

\subsection{Additional Experimental Details and Results}
\label{app:additional_experiments}

The results presented in this section are based on ablation studies conducted with our main model architecture.

\begin{table*}[ht]
\centering
\small
\setlength{\tabcolsep}{3.5pt}
\caption{Ablation study of single synthetic priors (trained for 500k iterations). 'Base Model' uses all priors and augmentations. Lower values are better. \textbf{Bold}: best, \underline{underline}: second-best. Novel priors are our contributions; Adapted are modified open-source versions.}
\label{tab:single_prior_ablations}
\vspace{-5pt} %
\resizebox{0.99\linewidth}{!}{
\begin{tabular}{l|c|cc|cc|cc|cc}
\toprule
& & \multicolumn{2}{c|}{\textbf{Gift-ZS Overall}} & \multicolumn{2}{c|}{\textbf{Gift-ZS Short}} & \multicolumn{2}{c|}{\textbf{Gift-ZS Medium}} & \multicolumn{2}{c}{\textbf{Gift-ZS Long}} \\
\cmidrule(lr){3-4} \cmidrule(lr){5-6} \cmidrule(lr){7-8} \cmidrule(lr){9-10}
\textbf{Ablation} & \textbf{Source} & \textbf{CRPS} & \textbf{MASE} & \textbf{CRPS} & \textbf{MASE} & \textbf{CRPS} & \textbf{MASE} & \textbf{CRPS} & \textbf{MASE} \\
\midrule
Base Model & -- & 0.577 & 0.842 & 0.563 & 0.763 & 0.566 & 0.900 & 0.631 & 1.019 \\
+ Cauker & Adapted & \textbf{0.600} & \textbf{0.875} & \textbf{0.583} & \textbf{0.789} & \textbf{0.615} & \textbf{0.964} & \textbf{0.631} & \textbf{1.043} \\
+ GP & Adapted & \underline{0.632} & \underline{0.897} & \underline{0.607} & \underline{0.812} & \underline{0.666} & \underline{0.993} & \underline{0.666} & \underline{1.053} \\
+ Kernel & Adapted & 0.638 & 0.926 & 0.622 & 0.835 & 0.656 & 1.042 & 0.661 & 1.082 \\
+ ForecastPFN & Adapted & 0.715 & 1.027 & 0.695 & 0.918 & 0.760 & 1.172 & 0.726 & 1.206 \\
+ SDE (OU Process) & Novel & \textbf{0.815} & \textbf{1.148} & \textbf{0.763} & \textbf{1.017} & \textbf{0.897} & \textbf{1.334} & \textbf{0.879} & \textbf{1.354} \\
+ Sinewave & Novel & \underline{0.868} & \underline{1.223} & \underline{0.854} & \underline{1.113} & \underline{0.901} & \underline{1.375} & \underline{0.872} & \underline{1.397} \\
+ Stochastic Rhythm & Novel & 0.953 & 1.337 & 0.940 & 1.252 & 1.004 & 1.472 & 0.938 & 1.440 \\
+ Sawtooth & Novel & 1.187 & 1.534 & 1.162 & 1.362 & 1.294 & 1.802 & 1.152 & 1.781 \\
+ Spike & Novel & 1.215 & 1.318 & 1.019 & 1.250 & 1.565 & 1.411 & 1.498 & 1.416 \\
+ Anomaly & Novel & 1.310 & 1.522 & 1.487 & 1.610 & 1.145 & 1.430 & 1.075 & 1.399 \\
+ Step & Novel & 2.199 & 1.702 & 1.272 & 1.280 & 4.325 & 2.398 & 4.693 & 2.549 \\
\midrule
seasonal\_naive & -- & 1.000 & 1.000 & 1.000 & 1.000 & 1.000 & 1.000 & 1.000 & 1.000 \\
\bottomrule
\end{tabular}
}
\end{table*}

\begin{table*}[ht]
\centering
\small
\setlength{\tabcolsep}{3.5pt}
\caption{Core architectural ablations (trained for 2M iterations with a per-GPU batch size of 40). Base config: $d=512$, $L=10$, conv size 16, $H=4$, weaving enabled, negative eigenvalues allowed. Sorted by overall CRPS. \textbf{Bold}: best, \underline{underline}: second-best.}
\label{tab:architectural_ablations_v2}
\vspace{-5pt} %
\resizebox{0.99\linewidth}{!}{
\begin{tabular}{l|cc|cc|cc|cc}
\toprule
& \multicolumn{2}{c|}{\textbf{Gift-ZS Overall}} & \multicolumn{2}{c|}{\textbf{Gift-ZS Short}} & \multicolumn{2}{c|}{\textbf{Gift-ZS Medium}} & \multicolumn{2}{c}{\textbf{Gift-ZS Long}} \\
\cmidrule(lr){2-3} \cmidrule(lr){4-5} \cmidrule(lr){6-7} \cmidrule(lr){8-9}
\textbf{Configuration} & \textbf{CRPS} & \textbf{MASE} & \textbf{CRPS} & \textbf{MASE} & \textbf{CRPS} & \textbf{MASE} & \textbf{CRPS} & \textbf{MASE} \\
\midrule
\multicolumn{9}{l}{\textit{Ablating Positional Encoding:}} \\
Base Model (Sin. Pos. Enc. Off) & \textbf{0.561} & \textbf{0.820} & \textbf{0.553} & \textbf{0.751} & \textbf{0.556} & \textbf{0.880} & \textbf{0.590} & \textbf{0.962} \\
Sinusoidal Positional Encoding & \underline{0.648} & \underline{0.937} & \underline{0.596} & \underline{0.809} & \underline{0.731} & \underline{1.120} & \underline{0.715} & \underline{1.152} \\
\midrule
\multicolumn{9}{l}{\textit{Ablating Number of Householder Matrices (H):}} \\
H=6 & \textbf{0.556} & \underline{0.823} & \textbf{0.545} & \textbf{0.750} & \textbf{0.552} & \underline{0.886} & \textbf{0.590} & \underline{0.972} \\
Base Model (H=4) & \underline{0.561} & \textbf{0.820} & \underline{0.553} & \textbf{0.751} & \underline{0.556} & \textbf{0.880} & \textbf{0.590} & \textbf{0.962} \\
H=2 & 0.562 & 0.822 & 0.549 & 0.750 & 0.560 & 0.892 & 0.598 & 0.964 \\
H=1 (DeltaNet equivalent) & 0.573 & 0.845 & 0.556 & 0.761 & 0.579 & 0.918 & 0.613 & 1.020 \\
\midrule
\multicolumn{9}{l}{\textit{Ablating Negative Eigenvalues and Weaving:}} \\
Neg. Eig. Off, Weaving On & \textbf{0.559} & \underline{0.821} & \textbf{0.553} & \underline{0.753} & \textbf{0.550} & \underline{0.881} & \textbf{0.584} & \underline{0.957} \\
Neg. Eig. Off, Weaving Off & \underline{0.560} & \textbf{0.818} & \underline{0.554} & \textbf{0.750} & \underline{0.548} & \textbf{0.879} & \underline{0.590} & \textbf{0.955} \\
Base Model (Neg. Eig. On, Weaving On) & 0.561 & 0.820 & 0.553 & 0.751 & 0.556 & 0.880 & 0.590 & 0.962 \\
\midrule
\multicolumn{9}{l}{\textit{Ablating Convolution Size:}} \\
Conv. size 32 & \textbf{0.559} & \textbf{0.816} & \textbf{0.543} & \textbf{0.737} & \underline{0.566} & \underline{0.897} & \underline{0.594} & \underline{0.968} \\
Base Model (Conv. size 16) & \underline{0.561} & \underline{0.820} & \underline{0.553} & \underline{0.751} & \textbf{0.556} & \textbf{0.880} & \textbf{0.590} & \textbf{0.962} \\
\midrule
seasonal\_naive & 1.000 & 1.000 & 1.000 & 1.000 & 1.000 & 1.000 & 1.000 & 1.000 \\
\bottomrule
\end{tabular}
}
\end{table*}

\begin{table*}[ht]
\centering
\small
\setlength{\tabcolsep}{3.5pt}
\caption{Ablation of model scale and depth (trained for 3M iterations with a batch size of 40). Base Model: $d=512, L=10$, $H=4$, conv size 32, weaving/neg eigenvalues on. Compares width vs. depth at constant parameter count. Sorted by overall CRPS. \textbf{Bold}: best, \underline{underline}: second-best.}
\label{tab:scale_depth_ablations}
\vspace{-5pt} %
\resizebox{0.99\linewidth}{!}{
\begin{tabular}{l|cc|cc|cc|cc}
\toprule
& \multicolumn{2}{c|}{\textbf{Gift-ZS Overall}} & \multicolumn{2}{c|}{\textbf{Gift-ZS Short}} & \multicolumn{2}{c|}{\textbf{Gift-ZS Medium}} & \multicolumn{2}{c}{\textbf{Gift-ZS Long}} \\
\cmidrule(lr){2-3} \cmidrule(lr){4-5} \cmidrule(lr){6-7} \cmidrule(lr){8-9}
\textbf{Configuration} & \textbf{CRPS} & \textbf{MASE} & \textbf{CRPS} & \textbf{MASE} & \textbf{CRPS} & \textbf{MASE} & \textbf{CRPS} & \textbf{MASE} \\
\midrule
Base Model (d=512, L=10) & \textbf{0.533} & \textbf{0.788} & \textbf{0.532} & \underline{0.727} & \textbf{0.523} & \textbf{0.840} & \textbf{0.544} & \textbf{0.912} \\
d=512, L=10, Weaving Off & \underline{0.537} & \underline{0.790} & \textbf{0.532} & \textbf{0.723} & \underline{0.533} & \underline{0.862} & \underline{0.553} & \underline{0.914} \\
d=384, L=16 (Narrower, Deeper) & 0.539 & 0.792 & 0.532 & 0.727 & 0.533 & 0.850 & 0.563 & 0.921 \\
d=576, L=8 (Wider, Shallower) & 0.540 & 0.794 & 0.536 & 0.732 & 0.529 & 0.849 & 0.561 & 0.921 \\
\midrule
seasonal\_naive & 1.000 & 1.000 & 1.000 & 1.000 & 1.000 & 1.000 & 1.000 & 1.000 \\
\bottomrule
\end{tabular}
}
\end{table*}

\begin{table*}[ht]
\centering
\small
\setlength{\tabcolsep}{3.5pt}
\caption{LR scheduler ablation (trained for 2M iterations with a batch size of 40). Base architecture: $d=512, L=10$, $H=4$, conv size 32, weaving enabled. \textit{WarmupStableDecay}: warmup (0.3\%), plateau (90\%), cosine decay (9.7\%). \textit{CosineWithRestarts}: 4 resets. Sorted by overall CRPS. \textbf{Bold}: best, \underline{underline}: second-best.}
\label{tab:scheduler_ablations}
\vspace{-5pt} %
\resizebox{0.99\linewidth}{!}{
\begin{tabular}{l|cc|cc|cc|cc}
\toprule
& \multicolumn{2}{c|}{\textbf{Gift-ZS Overall}} & \multicolumn{2}{c|}{\textbf{Gift-ZS Short}} & \multicolumn{2}{c|}{\textbf{Gift-ZS Medium}} & \multicolumn{2}{c}{\textbf{Gift-ZS Long}} \\
\cmidrule(lr){2-3} \cmidrule(lr){4-5} \cmidrule(lr){6-7} \cmidrule(lr){8-9}
\textbf{LR Scheduler} & \textbf{CRPS} & \textbf{MASE} & \textbf{CRPS} & \textbf{MASE} & \textbf{CRPS} & \textbf{MASE} & \textbf{CRPS} & \textbf{MASE} \\
\midrule
WarmupStableDecay & \textbf{0.554} & \textbf{0.812} & \textbf{0.544} & \textbf{0.740} & \textbf{0.550} & \underline{0.877} & \textbf{0.584} & \underline{0.956} \\
CosineWithWarmup & \underline{0.559} & \underline{0.817} & \underline{0.552} & \underline{0.751} & \textbf{0.550} & \textbf{0.873} & \underline{0.588} & \textbf{0.955} \\
CosineWithRestarts & \underline{0.559} & 0.820 & \underline{0.552} & 0.755 & 0.552 & 0.874 & 0.585 & 0.953 \\
Cosine (no warmup) & 0.561 & 0.820 & 0.553 & 0.751 & 0.556 & 0.880 & 0.590 & 0.962 \\
\midrule
seasonal\_naive & 1.000 & 1.000 & 1.000 & 1.000 & 1.000 & 1.000 & 1.000 & 1.000 \\
\bottomrule
\end{tabular}
}
\end{table*}

\begin{figure}
    \vspace{-5mm}
    \centering
    \includegraphics[width=0.4\linewidth, trim={0 0 0 0}, clip]{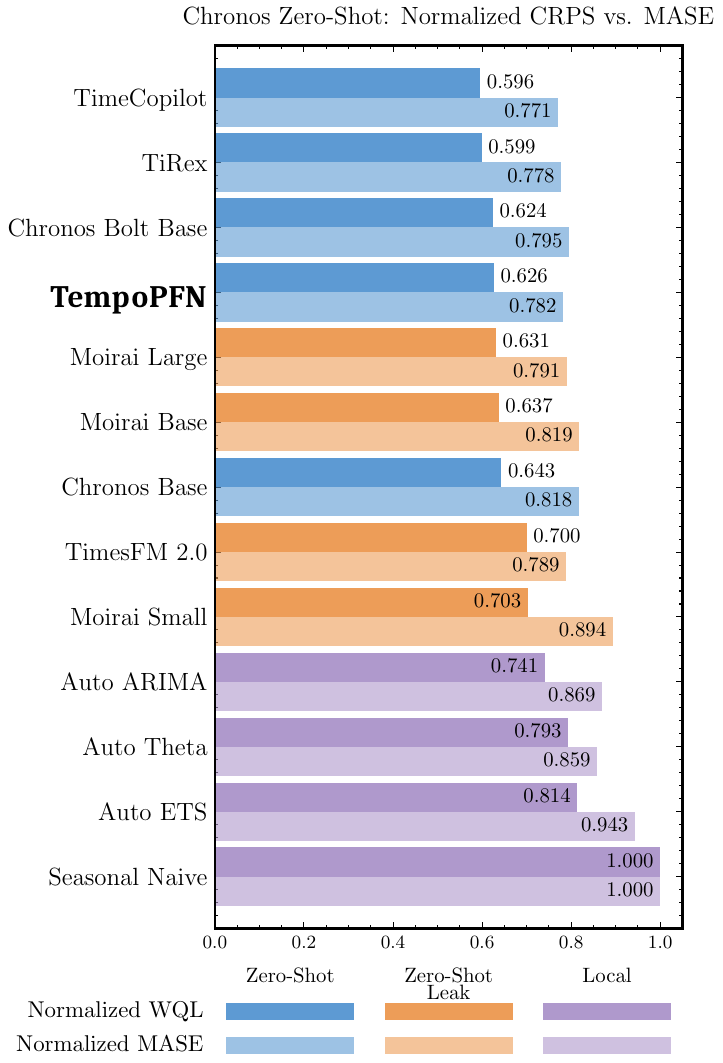}  
    \caption{Comparison of \methodname{} normalized CRPS and MASE (3M iterations with a batch size of 40) against other models on Chronos-ZS benchmark.}
    \label{fig:chronos_zs_scores}
    \vspace{5mm}
\end{figure}
\section{Comprehensive Quantitative Analysis}\label{app:comprehensive_analysis}

\subsection{Computational Complexity and Efficiency}\label{app:complexity_analysis}

Table~\ref{tab:model_comparison} summarizes the computational characteristics of \methodname{} relative to leading time-series foundation models, given sequence length $T$, horizon $H$, embedding dimension $d$, and layers $L$.

\textbf{Training Complexity.}
Transformer-based models (Chronos, TimesFM, MOIRAI, TabPFN-TS) require $\mathcal{O}(T^{2} d)$ compute and memory due to self-attention, which becomes prohibitive for long context windows. TiRex reduces quadratic memory growth but remains sequential along~$T$. In contrast, \methodname{} employs an associative GatedDeltaProduct recurrence, allowing parallel prefix-scan evaluation. This yields linear total work $\mathcal{O}(T L d^{2})$ and logarithmic parallel depth $\mathcal{O}(L \log T)$, enabling full sequence-length parallelism.

\textbf{Inference Latency.}
Autoregressive models such as TiRex and Chronos must unroll $H$ steps to predict a horizon $H$, yielding $\mathcal{O}(H)$ latency. Transformer encoder models also scale their inference cost with $T^{2}$ even when used non-autoregressively. \methodname{} performs \emph{direct forecasting}: concatenated query tokens allow the entire horizon to be produced in a single forward pass, giving constant $\mathcal{O}(1)$ latency with respect to~$H$.
\begin{table}[H]
\centering
\caption{Comparison of time-series foundation models. Complexities are reported with respect to sequence length $T$, horizon $H$, embedding dimension $d$, and layers $L$. We distinguish total work from parallel depth. \methodname{} benefits from scan-accelerated linear recurrences enabling sublinear depth.}
\label{tab:model_comparison}
\resizebox{\textwidth}{!}{
\begin{tabular}{lccccc}
\toprule
\textbf{Model} &
\textbf{Params (M)} &
\textbf{Training Time} &
\textbf{Inference Time} &
\textbf{Memory} &
\textbf{Parallelization} \\
\midrule

TiRex~\citep{auer2025tirex} &
35 &
\begin{tabular}{@{}c@{}}Work: $\mathcal{O}(T L d^{2})$ \\ Depth: $\mathcal{O}(L\,T/w)$ (window $w{=}32$)\end{tabular} &
\begin{tabular}{@{}c@{}}AR: $\mathcal{O}(H L d^{2})$ \\ Depth: $\mathcal{O}(L\,T/w)$\end{tabular} &
$\mathcal{O}(d)$ (streaming) &
Low (sequential sLSTM) \\

TabPFN-TS~\citep{hoo2024the} &
11 &
$\mathcal{O}(T^{2} d)$ &
$\mathcal{O}(T^{2} d)$ &
$\mathcal{O}(T d)$ &
Moderate (attention) \\

TimesFM-2.0~\citep{das2024decoder} &
500 &
$\mathcal{O}(T^{2} d)$ &
\begin{tabular}{@{}c@{}}Direct: $\mathcal{O}(T^{2} d)$\end{tabular} &
$\mathcal{O}(T d)$ &
High (transformer) \\

Chronos~\citep{ansari2024chronos} &
9--205 &
$\mathcal{O}(T^{2} d)$ &
\begin{tabular}{@{}c@{}}AR: $\mathcal{O}(H)$\end{tabular} &
$\mathcal{O}(T d)$ &
High (transformer) \\

MOIRAI-MoE~\citep{liu2024moirai} &
14--935 &
$\mathcal{O}(T^{2} d)$ &
$\mathcal{O}(T^{2} d)$ &
$\mathcal{O}(T d)$ &
High (MoE + transformer) \\

\midrule
\textbf{\methodname{} (ours)} &
40 &
\begin{tabular}{@{}c@{}}Work: $\mathcal{O}(T L d^{2})$ \\ Depth: $\mathcal{O}(L \log T)$\end{tabular} &
\begin{tabular}{@{}c@{}}Work: $\mathcal{O}(T L d^{2})$ \\ Depth: $\mathcal{O}(L \log T)$ \\ Horizon: $\mathcal{O}(1)$\end{tabular} &
\begin{tabular}{@{}c@{}}$\mathcal{O}(d)$ (streaming) \\ or $\mathcal{O}(T d)$ (cached)\end{tabular} &
\textbf{Full (sequence-parallel)} \\
\bottomrule
\end{tabular}}
\end{table}

\textbf{Memory Usage.}
Transformer-based models require caching Key-Value pairs with $\mathcal{O}(T d)$ memory and, in some implementations, up to $\mathcal{O}(T^{2})$ activations. TiRex maintains a hidden state of size $\mathcal{O}(T d)$ during training. \methodname{}, being a Linear RNN, compresses the entire past into a single hidden state and supports streaming inference with constant $\mathcal{O}(d)$ memory, while still allowing optional $\mathcal{O}(T d)$ state storage if needed for analysis or hybrid decoding.
\textbf{Parallelization.}
Transformer-based models benefit from substantial batch-level parallelism but cannot eliminate the quadratic attention bottleneck. TiRex processes its windowed token sequence sequentially in the sLSTM recurrence. \methodname{} achieves \emph{full} sequence-level parallelization: the entire recurrence is computed via parallel scans, providing both high throughput and sublinear parallel depth.
Overall, \methodname{} combines linear training cost, logarithmic parallel depth, constant-latency forecasting, and streaming memory usage, therefore, providing a zero-shot foundation model tailored for long-context forecasting settings.

\subsection{Additional Results on Chronos-ZS and \texttt{fev-bench}}
\label{app:chronos_fevbench_results}
\begin{figure}
    \vspace{-13mm}
    \centering
    \includegraphics[width=0.4\linewidth, trim={0 0 0 0}, clip]{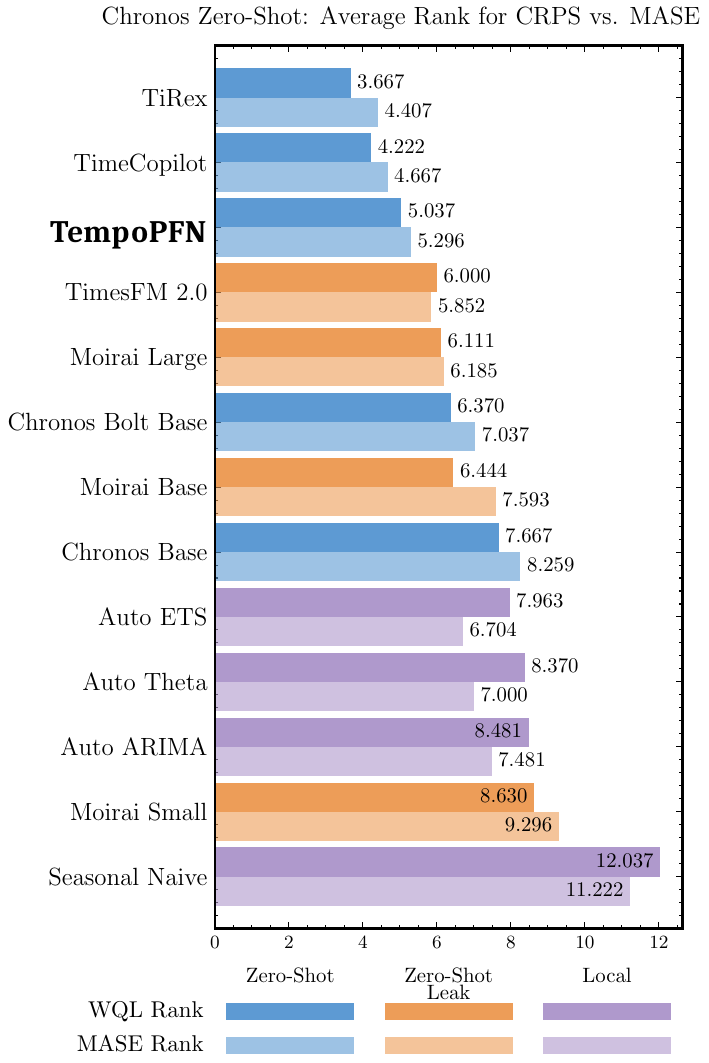}  
    \caption{CRPS and MASE average rank of TempoPFN (3M iterations with a per-GPU batch size of 40) on Chronos-ZS benchmark.}
    \label{fig:chronos_zs_ranks}
\end{figure}

We evaluate \methodname{} on the Chronos-ZS benchmark~\citep{ansari2024chronos}, comprising 27 diverse datasets from the GluonTS~\citep{gluonts} and Monash~\citep{godahewa-neuripsdbt21a} repositories, spanning multiple domains (e.g., energy, transport, healthcare) and frequencies. Figure~\ref{fig:chronos_zs_ranks} shows the aggregated performance in terms of average rank for both probabilistic (CRPS) and point (MASE) forecasting, where TempoPFN ranks again among the top-performing zero-shot models. In Figure~\ref{fig:chronos_zs_scores} in the appendix, we also show the normalized CRPS and MASE scores on this benchmark.

Figure~\ref{fig:chronos_zs_scores} shows the aggregated performance in terms of normalized probabilistic (CRPS) and point (MASE) forecasting, complementing Figure~\ref{fig:chronos_zs_ranks}.

\begin{figure}[t]
    \centering
    \begin{subfigure}[b]{0.49\textwidth}
        \centering
        \includegraphics[width=\linewidth]{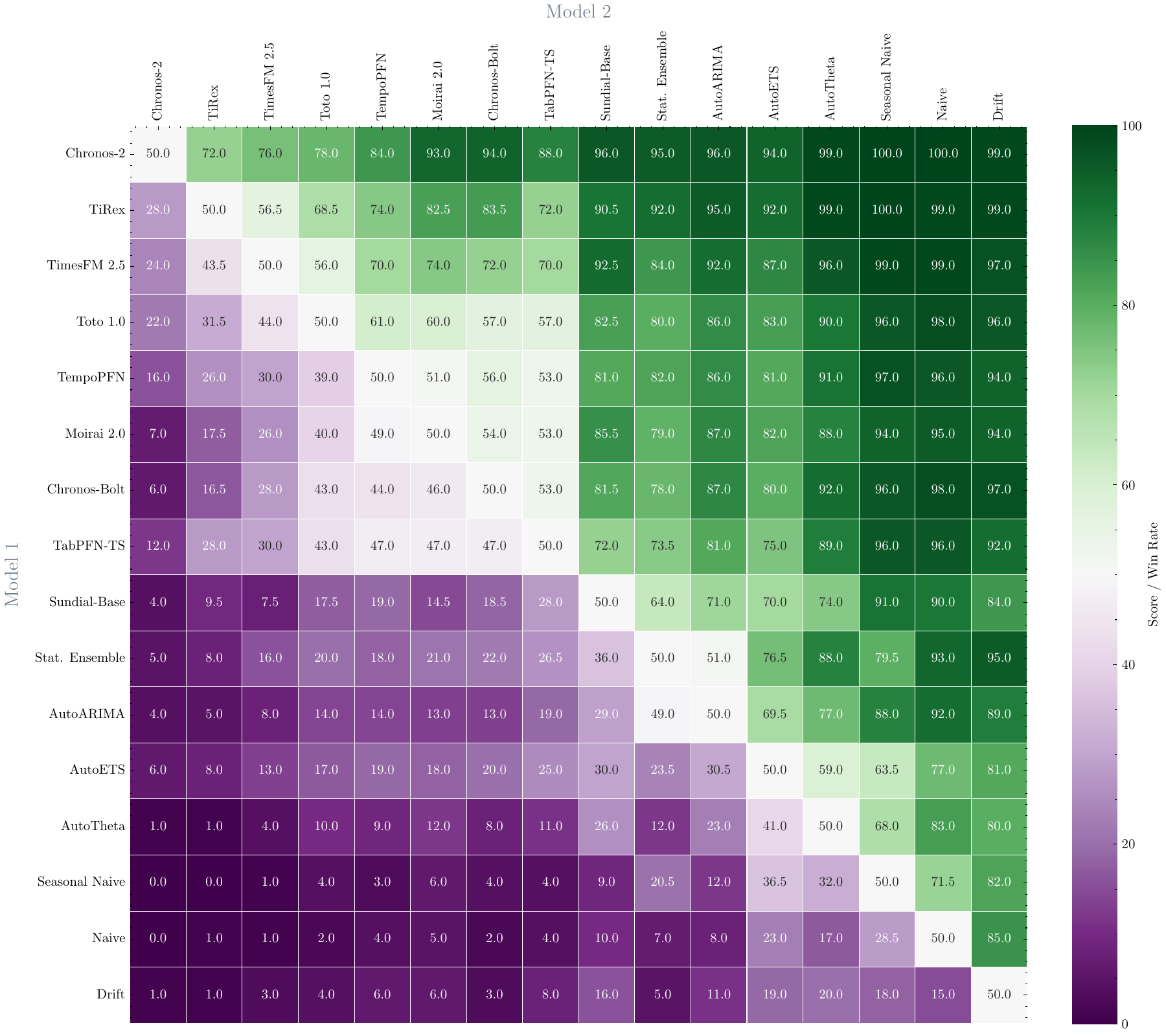}
        \caption{Pairwise Win Rate (SQL) with 95\% CIs}
        \label{fig:pairwise_win_rate_sql}
    \end{subfigure}
    \hfill
    \begin{subfigure}[b]{0.49\textwidth}
        \centering
        \includegraphics[width=\linewidth]{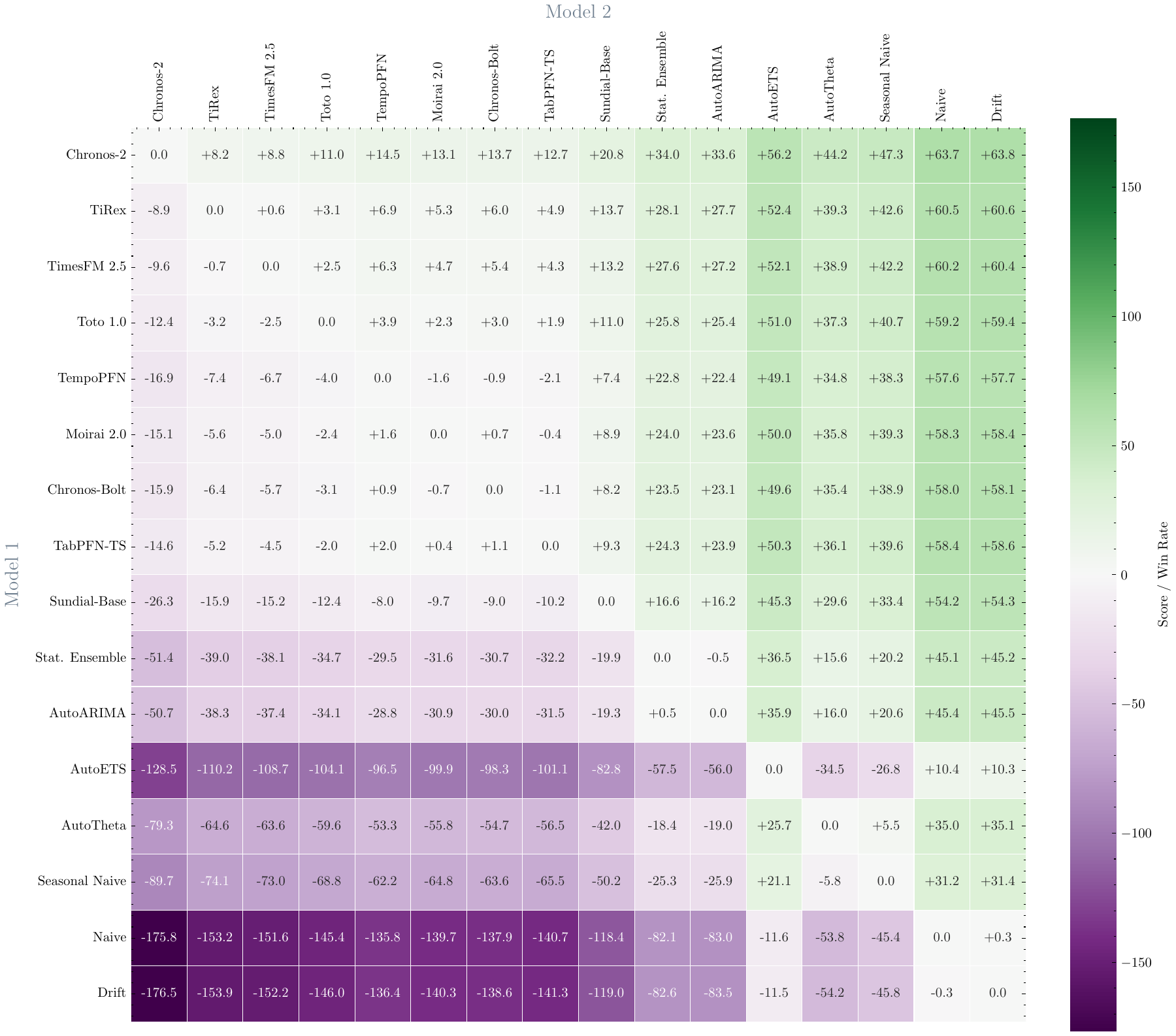}
        \caption{Pairwise Skill Score (SQL) with 95\% CIs}
        \label{fig:pairwise_skill_score_sql}
    \end{subfigure}
    \caption{Head-to-head comparisons on FEV-Bench using Scaled Quantile Loss (SQL). 
    \textbf{(a) Win Rate:} Percentage of tasks where the row model achieves lower error than the column model (ties count as half-wins); values $>$50\% indicate the row model is more accurate on average. 
    \textbf{(b) Skill Score:} Average relative error reduction of the row model with respect to the column model; positive values indicate error reduction. 
    Brackets indicate 95\% confidence intervals estimated via 1000 bootstrap samples.}
    \label{fig:pairwise_heatmaps_sql}
\end{figure}

On \texttt{fev-bench}, in addition to MASE (Table~\ref{tab:fev_bench_results_mase} in the main paper), we also present the leaderboard results for the \textit{Scaled Quantile Loss} (SQL) in Table~\ref{tab:fev_bench_results_sql}. SQL captures calibration quality by evaluating the quality of the entire predictive distribution at each time step. We can see that \methodname{} achieves \textit{Rank 5} again. In both metrics, \methodname{} achieves a higher win rate and rank than the other leading synthetic-only baseline, TabPFN-TS (Rank 8 in both), while TabPFN-TS retains a slightly higher skill score at a roughly 15$\times$ higher median runtime. Finally, to also visualize relative strengths in probabilistic forecasting, we show in Figure~\ref{fig:pairwise_heatmaps_sql} the head-to-head Win Rates and Skill Scores based on SQL.

\begin{table}[t]
\centering
\caption{\texttt{fev-bench} leaderboard based on Scaled Quantile Loss (SQL). Models are ranked by Win Rate and Skill Score. \methodname{} is evaluated with a sign-flip and context-length ensemble at a maximum context of 3072.}
\label{tab:fev_bench_results_sql}
\resizebox{\linewidth}{!}{%
\begin{tabular}{llccccccc}
\toprule
Rank & Model & Avg. Win Rate (\%) & Skill Score (\%) & Median Runtime (s) & Leakage (\%) & Failed Tasks (\%) & Organization & Zero-shot \\
\midrule
1 & Chronos-2 & 90.9 & 47.3 & 2.69 & 0 & 0 & AWS & \cmark \\
2 & TiRex & 82.1 & 42.6 & 1.40 & 1 & 0 & NX-AI & \cmark \\
3 & TimesFM 2.5 & 77.1 & 42.2 & 16.92 & 10 & 0 & Google & \cmark \\
4 & Toto 1.0 & 69.6 & 40.7 & 90.74 & 8 & 0 & Datadog & \cmark \\
\textbf{5} & \textbf{TempoPFN} & \textbf{65.3} & \textbf{38.3} & \textbf{20.67} & \textbf{0} & \textbf{0} & \textbf{Uni Freiburg / ELLIS} & \textbf{\cmark} \\
6 & Moirai 2.0 & 63.4 & 39.3 & 2.46 & 28 & 0 & Salesforce & \cmark \\
7 & Chronos-Bolt & 63.1 & 38.9 & 1.04 & 0 & 0 & AWS & \cmark \\
8 & TabPFN-TS & 61.9 & 39.6 & 305.51 & 0 & 2 & Prior Labs & \cmark \\
9 & Sundial-Base & 44.2 & 33.4 & 35.62 & 1 & 0 & Tsinghua University & \cmark \\
10 & Stat. Ensemble & 43.7 & 20.2 & 690.55 & 0 & 11 & — & \xmark \\
11 & AutoARIMA & 38.9 & 20.6 & 186.84 & 0 & 10 & — & \xmark \\
12 & AutoETS & 32.7 & -26.8 & 16.96 & 0 & 3 & — & \xmark \\
13 & AutoTheta & 25.9 & 5.5 & 9.27 & 0 & 0 & — & \xmark \\
14 & Seasonal Naive & 19.0 & 0.0 & 2.32 & 0 & 0 & — & \xmark \\
15 & Naive & 13.2 & -45.4 & 2.24 & 0 & 0 & — & \xmark \\
16 & Drift & 9.1 & -45.8 & 2.19 & 0 & 0 & — & \xmark \\
\bottomrule
\end{tabular}
}
\end{table}

\subsection{Feature-Space Alignment of Real and Synthetic Data Manifold}
\label{app:real_vs_synth_dist}

To empirically validate that our synthetic pre-training corpus effectively spans the manifold of real-world time series dynamics, we conducted a feature-space analysis comparing our synthetic data against the real-world benchmarks used for evaluation (GIFT-Eval, FEV-Bench, and Chronos).

\textbf{Methodology.} We randomly sampled up to 10,000 time series from each of our synthetic generators and the real-world datasets. For each series, we extracted a comprehensive vector of statistical time-series characteristics (including autocorrelation, approximate entropy, trend strength, spikiness, and seasonality metrics) using the \texttt{tsfresh} library~\citep{christ2018time}. To visualize the relationship between these distributions, we standardized the feature vectors and projected them into a latent space using Uniform Manifold Approximation and Projection (UMAP)~\citep{mcinnes2018umap}.

\begin{figure}[ht]
    \centering
    \pdfcompresslevel=0

    \begin{subfigure}[b]{0.75\linewidth}
        \centering
        \includegraphics[width=\linewidth]{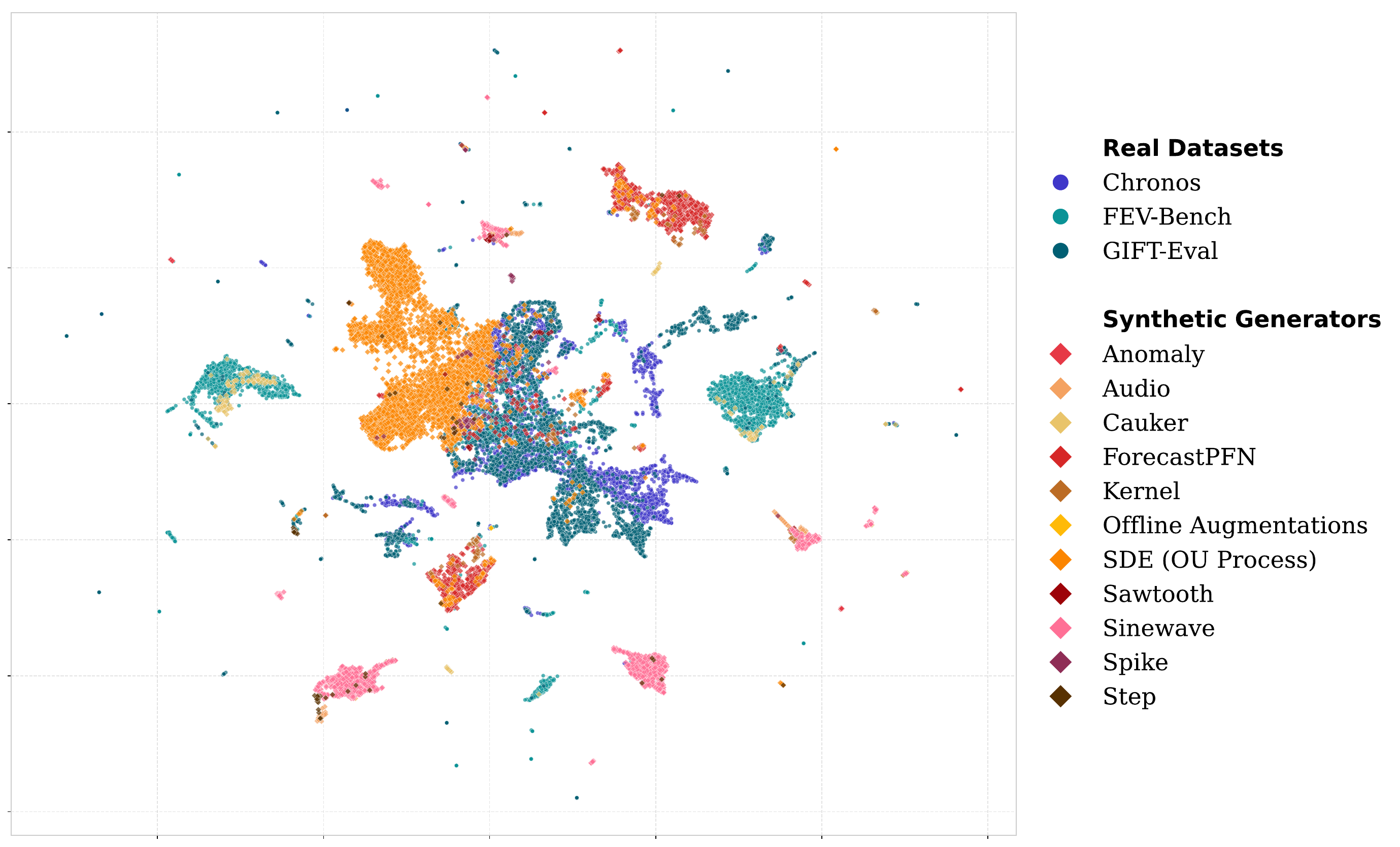}
        \caption{2D Projection ($N_{neigh}=5$, $d_{min}=0.1$, Silhouette Score: -0.054)}
        \label{fig:umap_2d}
    \end{subfigure}\\ \vspace{5mm}
    \begin{subfigure}[b]{0.75\linewidth}
        \centering
        \includegraphics[width=\linewidth]{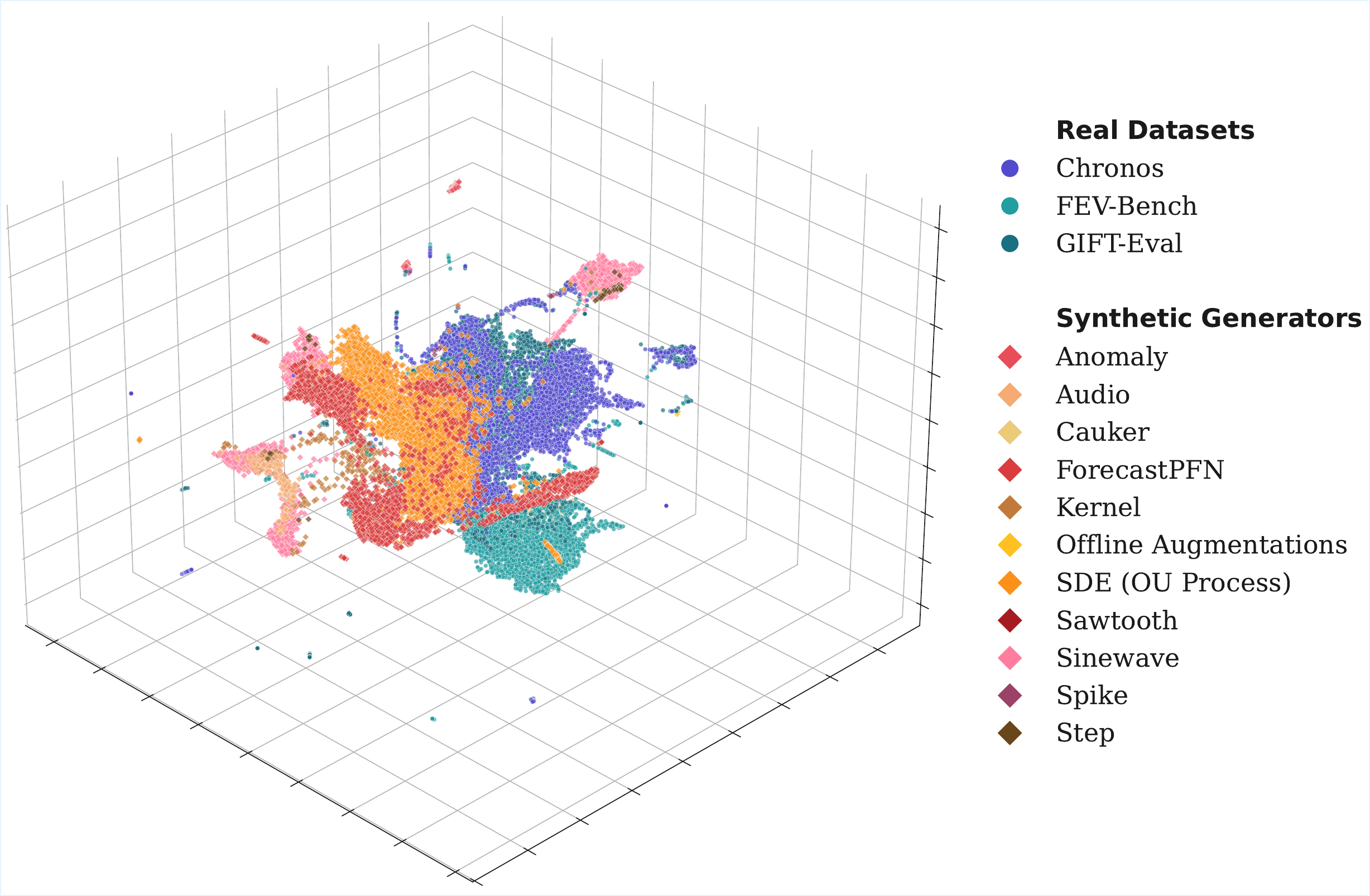}
        \caption{3D Projection ($N_{neigh}=5$, $d_{min}=0.4$, Silhouette Score: -0.001)}
        \label{fig:umap_3d}
    \end{subfigure}
    
    \caption{\textbf{Feature-Space Distribution of Real vs. Synthetic Data.} UMAP projections of time-series features extracted via \texttt{tsfresh}. Real-world benchmarks (GIFT-Eval, FEV-Bench, Chronos) are shown in cold colors (Blue/Teal) in the foreground, while our synthetic generators are shown in warm colors (Red/Orange/Yellow) in the background.}
    \label{fig:umap_distribution}
\end{figure}

\textbf{Analysis.} Figure~\ref{fig:umap_distribution} presents the resulting embeddings in both 2D and 3D projections. The fact that real-world clusters are visible directly \textit{on top} of the synthetic data confirms significant distributional overlap. The synthetic generators do not collapse into a single mode, but instead cover a vast region of the feature space, effectively ``underpainting'' the real-world benchmarks. This visual evidence supports our hypothesis that a diverse mixture of structurally distinct generators and data augmentations collectively covers the complex distribution of real-world temporal dynamics, enabling robust zero-shot transfer.

\newpage
\clearpage

\subsection{Qualitative Comparison on the Gift-Eval Benchmark}
\label{app:qualitative_results}

\begin{figure}[H]
    \centering
    \includegraphics[width=\linewidth]{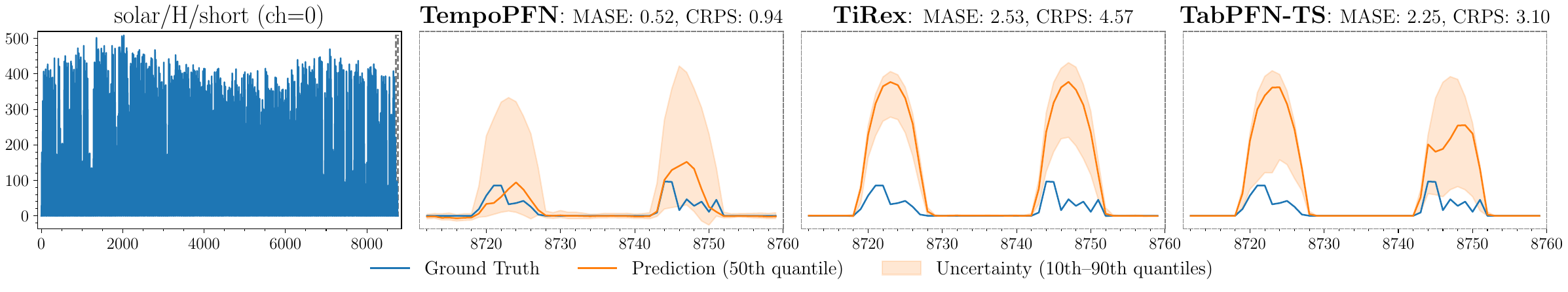}
    \includegraphics[width=\linewidth]{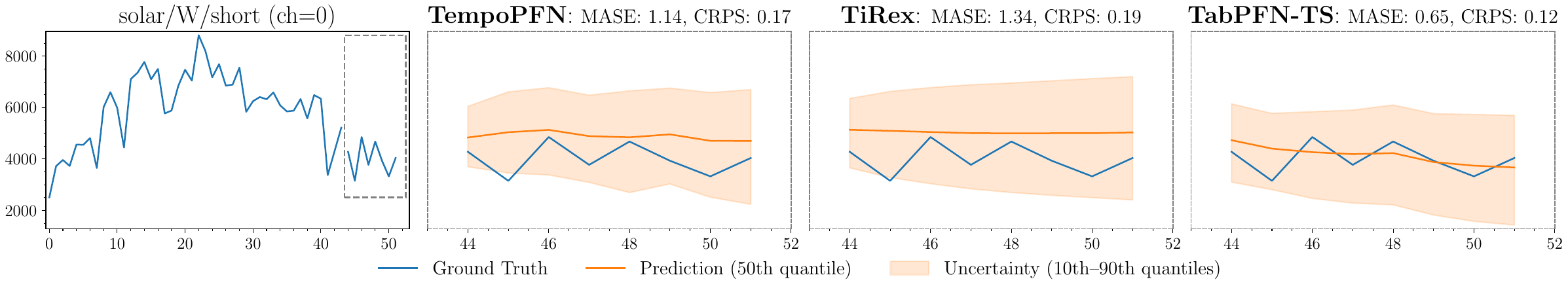}
    \includegraphics[width=\linewidth, trim={0 23 0 0}, clip=true]{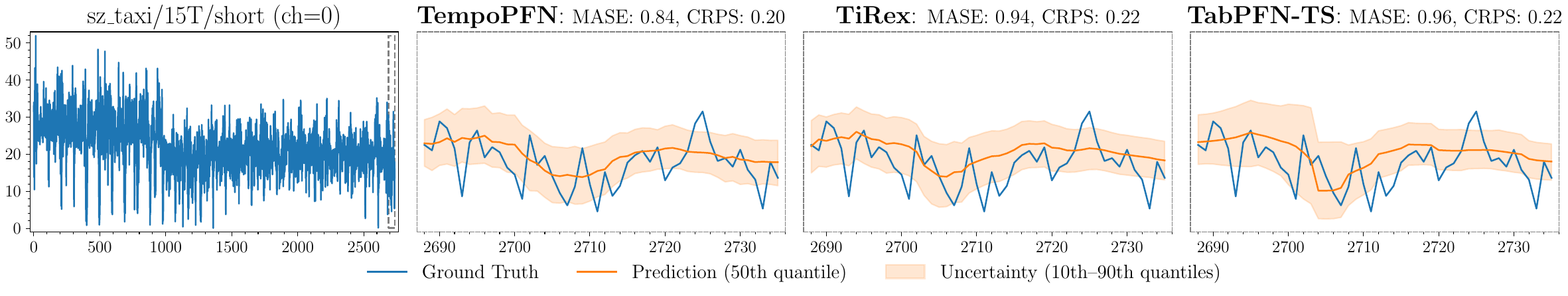}
    \includegraphics[width=\linewidth]{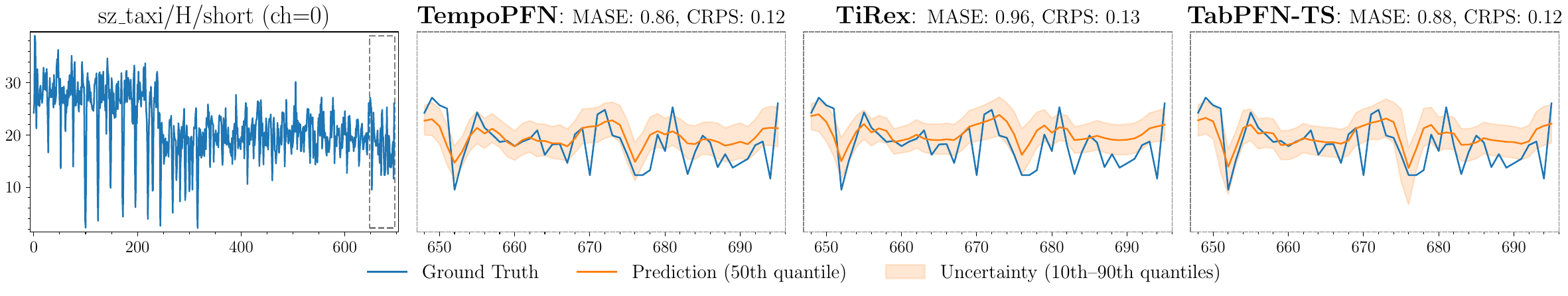}
    \includegraphics[width=\linewidth]{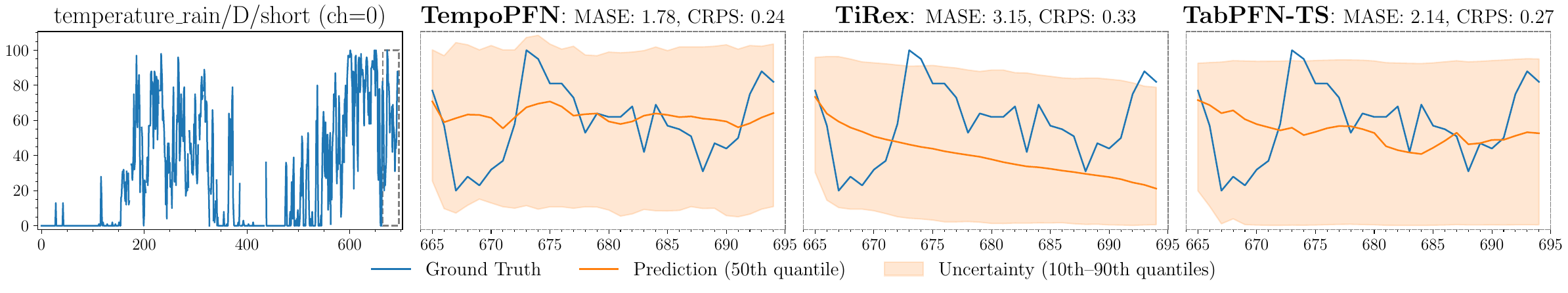}
    \includegraphics[width=\linewidth]{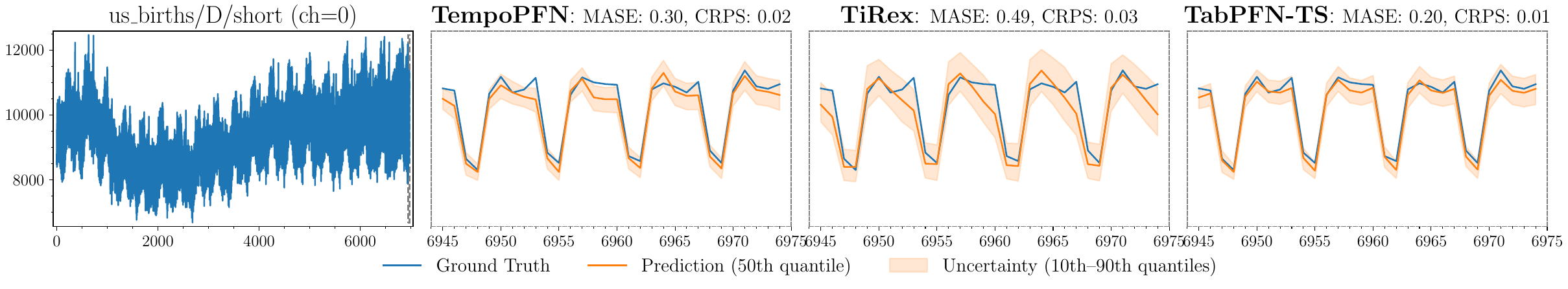}
    \includegraphics[width=\linewidth]{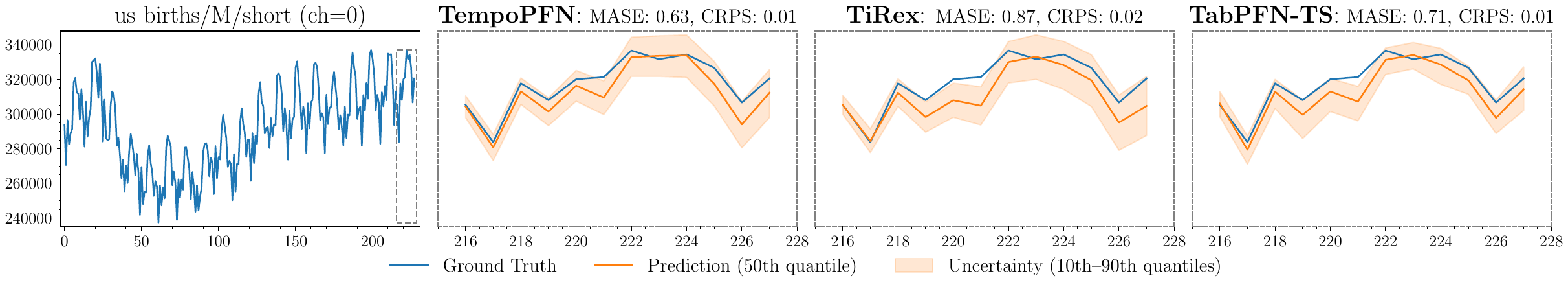}
    \includegraphics[width=\linewidth]{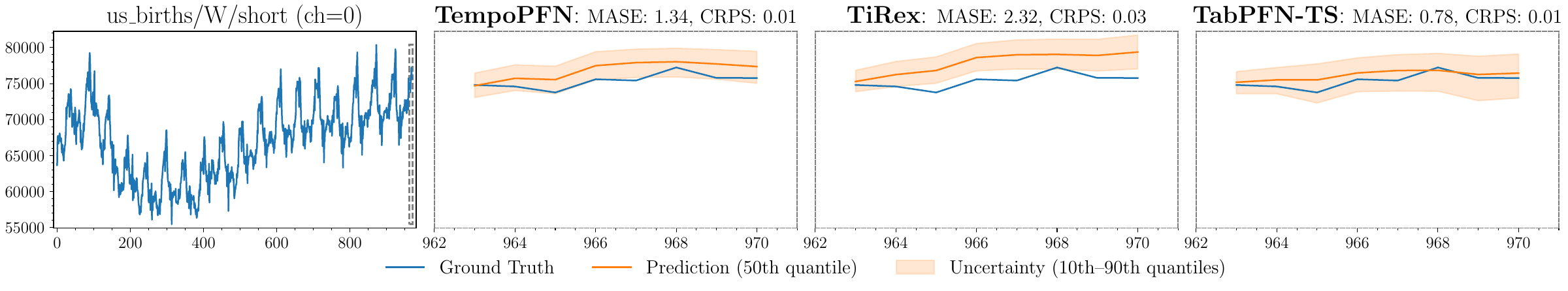}
    \caption{Qualitative comparison between \methodname{}, TiRex and TabPFN-TS on the GIFT-Eval Benchmark. (Left) Total context with prediction window in dashed grey box. (Right) Predictions between TempoPFN, TiRex, and TabPFN-TS.}
    \label{appfig:qualitative}
\end{figure}

This section presents qualitative Gift-Eval forecasts in Figure~\ref{appfig:qualitative}, showing the full history (left) alongside zoomed-in predictions for TempoPFN, TiRex, and TabPFN-TS. Note that evaluation context lengths vary: TiRex uses the full history, TabPFN-TS uses 4096 steps, and TempoPFN uses 3072.

\vfill

\begin{figure}[H]
    \centering
    \includegraphics[width=\linewidth, trim={0 23 0 0}, clip=true]{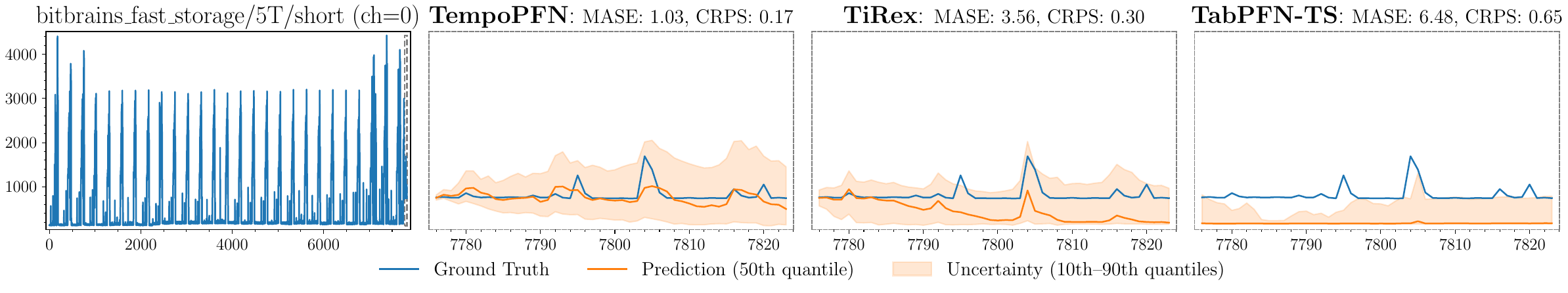}
    \includegraphics[width=\linewidth, trim={0 23 0 0}, clip=true]{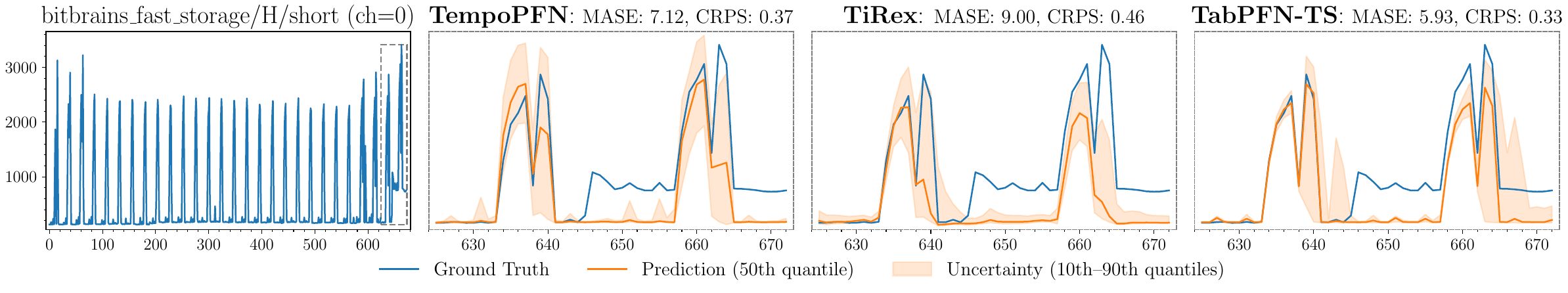}
    \includegraphics[width=\linewidth, trim={0 23 0 0}, clip=true]{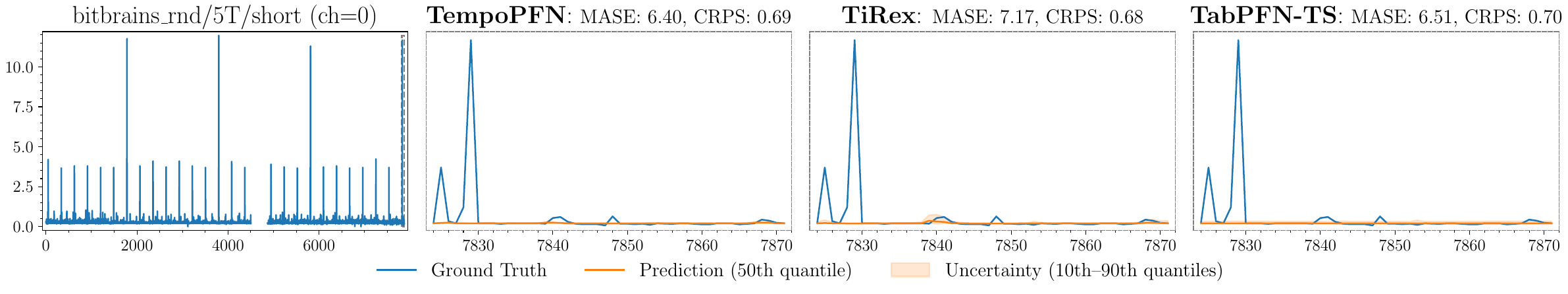}
    \includegraphics[width=\linewidth, trim={0 23 0 0}, clip=true]{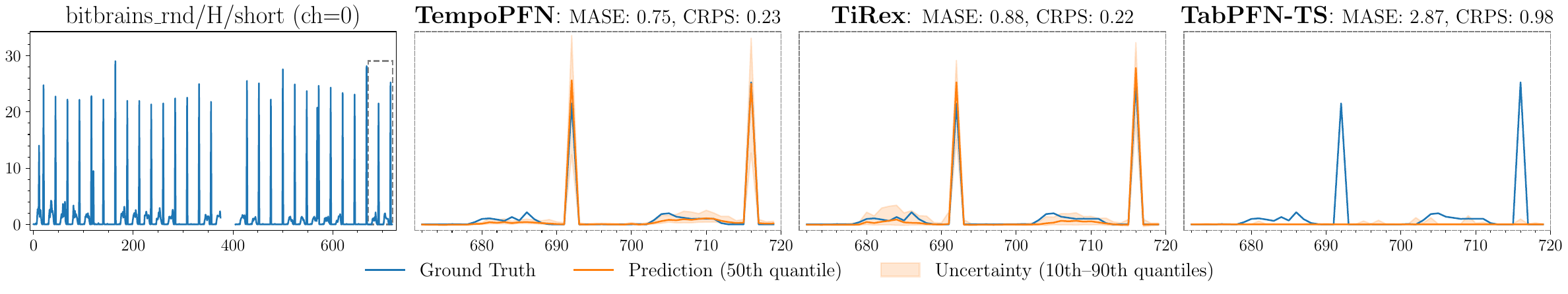}
    \includegraphics[width=\linewidth, trim={0 23 0 0}, clip=true]{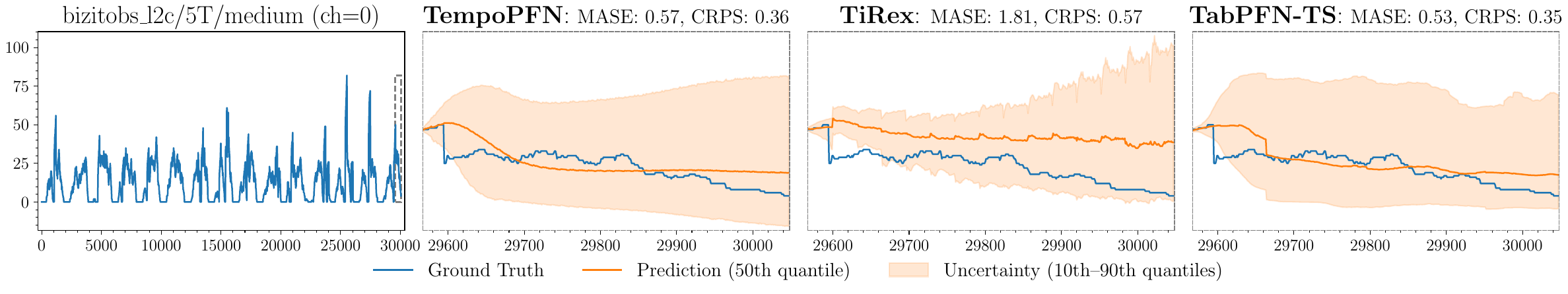}
    \includegraphics[width=\linewidth, trim={0 23 0 0}, clip=true]{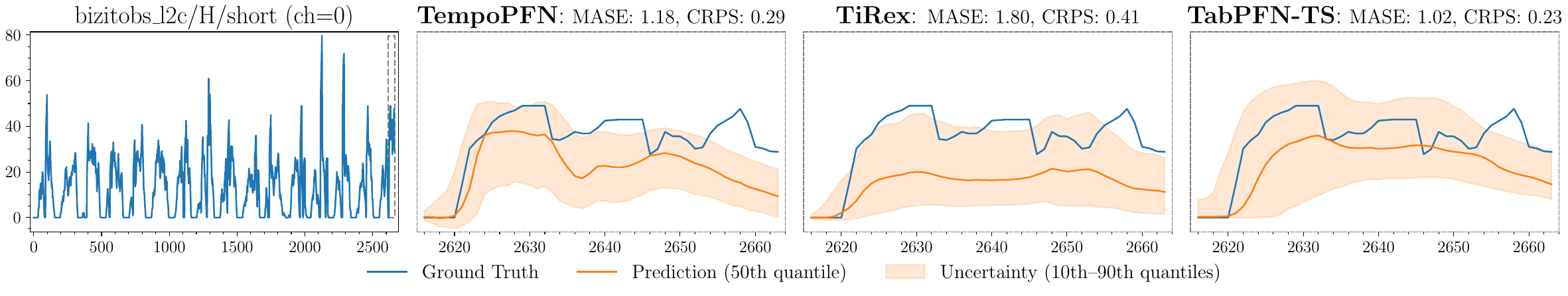}
    \includegraphics[width=\linewidth, trim={0 23 0 0}, clip=true]{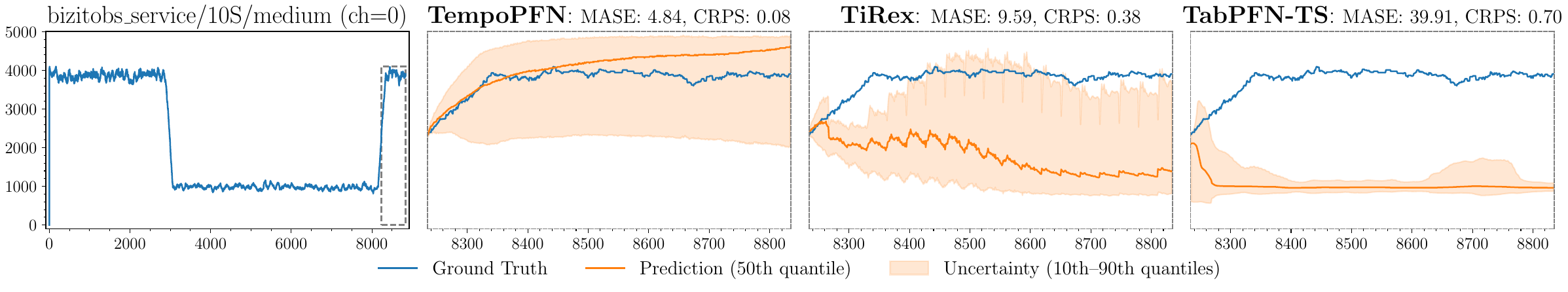}
    \includegraphics[width=\linewidth, trim={0 0 0 0}, clip=true]{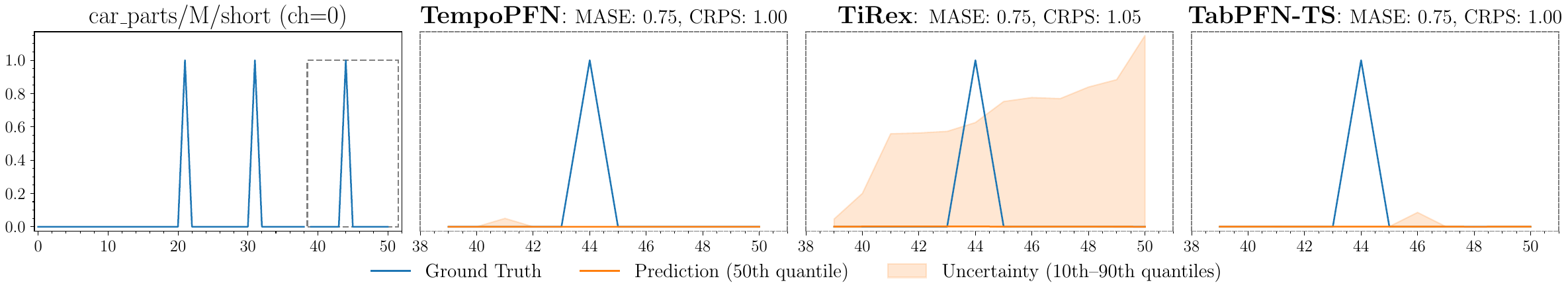}

\end{figure}

\begin{figure}[H]
    \centering
    \includegraphics[width=\linewidth, trim={0 23 0 0}, clip=true]{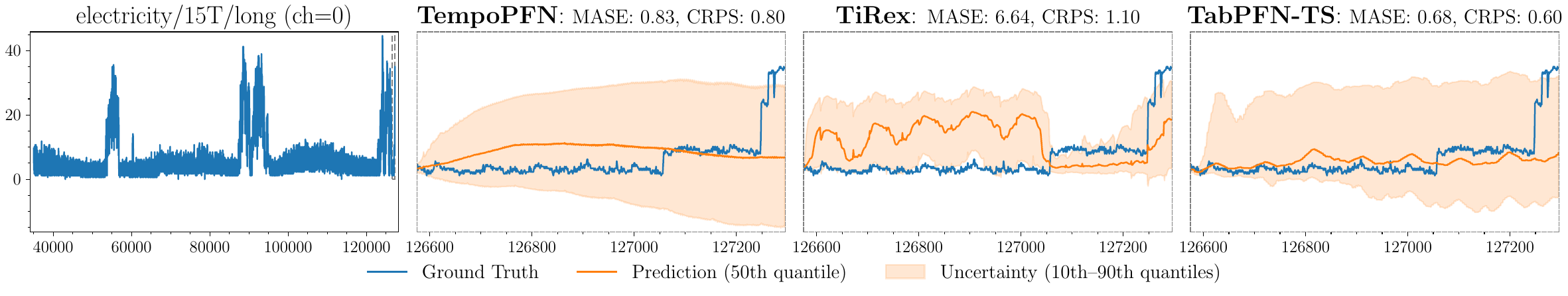}
    \includegraphics[width=\linewidth, trim={0 23 0 0}, clip=true]{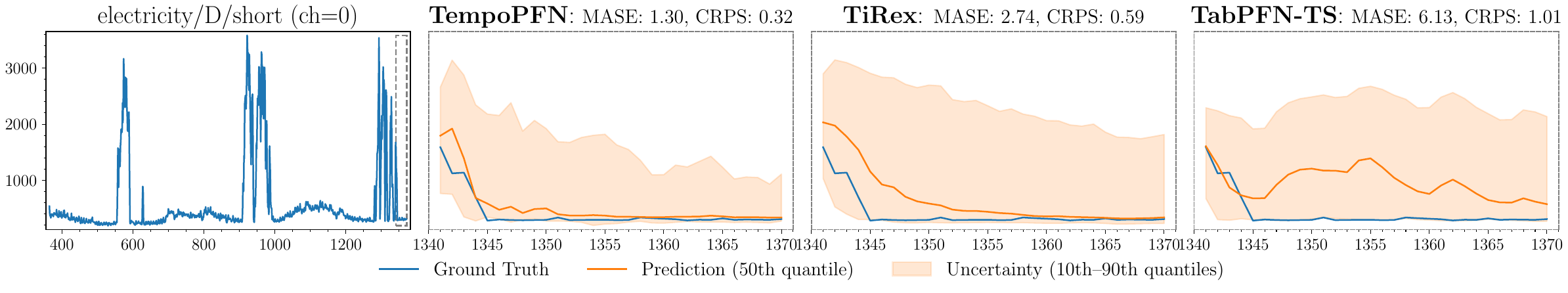}
    \includegraphics[width=\linewidth, trim={0 23 0 0}, clip=true]{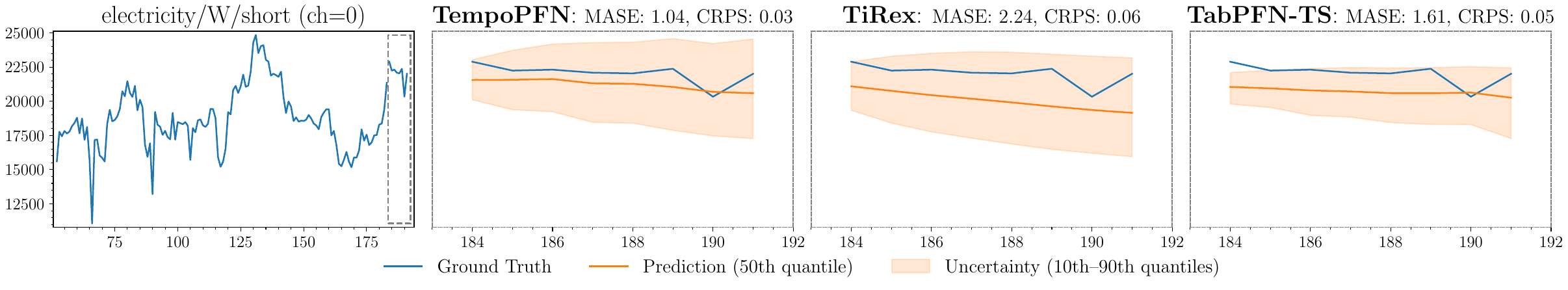}
    \includegraphics[width=\linewidth]{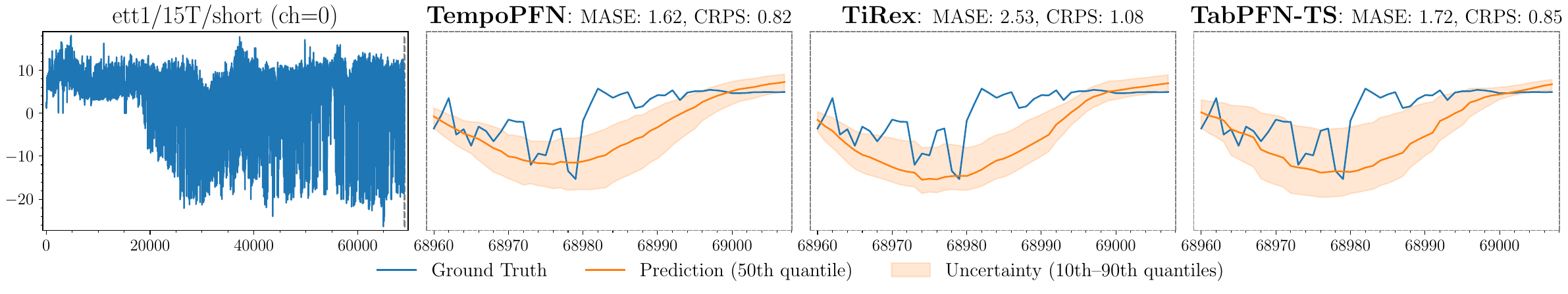}
    \includegraphics[width=\linewidth]{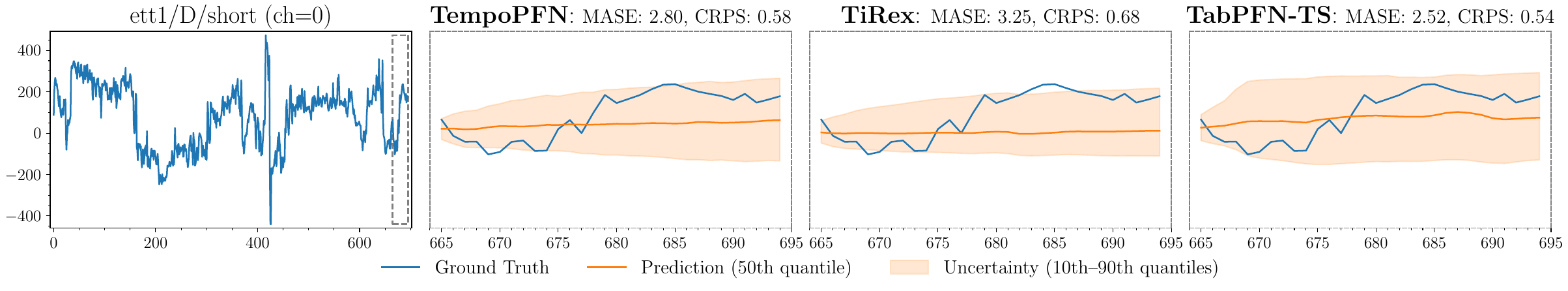}
    \includegraphics[width=\linewidth]{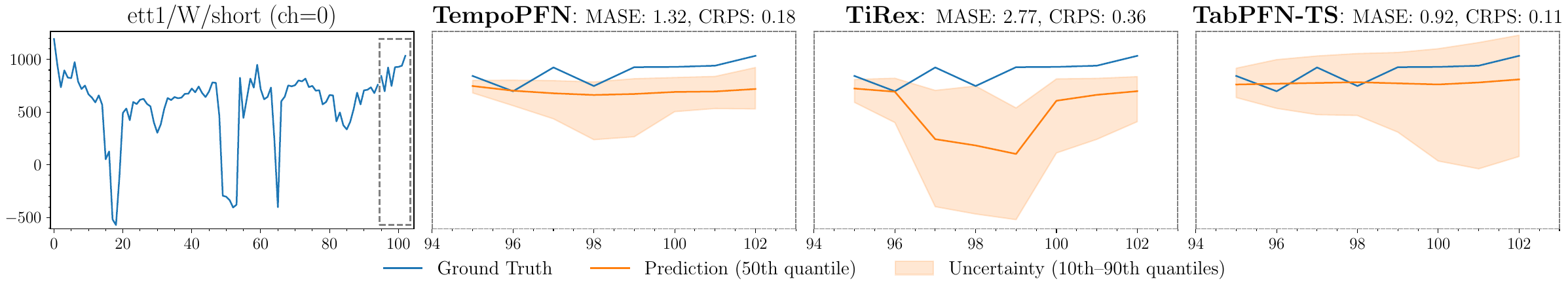}
    \includegraphics[width=\linewidth]{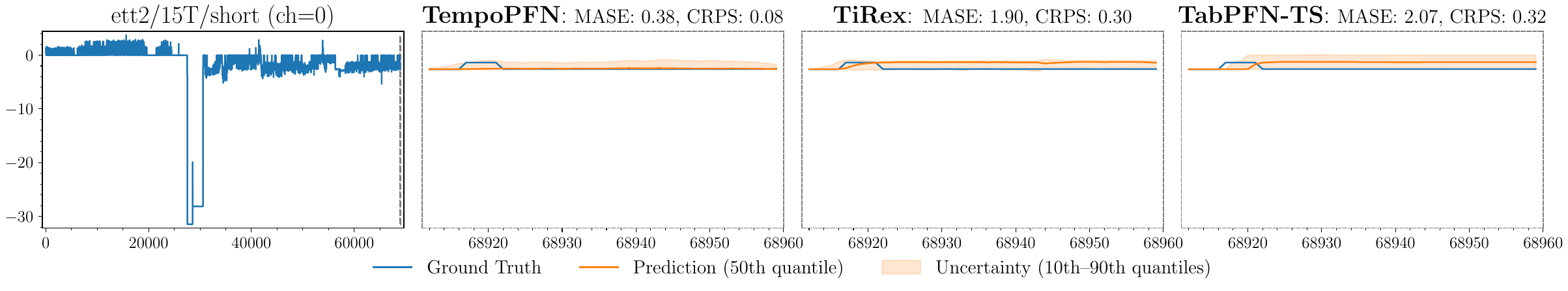}
    \includegraphics[width=\linewidth, trim={0 23 0 0}, clip=true]{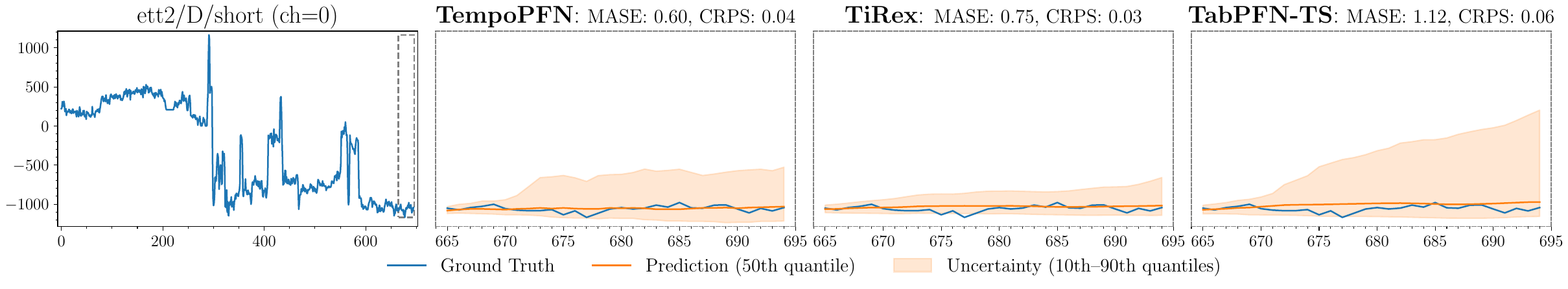}
    \includegraphics[width=\linewidth, trim={0 23 0 0}, clip=true]{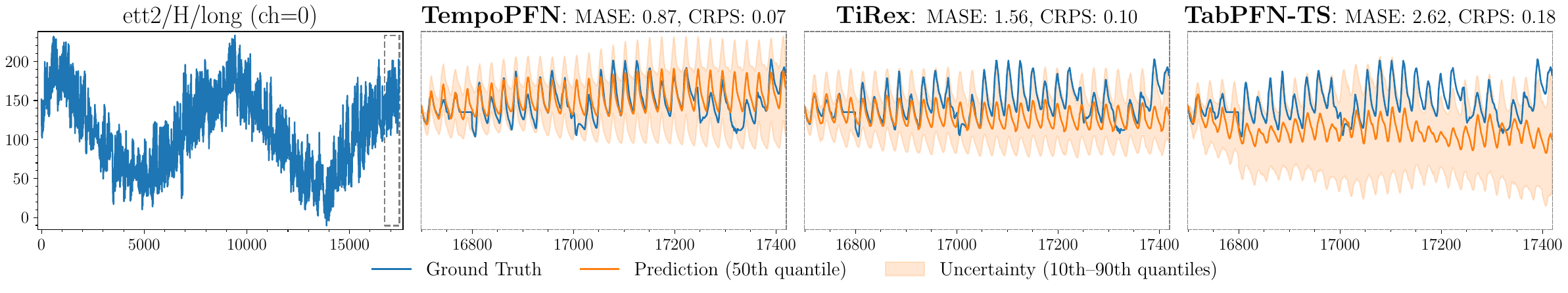}
\end{figure}

\begin{figure}[H]
    \centering
    \includegraphics[width=\linewidth, trim={0 23 0 0}, clip=true]{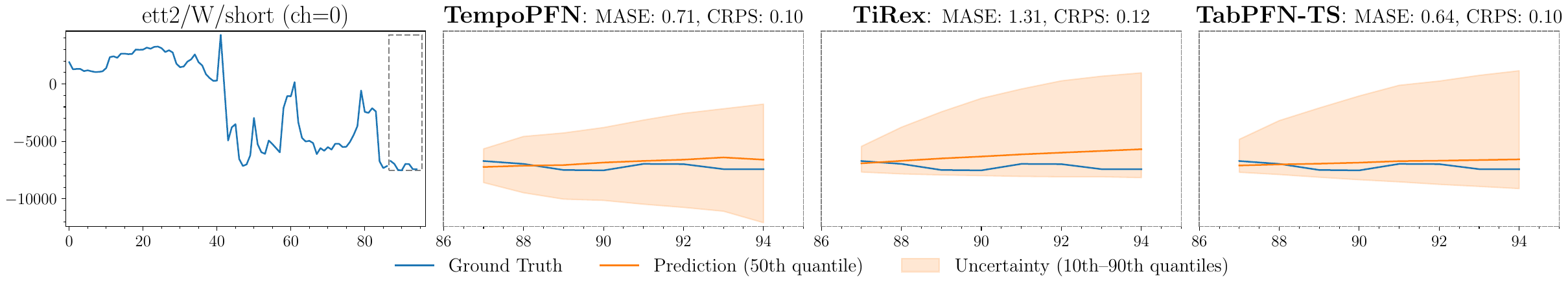}
    \includegraphics[width=\linewidth, trim={0 23 0 0}, clip=true]{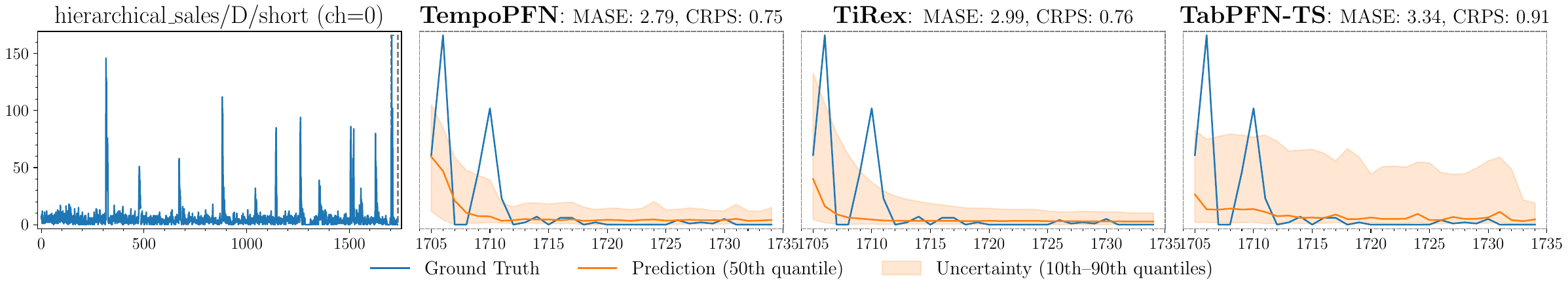}
    \includegraphics[width=\linewidth, trim={0 23 0 0}, clip=true]{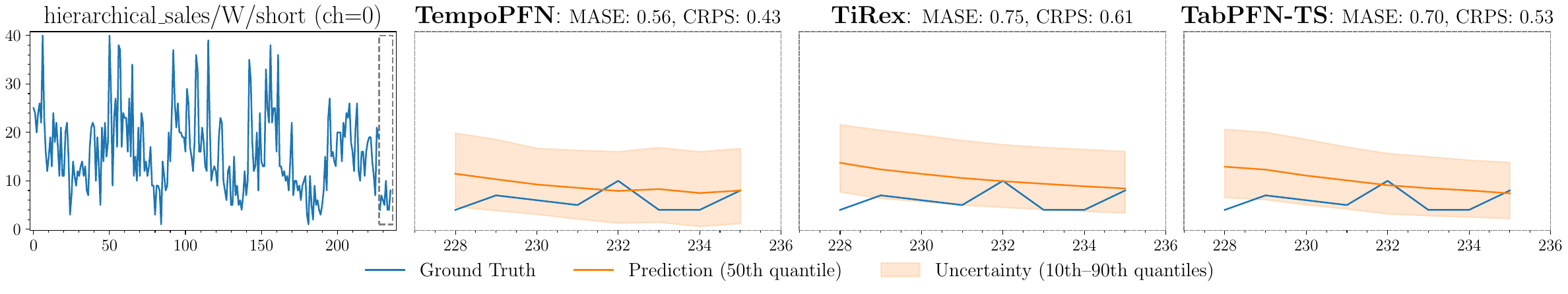}
    \includegraphics[width=\linewidth]{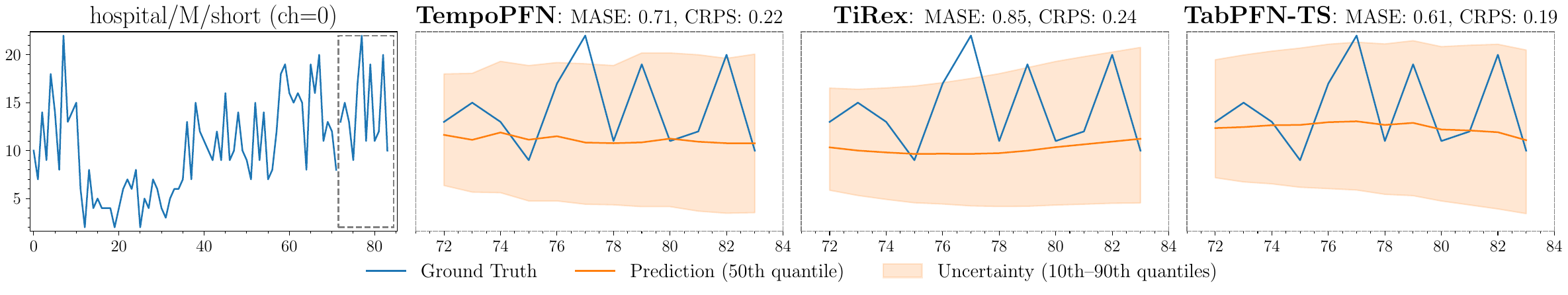}
    \includegraphics[width=\linewidth]{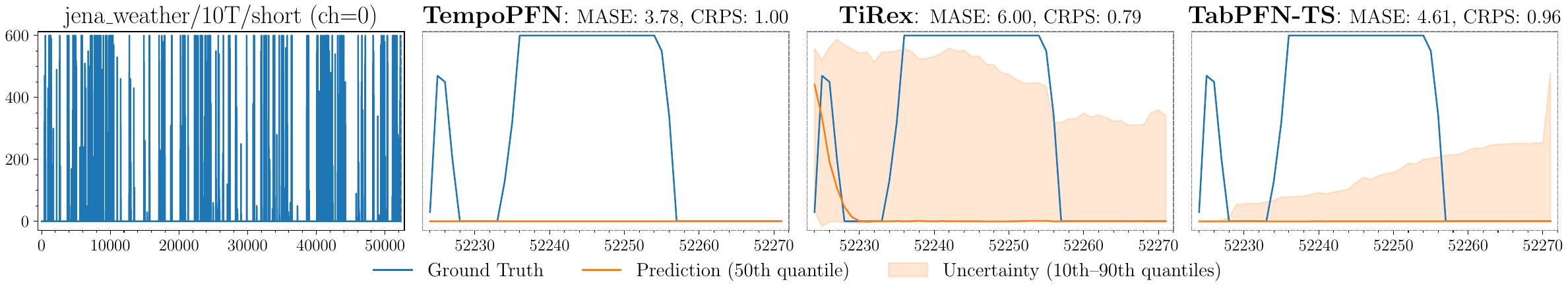}
    \includegraphics[width=\linewidth]{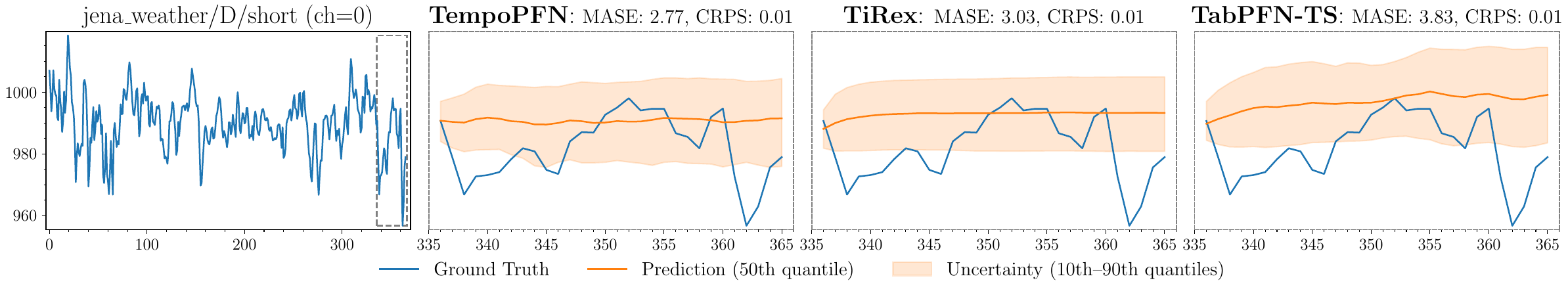}
    \includegraphics[width=\linewidth, trim={0 23 0 0}, clip=true]{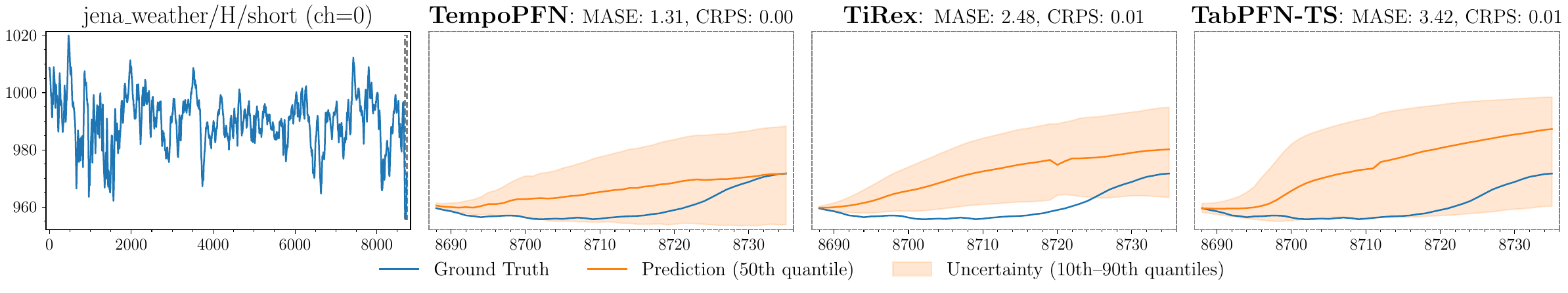}
    \includegraphics[width=\linewidth, trim={0 23 0 0}, clip=true]{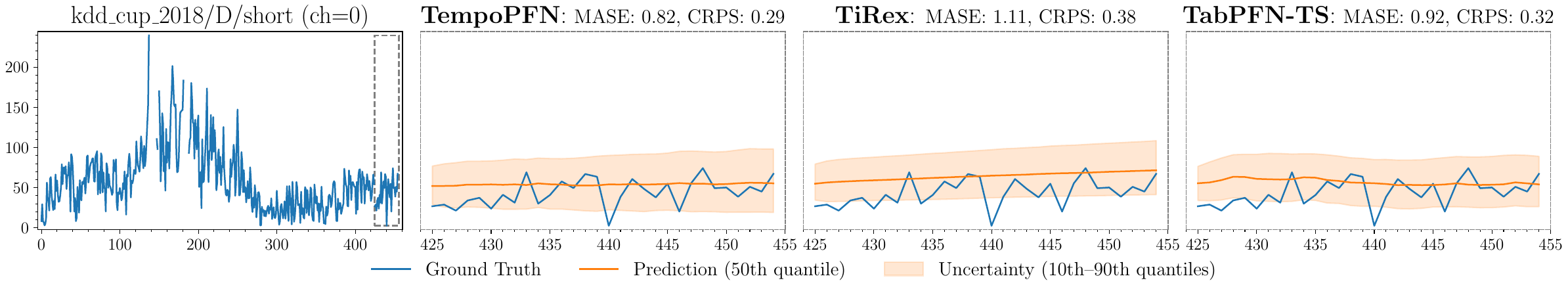}
    \includegraphics[width=\linewidth, trim={0 23 0 0}, clip=true]{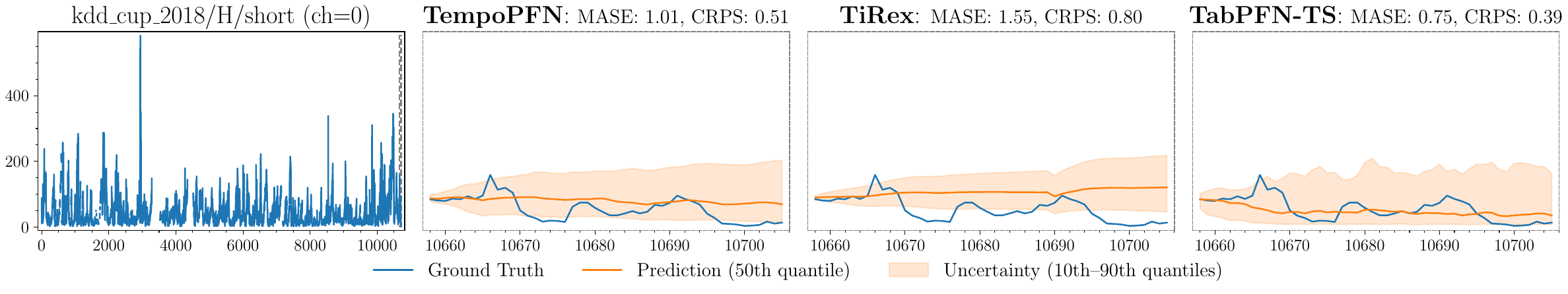}
\end{figure}

\begin{figure}[H]
    \centering
    \includegraphics[width=\linewidth, trim={0 23 0 0}, clip=true]{figures/qualtative_gift_eval_examples/loop_seattle/5T/3_w12_ch0.pdf}
    \includegraphics[width=\linewidth, trim={0 23 0 0}, clip=true]{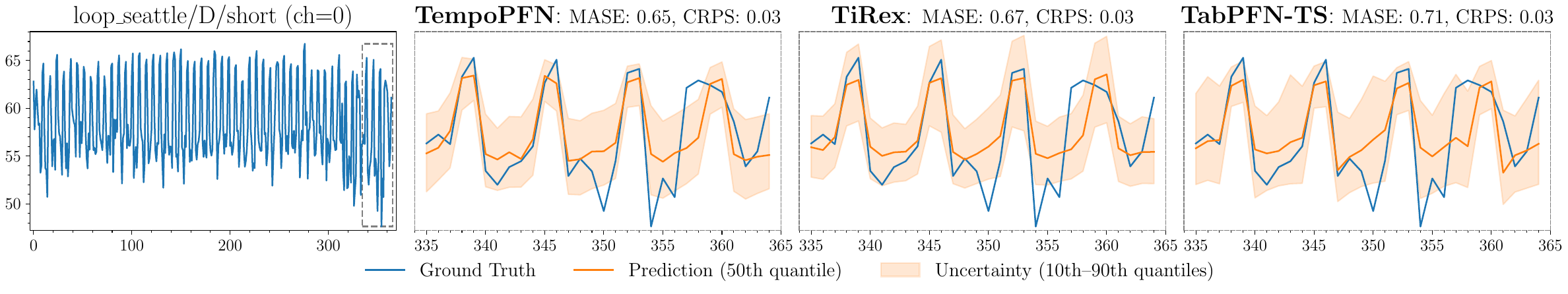}
    \includegraphics[width=\linewidth]{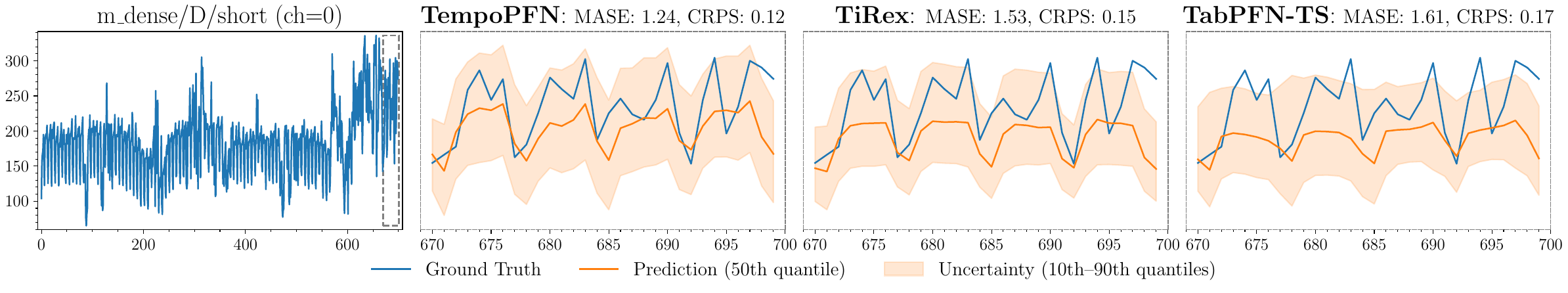}
    \includegraphics[width=\linewidth]{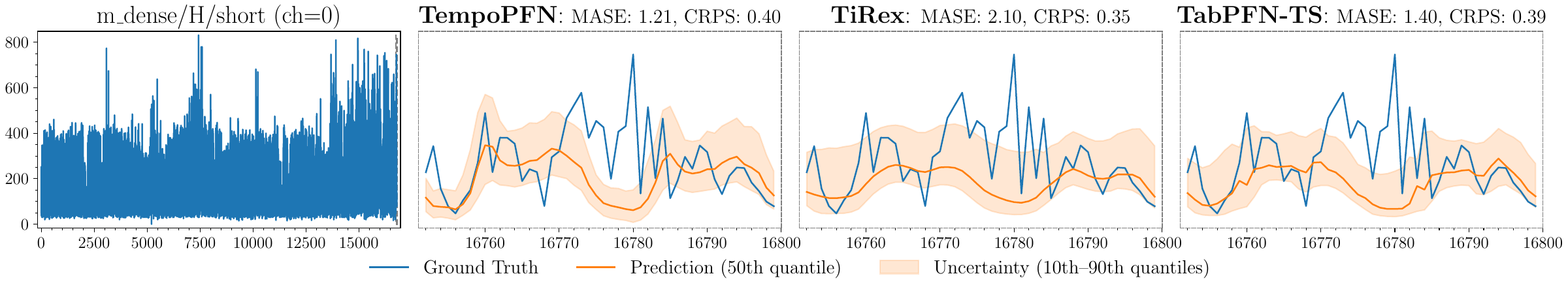}
    \includegraphics[width=\linewidth]{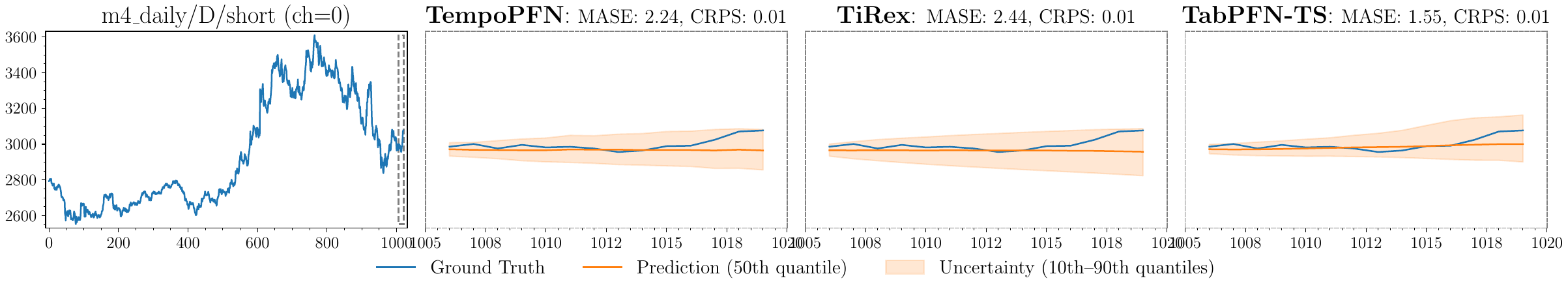}
    \includegraphics[width=\linewidth]{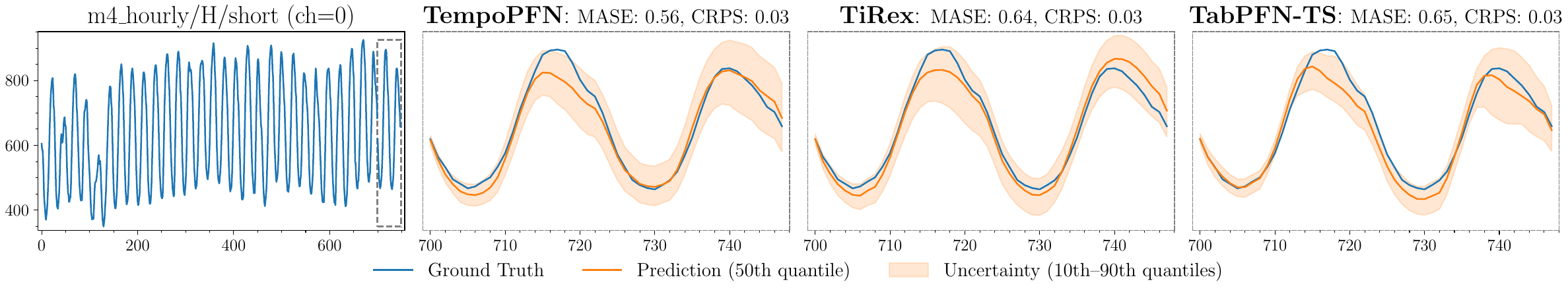}
    \includegraphics[width=\linewidth]{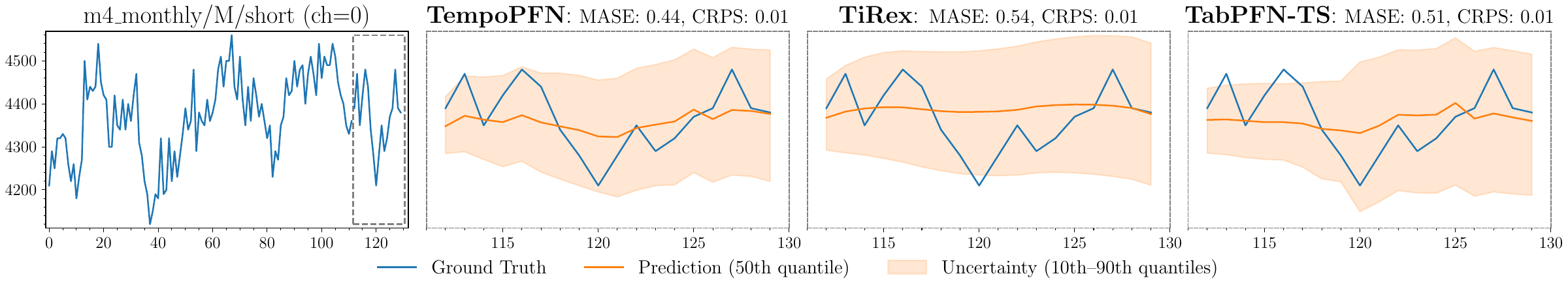}
    \includegraphics[width=\linewidth]{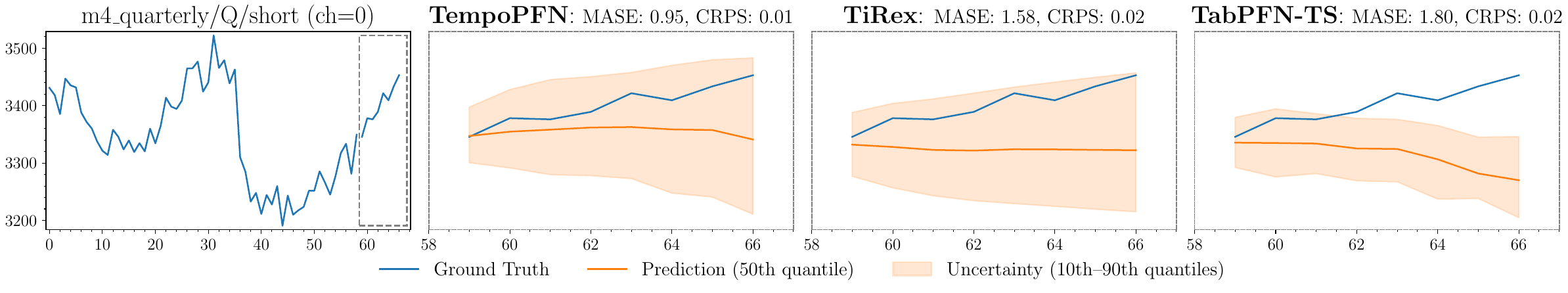}
\end{figure}

\begin{figure}[H]
    \centering
    \includegraphics[width=\linewidth]{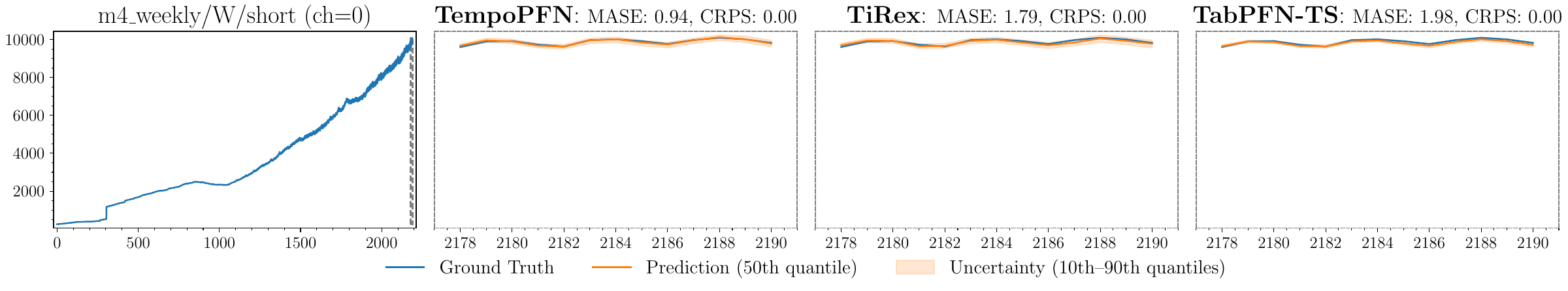}
    \includegraphics[width=\linewidth, trim={0 23 0 0}, clip=true]{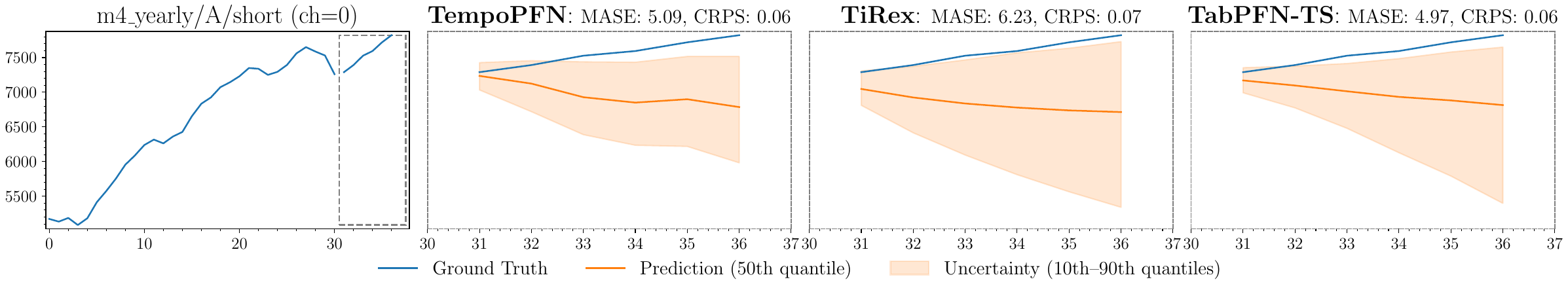}
    \includegraphics[width=\linewidth]{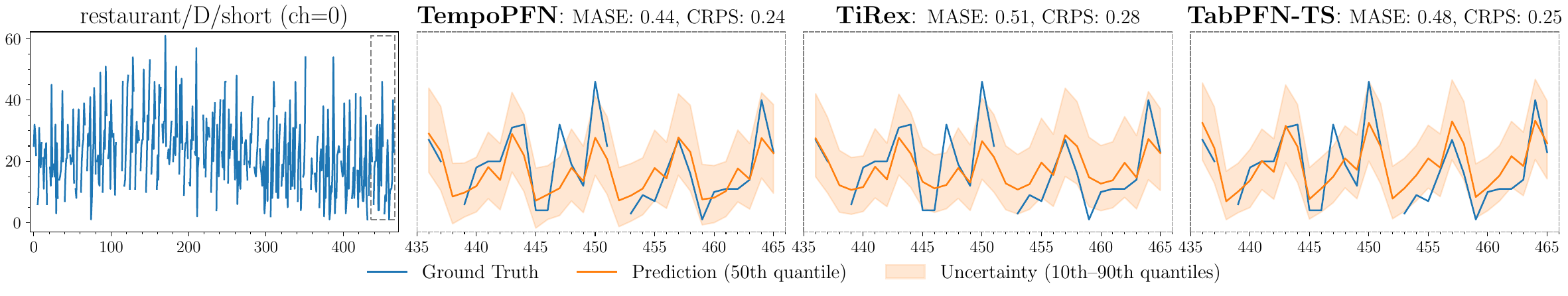}
    \includegraphics[width=\linewidth]{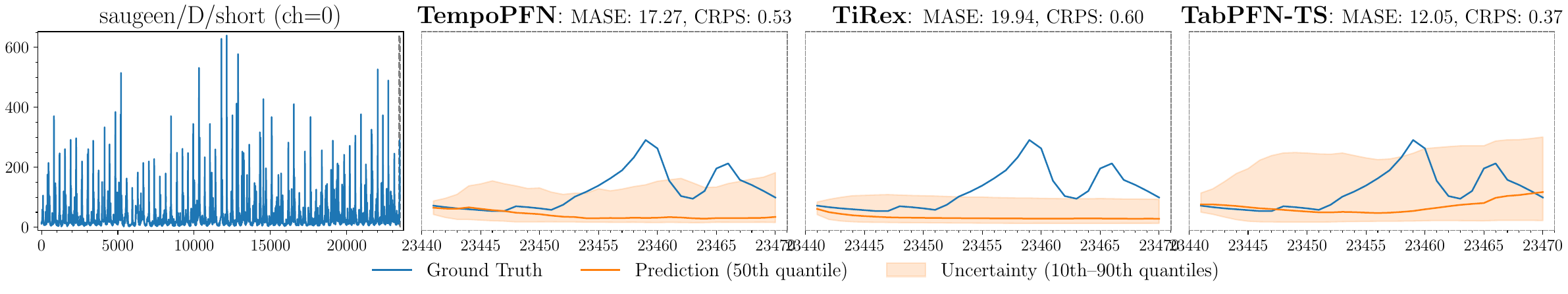}
    \includegraphics[width=\linewidth]{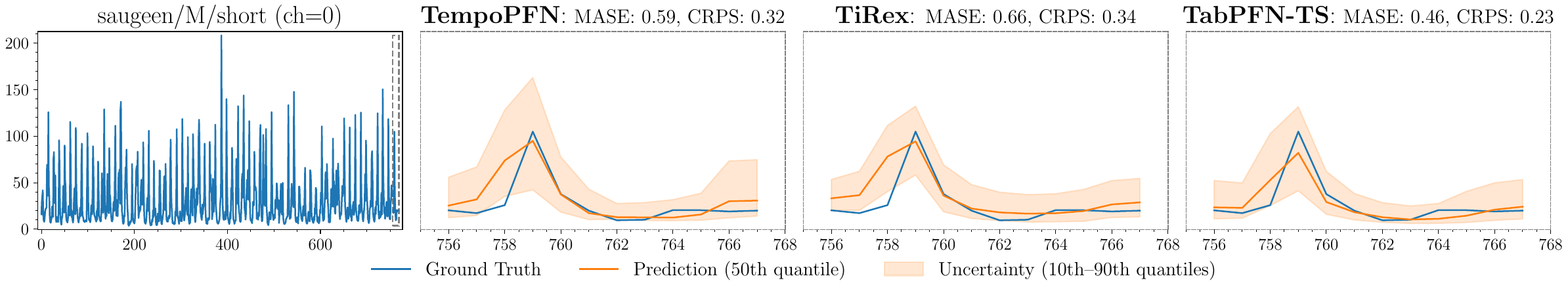}
    \includegraphics[width=\linewidth]{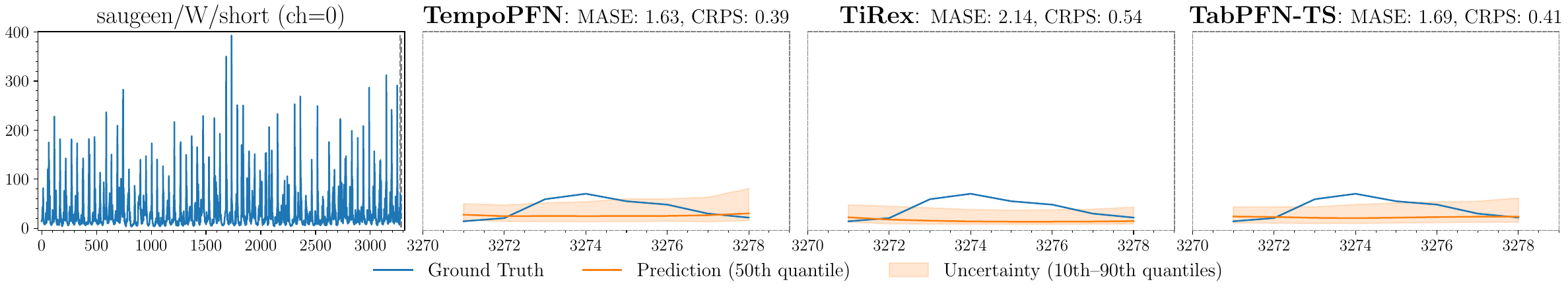}
    \includegraphics[width=\linewidth]{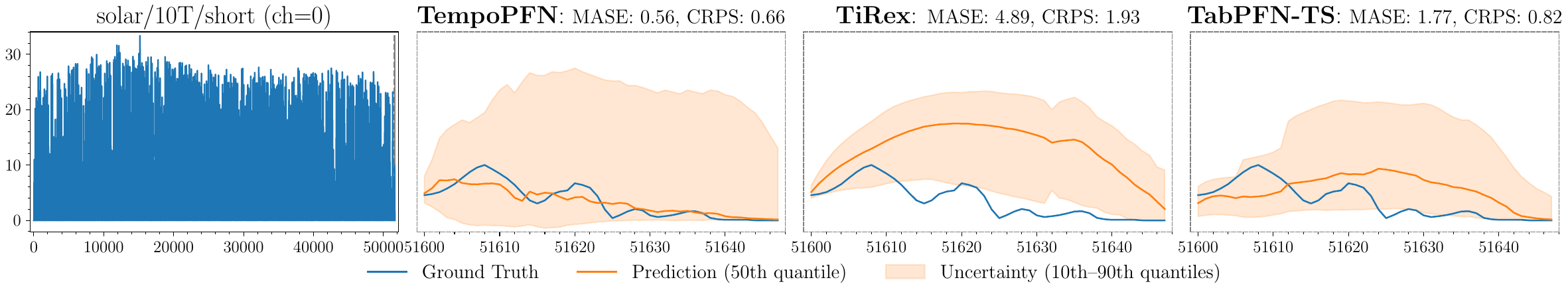}
    \includegraphics[width=\linewidth]{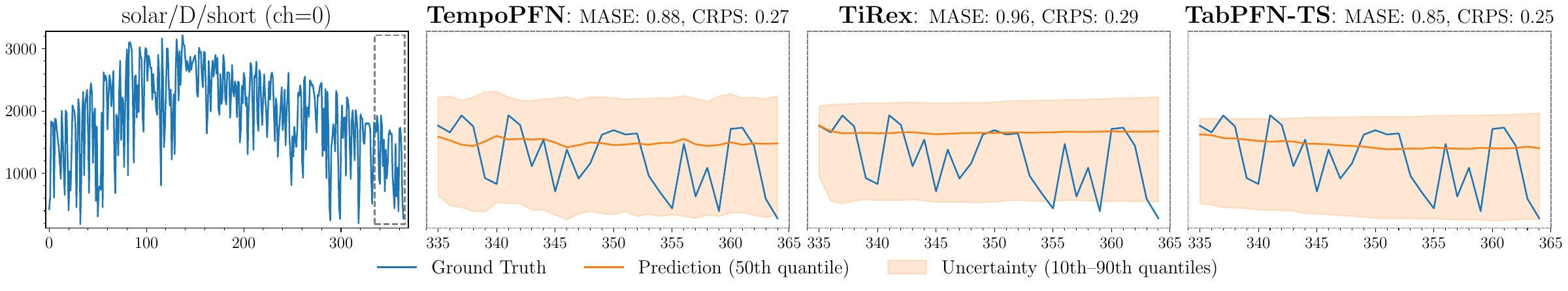}
\end{figure} 

\newpage
\subsection{Quantitative Comparison on the Gift-Eval Benchmark}\label{app:quantitative_results}

\begin{table}[ht]
\centering
\caption{Normalized CRPS (Continuous Ranked Probability Score) for various zero-shot models on the GiftEval benchmark. Scores are normalized against a seasonal naive baseline, with values less than 1.0 signifying superior probabilistic forecasts. The top two performing models are highlighted.}
\label{tab:normalized_crps}
\footnotesize
\resizebox{\textwidth}{!}{
\begin{tabular}{l | r r r r r r r r r r}
\toprule
\multirow{2}{*}{Dataset} & \rotatebox{10}{TempoPFN} & \rotatebox{10}{TiRex} & \rotatebox{10}{FlowState-9.1M} & \rotatebox{10}{Toto\_Open\_Base\_1.0} & \rotatebox{10}{TabPFN-TS} & \rotatebox{10}{YingLong\_50m} & \rotatebox{10}{Chronos Bolt B} & \rotatebox{10}{TTM-R2-Finetuned} & \rotatebox{10}{Moirai L 1.1} & \rotatebox{10}{Moirai B 1.1} \\
\midrule
bitbrains\_fast\_storage/5T/long & \textbf{0.555} & \underline{0.558} & 0.701 & 0.568 & 0.752 & 0.586 & 0.635 & 0.796 & 0.608 & 0.622 \\
bitbrains\_fast\_storage/5T/medium & \textbf{0.516} & \textbf{0.516} & 0.651 & \underline{0.525} & 0.792 & 0.537 & 0.631 & 0.690 & 0.531 & 0.553 \\
bitbrains\_fast\_storage/5T/short & \underline{0.334} & \underline{0.334} & 0.422 & \textbf{0.306} & 0.547 & 0.343 & 0.375 & 0.465 & 0.340 & 0.341 \\
bitbrains\_fast\_storage/H/short & 0.632 & 0.683 & 0.684 & \underline{0.609} & 0.655 & 0.610 & 0.757 & 1.007 & 0.632 & \textbf{0.600} \\
bitbrains\_rnd/5T/long & 0.602 & \underline{0.541} & 0.619 & \textbf{0.501} & 0.697 & 0.563 & 0.643 & 0.648 & 0.577 & 0.566 \\
bitbrains\_rnd/5T/medium & 0.530 & \textbf{0.500} & 0.619 & 0.537 & 0.701 & 0.546 & 0.517 & 0.670 & \underline{0.508} & 0.527 \\
bitbrains\_rnd/5T/short & 0.477 & \underline{0.367} & 0.420 & \textbf{0.362} & 0.552 & 0.390 & 0.398 & 0.424 & 0.379 & 0.405 \\
bitbrains\_rnd/H/short & 0.523 & 0.522 & 0.535 & 0.477 & 0.597 & 0.546 & 0.502 & 0.718 & \textbf{0.455} & \underline{0.467} \\
bizitobs\_application/10S/long & \textbf{1.041} & 1.142 & 1.152 & 1.148 & \underline{1.062} & 1.315 & 2.383 & 1.261 & 2.053 & 2.624 \\
bizitobs\_application/10S/medium & \textbf{0.560} & 0.944 & 0.985 & \underline{0.802} & 0.955 & 1.147 & 2.425 & 1.138 & 1.958 & 2.436 \\
bizitobs\_application/10S/short & \textbf{0.293} & 0.377 & 0.389 & \underline{0.333} & 0.426 & 0.506 & 1.549 & 0.638 & 1.099 & 0.941 \\
bizitobs\_l2c/5T/long & 0.885 & 0.917 & \textbf{0.355} & 0.821 & 0.472 & 0.848 & 1.138 & \underline{0.373} & 0.783 & 0.854 \\
bizitobs\_l2c/5T/medium & 0.681 & 0.684 & \textbf{0.411} & 0.606 & 0.502 & 0.690 & 0.856 & \underline{0.475} & 0.788 & 0.730 \\
bizitobs\_l2c/5T/short & 0.276 & 0.303 & 0.323 & \underline{0.265} & 0.321 & 0.298 & 0.284 & \textbf{0.263} & 0.303 & 0.297 \\
bizitobs\_l2c/H/long & 0.333 & 0.297 & \textbf{0.263} & 0.392 & 0.311 & 0.401 & \underline{0.295} & 0.736 & 0.638 & 0.526 \\
bizitobs\_l2c/H/medium & 0.301 & 0.284 & \textbf{0.245} & 0.394 & \underline{0.263} & 0.396 & 0.281 & 0.746 & 0.685 & 0.761 \\
bizitobs\_l2c/H/short & 0.381 & 0.413 & \textbf{0.362} & 0.382 & 0.402 & 0.516 & \underline{0.363} & 0.516 & 1.073 & 0.946 \\
bizitobs\_service/10S/long & 1.043 & 1.001 & 1.001 & \textbf{0.955} & \underline{0.965} & 1.220 & 2.111 & 1.095 & 1.946 & 2.151 \\
bizitobs\_service/10S/medium & \textbf{0.393} & 0.639 & \underline{0.562} & 0.574 & 0.860 & 0.981 & 2.022 & 0.896 & 1.455 & 1.892 \\
bizitobs\_service/10S/short & \underline{0.289} & 0.327 & 0.318 & \textbf{0.286} & 0.482 & 0.444 & 1.267 & 0.354 & 0.788 & 1.053 \\
car\_parts/M/short & 0.590 & 0.576 & 0.585 & \textbf{0.522} & \underline{0.563} & 0.817 & 0.578 & 0.641 & 0.685 & 0.580 \\
covid\_deaths/D/short & \underline{0.234} & 0.273 & 0.339 & \textbf{0.215} & 0.324 & 0.468 & 0.371 & 0.287 & 0.362 & 0.346 \\
electricity/15T/long & 1.010 & \textbf{0.656} & \underline{0.666} & 0.761 & 0.716 & 0.744 & 0.748 & 0.690 & 0.878 & 1.022 \\
electricity/15T/medium & 0.907 & \textbf{0.655} & \underline{0.671} & 0.762 & 0.734 & 0.754 & 0.734 & 0.683 & 0.913 & 0.940 \\
electricity/15T/short & 0.682 & 0.503 & 0.562 & 0.603 & 0.587 & 0.565 & \textbf{0.496} & \underline{0.497} & 0.776 & 0.728 \\
electricity/D/short & 0.639 & \textbf{0.525} & \underline{0.529} & 0.566 & 0.603 & 0.560 & 0.531 & 0.560 & 0.666 & 0.583 \\
electricity/H/long & 0.652 & 0.607 & 0.571 & \textbf{0.544} & 0.706 & 0.682 & 0.637 & 0.587 & 0.671 & \underline{0.557} \\
electricity/H/medium & 0.685 & 0.617 & \underline{0.603} & \textbf{0.589} & 0.688 & 0.677 & 0.636 & 0.615 & 0.685 & 0.645 \\
electricity/H/short & 0.730 & \textbf{0.571} & 0.632 & 0.652 & 0.677 & 0.744 & \underline{0.602} & 0.643 & 0.732 & 0.711 \\
electricity/W/short & 0.611 & \underline{0.459} & \textbf{0.439} & 0.641 & 0.550 & 0.636 & 0.477 & 0.605 & 0.625 & 0.775 \\
ett1/15T/long & 0.847 & 0.721 & \textbf{0.676} & 0.738 & 0.763 & \underline{0.712} & 0.875 & 0.750 & 1.052 & 0.803 \\
ett1/15T/medium & 0.864 & 0.777 & \textbf{0.713} & 0.810 & 0.785 & \underline{0.763} & 0.874 & 0.847 & 1.064 & 1.008 \\
ett1/15T/short & 0.753 & \underline{0.662} & 0.670 & 0.670 & 0.691 & 0.704 & \textbf{0.654} & 0.724 & 0.936 & 0.800 \\
ett1/D/short & \textbf{0.658} & 0.678 & 0.685 & 0.695 & 0.729 & \underline{0.664} & 0.703 & 0.838 & 0.700 & 0.737 \\
ett1/H/long & 0.598 & \underline{0.551} & 0.573 & 0.566 & 0.626 & \textbf{0.550} & 0.660 & 0.597 & 0.628 & 0.609 \\
ett1/H/medium & 0.634 & \underline{0.578} & \textbf{0.569} & 0.584 & 0.651 & \underline{0.578} & 0.696 & 0.628 & 0.621 & 0.648 \\
ett1/H/short & 0.775 & 0.747 & \textbf{0.733} & 0.806 & 0.807 & \underline{0.746} & 0.753 & 0.814 & 0.786 & 0.820 \\
ett1/W/short & \underline{0.827} & 1.000 & \textbf{0.786} & 0.843 & 0.911 & 0.931 & 0.949 & 0.906 & 0.834 & 0.837 \\
ett2/15T/long & 0.768 & 0.721 & \underline{0.680} & \textbf{0.665} & 0.761 & 0.702 & 0.839 & 0.718 & 0.866 & 1.031 \\
ett2/15T/medium & 0.776 & 0.744 & \textbf{0.703} & 0.748 & 0.808 & \underline{0.741} & 0.889 & 0.776 & 0.846 & 0.878 \\
ett2/15T/short & 0.711 & 0.685 & \underline{0.683} & 0.706 & 0.756 & 0.696 & 0.691 & \textbf{0.662} & 0.833 & 0.805 \\
ett2/D/short & 0.783 & \underline{0.594} & 0.674 & 0.727 & 0.821 & 0.613 & 0.610 & \textbf{0.553} & 0.611 & 0.618 \\
ett2/H/long & \textbf{0.498} & 0.549 & 0.536 & 0.521 & 0.669 & 0.514 & 0.562 & \underline{0.504} & 0.601 & 0.529 \\
ett2/H/medium & 0.567 & 0.568 & 0.560 & \underline{0.547} & 0.648 & 0.548 & 0.616 & 0.560 & 0.634 & \textbf{0.537} \\
ett2/H/short & 0.730 & 0.744 & \textbf{0.710} & 0.725 & 0.821 & 0.730 & \underline{0.713} & 0.742 & 0.775 & 0.807 \\
ett2/W/short & 0.687 & \textbf{0.645} & 0.779 & 0.792 & 0.738 & 0.817 & 0.660 & 0.702 & 0.815 & \underline{0.652} \\
hierarchical\_sales/D/short & 0.336 & \textbf{0.327} & 0.336 & \underline{0.328} & 0.341 & 0.340 & 0.332 & 0.349 & 0.334 & 0.331 \\
hierarchical\_sales/W/short & \underline{0.412} & 0.416 & \textbf{0.408} & 0.428 & 0.415 & 0.467 & 0.424 & 0.435 & 0.431 & 0.429 \\
hospital/M/short & 0.840 & 0.830 & \textbf{0.766} & 0.838 & 0.860 & 0.926 & 0.913 & 0.841 & 0.821 & \underline{0.810} \\
jena\_weather/10T/long & 0.240 & 0.220 & \underline{0.214} & \textbf{0.210} & 0.224 & 0.254 & 0.269 & 0.286 & 0.325 & 0.297 \\
jena\_weather/10T/medium & 0.248 & 0.239 & \underline{0.234} & \textbf{0.231} & 0.254 & 0.272 & 0.270 & 0.324 & 0.338 & 0.320 \\
jena\_weather/10T/short & 0.204 & 0.200 & \underline{0.185} & \textbf{0.172} & 0.219 & 0.238 & 0.210 & 0.285 & 0.331 & 0.345 \\
jena\_weather/D/short & 0.239 & \underline{0.218} & 0.226 & 0.240 & 0.224 & 0.233 & \textbf{0.214} & 0.343 & 0.243 & 0.236 \\
jena\_weather/H/long & 0.147 & \textbf{0.134} & 0.159 & \underline{0.136} & 0.245 & 0.145 & 0.147 & 0.449 & 0.145 & 0.156 \\
jena\_weather/H/medium & 0.164 & \textbf{0.153} & \underline{0.154} & \underline{0.154} & 0.169 & 0.165 & 0.158 & 0.204 & 0.168 & 0.166 \\
jena\_weather/H/short & 0.273 & \textbf{0.266} & \underline{0.272} & 0.274 & 0.273 & 0.284 & 0.274 & 0.403 & 0.291 & 0.284 \\
kdd\_cup\_2018/D/short & 0.561 & 0.577 & 0.565 & 0.574 & \textbf{0.537} & \underline{0.551} & 0.552 & 0.593 & 0.565 & 0.557 \\
kdd\_cup\_2018/H/long & 0.476 & \underline{0.353} & 0.489 & 0.488 & 0.510 & 0.475 & \textbf{0.320} & 0.507 & 0.404 & 0.446 \\
kdd\_cup\_2018/H/medium & 0.563 & \underline{0.437} & 0.576 & 0.582 & 0.593 & 0.553 & \textbf{0.397} & 0.597 & 0.510 & 0.581 \\
kdd\_cup\_2018/H/short & 0.701 & \underline{0.498} & 0.697 & 0.736 & 0.763 & 0.694 & \textbf{0.450} & 0.758 & 0.661 & 0.710 \\
loop\_seattle/5T/long & 0.867 & 0.680 & 0.582 & 0.602 & 0.710 & 0.790 & 1.015 & 0.659 & \textbf{0.387} & \underline{0.407} \\
loop\_seattle/5T/medium & 0.910 & 0.692 & 0.589 & 0.617 & 0.741 & 0.848 & 0.989 & 0.661 & \textbf{0.328} & \underline{0.386} \\
loop\_seattle/5T/short & 0.701 & 0.606 & 0.618 & 0.598 & 0.650 & 0.710 & 0.675 & 0.634 & \textbf{0.512} & \underline{0.565} \\
loop\_seattle/D/short & 0.417 & \underline{0.409} & \textbf{0.407} & 0.430 & 0.418 & 0.425 & 0.424 & 0.441 & 0.438 & 0.428 \\
loop\_seattle/H/long & 0.405 & \textbf{0.328} & 0.338 & 0.345 & \underline{0.334} & 0.364 & 0.406 & 0.360 & 0.397 & 0.425 \\
loop\_seattle/H/medium & 0.451 & \underline{0.400} & 0.420 & \textbf{0.398} & 0.412 & 0.430 & 0.470 & 0.429 & 0.433 & 0.492 \\
loop\_seattle/H/short & 0.669 & \textbf{0.569} & \underline{0.583} & 0.609 & 0.608 & 0.609 & 0.624 & 0.642 & 0.634 & 0.709 \\
m4\_daily/D/short & 1.048 & \textbf{0.827} & 0.998 & 0.898 & 0.928 & 0.975 & \underline{0.865} & 0.970 & 1.244 & 1.646 \\
m4\_hourly/H/short & 0.759 & \underline{0.536} & 0.545 & 0.921 & 0.788 & 0.641 & 0.674 & 0.919 & \textbf{0.527} & 0.591 \\
m4\_monthly/M/short & \textbf{0.749} & \underline{0.756} & 0.759 & 0.795 & 0.768 & 0.860 & 0.769 & 0.821 & 0.780 & 0.769 \\
m4\_quarterly/Q/short & 0.763 & \underline{0.753} & 0.770 & 0.791 & 0.798 & 0.892 & 0.786 & 0.809 & \textbf{0.749} & \textbf{0.749} \\
m4\_weekly/W/short & 0.689 & \textbf{0.589} & \underline{0.602} & 0.806 & 0.608 & 0.681 & 0.627 & 0.722 & 0.764 & 0.792 \\
m4\_yearly/A/short & 0.872 & 0.857 & 0.780 & 0.887 & 0.858 & 1.096 & 0.886 & 0.864 & \textbf{0.758} & \underline{0.766} \\
m\_dense/D/short & \underline{0.291} & 0.292 & 0.308 & 0.328 & \textbf{0.269} & 0.354 & 0.304 & 0.311 & 0.420 & 0.458 \\
m\_dense/H/long & 0.438 & 0.291 & \underline{0.289} & 0.307 & 0.394 & 0.417 & 0.405 & 0.306 & \textbf{0.272} & 0.291 \\
m\_dense/H/medium & 0.459 & 0.318 & \underline{0.304} & 0.321 & 0.424 & 0.415 & 0.417 & 0.322 & \textbf{0.297} & 0.326 \\
m\_dense/H/short & 0.627 & 0.472 & 0.478 & 0.540 & 0.564 & 0.573 & \textbf{0.455} & 0.514 & \underline{0.466} & 0.509 \\
restaurant/D/short & 0.381 & \textbf{0.377} & \underline{0.378} & 0.439 & 0.388 & 0.401 & 0.390 & 0.397 & 0.399 & 0.393 \\
saugeen/D/short & 0.642 & 0.669 & \textbf{0.562} & 0.604 & 0.638 & 0.627 & \underline{0.578} & 0.694 & 0.694 & 0.605 \\
saugeen/M/short & 0.697 & 0.717 & \underline{0.664} & 0.671 & \textbf{0.621} & 0.707 & 0.665 & 0.764 & 0.728 & 0.782 \\
saugeen/W/short & 0.546 & \textbf{0.477} & \underline{0.486} & 0.531 & 0.539 & 0.512 & 0.494 & 0.606 & 0.586 & 0.576 \\
solar/10T/long & 0.748 & \textbf{0.480} & 0.501 & 0.523 & \underline{0.491} & 0.830 & 0.657 & 0.722 & 1.144 & 1.340 \\
solar/10T/medium & 0.664 & 0.549 & 0.544 & \underline{0.539} & \textbf{0.498} & 0.819 & 0.666 & 0.759 & 1.140 & 1.270 \\
solar/10T/short & \underline{0.538} & 0.640 & 0.605 & 0.629 & \textbf{0.533} & 0.663 & 0.595 & 0.633 & 0.693 & 0.714 \\
solar/D/short & 0.501 & 0.506 & \underline{0.497} & 0.518 & \textbf{0.481} & 0.516 & 0.514 & 0.541 & 0.522 & 0.528 \\
solar/H/long & 0.339 & \textbf{0.237} & 0.309 & \underline{0.307} & 0.325 & 0.326 & 0.376 & 0.563 & 0.322 & 0.334 \\
solar/H/medium & 0.409 & \textbf{0.281} & 0.365 & 0.350 & \underline{0.331} & 0.380 & 0.390 & 0.591 & 0.366 & 0.350 \\
solar/H/short & 0.573 & \textbf{0.457} & 0.563 & 0.555 & 0.567 & 0.576 & \underline{0.504} & 0.691 & 0.563 & 0.571 \\
solar/W/short & 0.763 & 0.779 & \underline{0.583} & 0.887 & \textbf{0.572} & 1.452 & 0.632 & 0.817 & 1.016 & 1.120 \\
sz\_taxi/15T/long & 0.497 & \textbf{0.460} & 0.483 & \underline{0.472} & 0.567 & 0.477 & 0.581 & 0.558 & 0.498 & 0.488 \\
sz\_taxi/15T/medium & 0.558 & \textbf{0.533} & 0.548 & \underline{0.542} & 0.608 & 0.549 & 0.644 & 0.567 & 0.567 & 0.556 \\
sz\_taxi/15T/short & 0.669 & \textbf{0.649} & 0.656 & 0.657 & 0.676 & 0.658 & \underline{0.654} & 0.682 & 0.696 & 0.690 \\
sz\_taxi/H/short & 0.650 & \textbf{0.634} & 0.647 & 0.642 & 0.656 & 0.640 & \underline{0.636} & 0.662 & 0.683 & 0.669 \\
temperature\_rain/D/short & 0.451 & 0.434 & 0.434 & 0.441 & 0.449 & 0.462 & 0.424 & 0.518 & \textbf{0.378} & \underline{0.422} \\
us\_births/D/short & \underline{0.169} & 0.179 & 0.198 & 0.217 & \textbf{0.133} & 0.258 & 0.217 & 0.171 & 0.224 & 0.223 \\
us\_births/M/short & 0.814 & 0.894 & 0.887 & \underline{0.754} & 0.909 & 0.766 & 1.156 & \textbf{0.680} & 0.939 & 0.886 \\
us\_births/W/short & \underline{0.553} & 0.654 & 0.616 & 0.750 & \textbf{0.550} & 0.743 & 0.668 & 0.739 & 0.922 & 0.897 \\
\bottomrule
\end{tabular}
}
\end{table}

\begin{table}
\centering
\caption{Normalized MASE scores of different zero-shot models on the GiftEval benchmark. Scores are relative to a seasonal naive baseline, where values below 1.0 indicate better performance. The models achieving the best and second-best scores are highlighted.}
\label{tab:normalized_mase}
\footnotesize
\resizebox{\textwidth}{!}{
\begin{tabular}{l | r r r r r r r r r r}
\toprule
\multirow{2}{*}{Dataset} & \rotatebox{45}{TempoPFN} & \rotatebox{45}{TiRex} & \rotatebox{45}{FlowState-9.1M} & \rotatebox{45}{Toto\_Open\_Base\_1.0} & \rotatebox{45}{TTM-R2-Finetuned} & \rotatebox{45}{TabPFN-TS} & \rotatebox{45}{Chronos Bolt B} & \rotatebox{45}{YingLong\_50m} & \rotatebox{45}{Moirai L 1.1} & \rotatebox{45}{Moirai B 1.1} \\
\midrule
bitbrains\_fast\_storage/5T/long & 0.846 & \underline{0.808} & 0.913 & \textbf{0.789} & 0.827 & 1.014 & 0.834 & 0.886 & 0.840 & 0.851 \\
bitbrains\_fast\_storage/5T/medium & 0.896 & \underline{0.815} & 1.033 & \textbf{0.807} & 0.876 & 1.072 & 0.871 & 0.889 & 0.836 & 0.860 \\
bitbrains\_fast\_storage/5T/short & 0.716 & \underline{0.609} & 0.883 & \textbf{0.591} & 0.648 & 0.879 & 0.662 & 0.719 & 0.728 & 0.697 \\
bitbrains\_fast\_storage/H/short & 0.910 & 0.825 & 0.851 & \textbf{0.728} & 0.927 & 0.912 & \underline{0.824} & 0.856 & 0.839 & 0.909 \\
bitbrains\_rnd/5T/long & 1.005 & \underline{0.954} & 1.007 & \textbf{0.953} & 1.001 & 1.107 & 0.970 & 1.002 & 0.977 & 0.985 \\
bitbrains\_rnd/5T/medium & 1.001 & \textbf{0.967} & 1.009 & \underline{0.972} & 0.997 & 1.064 & 0.979 & 1.004 & 0.982 & 0.997 \\
bitbrains\_rnd/5T/short & 0.940 & \underline{0.845} & 0.996 & \textbf{0.837} & 0.889 & 1.030 & 0.865 & 0.915 & 0.888 & 0.923 \\
bitbrains\_rnd/H/short & 0.999 & \underline{0.971} & 0.982 & \textbf{0.934} & 1.004 & 1.106 & 0.977 & 0.972 & 0.982 & 1.005 \\
bizitobs\_application/10S/long & 1.086 & 1.150 & 1.042 & \underline{1.021} & 1.187 & \textbf{0.965} & 3.270 & 1.518 & 2.445 & 4.210 \\
bizitobs\_application/10S/medium & \textbf{0.800} & 1.035 & 0.962 & \underline{0.856} & 1.134 & 0.925 & 3.612 & 1.516 & 2.746 & 4.756 \\
bizitobs\_application/10S/short & \textbf{0.466} & 0.571 & 0.597 & \underline{0.556} & 0.734 & 0.563 & 2.468 & 0.841 & 2.011 & 2.373 \\
bizitobs\_l2c/5T/long & 0.840 & 0.819 & \underline{0.356} & 0.809 & \textbf{0.347} & 0.457 & 0.853 & 0.816 & 0.770 & 0.770 \\
bizitobs\_l2c/5T/medium & 0.629 & 0.668 & \textbf{0.418} & 0.606 & \underline{0.441} & 0.513 & 0.706 & 0.677 & 0.794 & 0.686 \\
bizitobs\_l2c/5T/short & 0.278 & 0.299 & 0.317 & \underline{0.262} & \textbf{0.250} & 0.311 & 0.282 & 0.292 & 0.289 & 0.295 \\
bizitobs\_l2c/H/long & 0.478 & 0.425 & \textbf{0.384} & 0.559 & 0.844 & 0.466 & \underline{0.390} & 0.554 & 0.891 & 0.764 \\
bizitobs\_l2c/H/medium & 0.403 & 0.358 & \textbf{0.312} & 0.501 & 0.753 & \underline{0.324} & 0.328 & 0.512 & 0.828 & 0.874 \\
bizitobs\_l2c/H/short & 0.381 & 0.420 & \underline{0.371} & 0.387 & 0.448 & 0.400 & \textbf{0.356} & 0.532 & 0.947 & 0.823 \\
bizitobs\_service/10S/long & 1.216 & 1.104 & 1.011 & \textbf{0.953} & 1.086 & \underline{0.999} & 3.875 & 1.715 & 3.167 & 4.447 \\
bizitobs\_service/10S/medium & 1.076 & 0.946 & \underline{0.866} & \textbf{0.820} & 0.953 & 0.928 & 3.768 & 1.605 & 2.931 & 4.536 \\
bizitobs\_service/10S/short & 0.947 & 0.677 & 0.658 & \textbf{0.644} & \underline{0.653} & 0.721 & 2.706 & 0.953 & 1.885 & 2.799 \\
car\_parts/M/short & 0.700 & 0.698 & 0.744 & \textbf{0.675} & 0.698 & 0.706 & 0.712 & 1.057 & 0.752 & \underline{0.695} \\
covid\_deaths/D/short & 0.785 & 0.830 & 0.738 & \underline{0.695} & \textbf{0.657} & 0.837 & 0.828 & 0.929 & 0.778 & 0.738 \\
electricity/15T/long & 1.170 & \underline{0.759} & 0.778 & 0.897 & \textbf{0.747} & 0.812 & 0.801 & 0.863 & 1.126 & 1.134 \\
electricity/15T/medium & 1.027 & \underline{0.726} & 0.744 & 0.858 & \textbf{0.714} & 0.773 & 0.749 & 0.825 & 1.121 & 1.156 \\
electricity/15T/short & 0.751 & 0.557 & 0.630 & 0.667 & \textbf{0.530} & 0.670 & \underline{0.545} & 0.628 & 0.996 & 0.897 \\
electricity/D/short & 0.787 & \textbf{0.716} & \underline{0.722} & 0.747 & 0.730 & 0.751 & 0.729 & 0.744 & 0.760 & 0.755 \\
electricity/H/long & 0.896 & 0.802 & \underline{0.795} & 0.811 & \textbf{0.781} & 0.861 & 0.812 & 0.869 & 0.892 & 0.827 \\
electricity/H/medium & 0.871 & \underline{0.780} & 0.785 & 0.792 & 0.782 & 0.838 & \textbf{0.774} & 0.831 & 0.862 & 0.855 \\
electricity/H/short & 0.794 & \textbf{0.641} & 0.689 & 0.719 & 0.663 & 0.763 & \underline{0.643} & 0.807 & 0.795 & 0.803 \\
electricity/W/short & 0.751 & \underline{0.691} & \textbf{0.653} & 0.857 & 0.729 & 0.740 & 0.707 & 0.832 & 0.857 & 0.919 \\
ett1/15T/long & 1.050 & 0.871 & \textbf{0.850} & 0.898 & 0.870 & 0.939 & 0.954 & \underline{0.869} & 1.176 & 0.940 \\
ett1/15T/medium & 0.995 & 0.868 & \textbf{0.833} & 0.909 & 0.902 & 0.919 & 0.893 & \underline{0.865} & 1.094 & 1.044 \\
ett1/15T/short & 0.830 & 0.745 & \underline{0.737} & 0.742 & 0.751 & 0.794 & \textbf{0.728} & 0.769 & 0.990 & 0.883 \\
ett1/D/short & \underline{0.918} & 0.952 & \textbf{0.916} & 0.933 & 1.056 & 0.922 & 0.940 & 0.953 & 0.984 & 0.978 \\
ett1/H/long & 0.937 & \textbf{0.886} & \underline{0.906} & 0.926 & 0.941 & 0.996 & 0.916 & 0.909 & 0.981 & 0.933 \\
ett1/H/medium & 0.860 & \underline{0.785} & \textbf{0.782} & 0.811 & 0.819 & 0.896 & 0.877 & 0.808 & 0.855 & 0.861 \\
ett1/H/short & 0.875 & 0.849 & \textbf{0.834} & 0.885 & 0.869 & 0.908 & \underline{0.847} & 0.850 & 0.875 & 0.906 \\
ett1/W/short & \underline{0.851} & 1.003 & \textbf{0.820} & 0.893 & 0.902 & 0.938 & 0.959 & 0.980 & 0.854 & 0.871 \\
ett2/15T/long & 1.014 & 0.907 & \textbf{0.855} & \underline{0.860} & 0.918 & 0.965 & 0.928 & 0.908 & 1.126 & 1.284 \\
ett2/15T/medium & 0.923 & \underline{0.859} & \textbf{0.806} & 0.878 & 0.872 & 0.933 & 0.877 & 0.865 & 1.008 & 1.046 \\
ett2/15T/short & 0.731 & \underline{0.695} & 0.707 & 0.736 & \textbf{0.681} & 0.788 & 0.718 & 0.724 & 0.937 & 0.899 \\
ett2/D/short & 1.648 & \underline{0.917} & 1.053 & 1.161 & \textbf{0.868} & 1.029 & 0.951 & 0.943 & 1.036 & 0.942 \\
ett2/H/long & \textbf{0.879} & 1.008 & 0.972 & 0.955 & 0.919 & 1.280 & \underline{0.918} & 0.940 & 1.134 & 0.993 \\
ett2/H/medium & 0.843 & 0.846 & 0.838 & \underline{0.821} & 0.831 & 1.008 & 0.830 & \textbf{0.816} & 0.953 & 0.832 \\
ett2/H/short & 0.809 & 0.809 & \textbf{0.782} & 0.796 & 0.811 & 0.894 & \underline{0.794} & 0.815 & 0.848 & 0.874 \\
ett2/W/short & 1.150 & 1.021 & 1.256 & 1.267 & 1.009 & \underline{0.983} & \textbf{0.949} & 1.221 & 1.683 & 1.093 \\
hierarchical\_sales/D/short & 0.662 & \underline{0.653} & 0.660 & \textbf{0.648} & 0.666 & 0.669 & 0.655 & 0.677 & 0.657 & 0.657 \\
hierarchical\_sales/W/short & 0.711 & \underline{0.704} & \textbf{0.695} & 0.726 & 0.709 & 0.713 & 0.715 & 0.757 & 0.731 & 0.729 \\
hospital/M/short & 0.834 & 0.829 & \underline{0.824} & 0.851 & \textbf{0.815} & 0.830 & 0.860 & 0.899 & 0.834 & 0.842 \\
jena\_weather/10T/long & 0.851 & \underline{0.836} & 0.868 & \textbf{0.833} & 1.008 & 0.876 & 0.862 & 0.843 & 1.040 & 1.001 \\
jena\_weather/10T/medium & 0.867 & \underline{0.842} & 0.873 & \textbf{0.835} & 1.001 & 0.874 & 0.852 & 0.888 & 0.969 & 0.994 \\
jena\_weather/10T/short & 0.401 & 0.402 & \underline{0.373} & \textbf{0.358} & 0.431 & 0.418 & 0.411 & 0.452 & 0.455 & 0.471 \\
jena\_weather/D/short & 0.811 & \textbf{0.648} & \underline{0.666} & 0.760 & 0.840 & 0.781 & 0.668 & 0.687 & 0.725 & 0.731 \\
jena\_weather/H/long & 0.907 & \underline{0.778} & 0.856 & 0.800 & 1.153 & 1.109 & 0.811 & 0.832 & \textbf{0.695} & 0.836 \\
jena\_weather/H/medium & 0.934 & 0.949 & 0.965 & \underline{0.847} & 0.969 & 1.225 & \textbf{0.841} & 0.991 & 1.003 & 0.919 \\
jena\_weather/H/short & 0.739 & \textbf{0.716} & \underline{0.728} & 0.752 & 0.802 & 0.759 & 0.741 & 0.740 & 0.809 & 0.766 \\
kdd\_cup\_2018/D/short & 0.803 & 0.820 & 0.817 & 0.810 & 0.800 & \underline{0.784} & 0.799 & \textbf{0.782} & 0.802 & 0.802 \\
kdd\_cup\_2018/H/long & 0.776 & \underline{0.554} & 0.789 & 0.779 & 0.752 & 0.819 & \textbf{0.512} & 0.756 & 0.649 & 0.719 \\
kdd\_cup\_2018/H/medium & 0.753 & \underline{0.561} & 0.759 & 0.754 & 0.721 & 0.790 & \textbf{0.490} & 0.721 & 0.668 & 0.735 \\
kdd\_cup\_2018/H/short & 0.727 & \underline{0.490} & 0.712 & 0.739 & 0.709 & 0.784 & \textbf{0.448} & 0.697 & 0.667 & 0.704 \\
loop\_seattle/5T/long & 0.952 & 0.783 & 0.660 & 0.678 & 0.697 & 0.805 & 0.990 & 0.882 & \textbf{0.445} & \underline{0.473} \\
loop\_seattle/5T/medium & 1.004 & 0.796 & 0.671 & 0.697 & 0.701 & 0.836 & 0.985 & 0.949 & \textbf{0.390} & \underline{0.453} \\
loop\_seattle/5T/short & 0.856 & 0.745 & 0.756 & 0.737 & 0.732 & 0.784 & 0.823 & 0.864 & \textbf{0.638} & \underline{0.703} \\
loop\_seattle/D/short & \underline{0.518} & \textbf{0.505} & \textbf{0.505} & 0.534 & 0.519 & 0.524 & 0.521 & 0.541 & 0.529 & 0.521 \\
loop\_seattle/H/long & 0.727 & \textbf{0.583} & 0.595 & 0.610 & \underline{0.585} & 0.599 & 0.644 & 0.636 & 0.679 & 0.744 \\
loop\_seattle/H/medium & 0.721 & 0.634 & 0.662 & \underline{0.628} & \textbf{0.623} & 0.661 & 0.688 & 0.684 & 0.675 & 0.770 \\
loop\_seattle/H/short & 0.775 & \textbf{0.657} & \underline{0.664} & 0.696 & 0.678 & 0.705 & 0.696 & 0.693 & 0.731 & 0.820 \\
m4\_daily/D/short & 1.352 & \textbf{0.942} & 1.096 & 1.010 & 1.007 & 1.279 & \underline{0.976} & 1.131 & 1.275 & 1.638 \\
m4\_hourly/H/short & 0.695 & \textbf{0.589} & \underline{0.613} & 0.721 & 0.862 & 0.619 & 0.701 & 0.808 & 0.743 & 0.814 \\
m4\_monthly/M/short & \underline{0.733} & \textbf{0.732} & \textbf{0.732} & 0.780 & 0.751 & 0.760 & 0.753 & 0.836 & 0.776 & 0.757 \\
m4\_quarterly/Q/short & 0.733 & 0.725 & \underline{0.720} & 0.766 & 0.730 & 0.764 & 0.764 & 0.875 & \textbf{0.712} & \textbf{0.712} \\
m4\_weekly/W/short & 0.903 & \textbf{0.679} & 0.728 & 0.864 & \underline{0.700} & 0.738 & 0.748 & 0.816 & 0.929 & 1.012 \\
m4\_yearly/A/short & 0.863 & 0.864 & \textbf{0.746} & 0.857 & 0.819 & 0.834 & 0.884 & 1.098 & \underline{0.749} & 0.759 \\
m\_dense/D/short & \underline{0.413} & 0.414 & 0.439 & 0.457 & 0.438 & \textbf{0.406} & 0.429 & 0.498 & 0.573 & 0.659 \\
m\_dense/H/long & 0.751 & \underline{0.491} & 0.504 & 0.528 & 0.497 & 0.692 & 0.635 & 0.666 & \textbf{0.471} & 0.497 \\
m\_dense/H/medium & 0.660 & 0.464 & \underline{0.441} & 0.464 & 0.452 & 0.633 & 0.561 & 0.567 & \textbf{0.436} & 0.468 \\
m\_dense/H/short & 0.683 & 0.528 & 0.534 & 0.591 & 0.544 & 0.609 & \textbf{0.521} & 0.630 & \underline{0.522} & 0.563 \\
restaurant/D/short & 0.684 & \textbf{0.674} & \underline{0.678} & 0.779 & 0.695 & 0.694 & 0.696 & 0.713 & 0.711 & 0.700 \\
saugeen/D/short & 0.917 & 0.933 & \textbf{0.802} & 0.869 & 0.878 & 0.922 & \underline{0.832} & 0.888 & 0.964 & 0.853 \\
saugeen/M/short & 0.809 & 0.790 & 0.766 & 0.775 & 0.773 & \textbf{0.720} & \underline{0.757} & 0.763 & 0.774 & 0.854 \\
saugeen/W/short & 0.678 & \underline{0.588} & \textbf{0.586} & 0.657 & 0.690 & 0.662 & 0.611 & 0.623 & 0.693 & 0.708 \\
solar/10T/long & 1.481 & \textbf{0.928} & \underline{0.977} & 1.011 & 1.144 & 1.000 & 1.229 & 1.564 & 2.239 & 2.319 \\
solar/10T/medium & 1.196 & 0.969 & 0.970 & \underline{0.951} & 1.075 & \textbf{0.906} & 1.108 & 1.324 & 1.963 & 2.039 \\
solar/10T/short & \underline{0.823} & 0.982 & 0.928 & 0.934 & \textbf{0.745} & 0.854 & 0.896 & 1.015 & 1.004 & 0.995 \\
solar/D/short & \underline{0.846} & \textbf{0.841} & 0.858 & 0.875 & 0.851 & 0.854 & 0.849 & 0.856 & 0.854 & 0.882 \\
solar/H/long & 0.929 & \textbf{0.688} & \underline{0.879} & 0.894 & 1.311 & 1.011 & 0.964 & 0.911 & 0.952 & 0.999 \\
solar/H/medium & 1.061 & \textbf{0.792} & 0.937 & 0.926 & 1.323 & \underline{0.924} & 0.996 & 0.993 & 0.981 & 0.954 \\
solar/H/short & 0.962 & \textbf{0.759} & 0.916 & 0.870 & 0.926 & 0.943 & \underline{0.854} & 0.922 & 0.919 & 0.938 \\
solar/W/short & 0.804 & 0.830 & \underline{0.570} & 0.970 & 0.678 & \textbf{0.539} & 0.666 & 1.415 & 1.041 & 1.129 \\
sz\_taxi/15T/long & 0.787 & \textbf{0.736} & 0.766 & \underline{0.749} & 0.851 & 0.810 & 0.789 & 0.754 & 0.802 & 0.777 \\
sz\_taxi/15T/medium & 0.789 & \textbf{0.753} & 0.772 & \underline{0.764} & \underline{0.764} & 0.793 & 0.784 & 0.774 & 0.797 & 0.782 \\
sz\_taxi/15T/short & 0.733 & \textbf{0.712} & 0.721 & 0.719 & 0.733 & 0.730 & \underline{0.717} & 0.721 & 0.760 & 0.754 \\
sz\_taxi/H/short & 0.781 & \underline{0.763} & 0.780 & 0.769 & 0.780 & 0.776 & \textbf{0.762} & 0.768 & 0.814 & 0.797 \\
temperature\_rain/D/short & 0.689 & 0.666 & 0.671 & 0.679 & 0.702 & 0.685 & \underline{0.648} & 0.725 & \textbf{0.597} & 0.651 \\
us\_births/D/short & 0.211 & 0.219 & 0.245 & 0.267 & \underline{0.208} & \textbf{0.170} & 0.260 & 0.324 & 0.270 & 0.273 \\
us\_births/M/short & 0.875 & 0.941 & 0.920 & \underline{0.764} & \textbf{0.673} & 0.941 & 1.215 & 0.799 & 1.014 & 0.951 \\
us\_births/W/short & \underline{0.583} & 0.687 & 0.651 & 0.790 & 0.729 & \textbf{0.572} & 0.696 & 0.789 & 0.940 & 0.921 \\
\bottomrule
\end{tabular}
}
\end{table}

\begin{table}
\centering
\caption{CRPS performance summarized by average rank for zero-shot models on the GiftEval benchmark. A lower rank signifies superior probabilistic forecasting performance. The models achieving the first and second-best overall average ranks are highlighted.}
\label{tab:crps_rank}
\footnotesize
\resizebox{\textwidth}{!}{
\begin{tabular}{l | r r r r r r r r r r}
\toprule
\multirow{2}{*}{Dataset} & \rotatebox{45}{TempoPFN} & \rotatebox{45}{TiRex} & \rotatebox{45}{FlowState-9.1M} & \rotatebox{45}{Toto\_Open\_Base\_1.0} & \rotatebox{45}{Chronos Bolt B} & \rotatebox{45}{TabPFN-TS} & \rotatebox{45}{YingLong\_50m} & \rotatebox{45}{TTM-R2-Finetuned} & \rotatebox{45}{Moirai L 1.1} & \rotatebox{45}{Moirai B 1.1} \\
\midrule
bitbrains\_fast\_storage/5T/long & \textbf{1.000} & \underline{2.000} & 8.000 & 3.000 & 7.000 & 9.000 & 4.000 & 10.000 & 5.000 & 6.000 \\
bitbrains\_fast\_storage/5T/medium & \textbf{1.000} & \underline{2.000} & 8.000 & 3.000 & 7.000 & 10.000 & 5.000 & 9.000 & 4.000 & 6.000 \\
bitbrains\_fast\_storage/5T/short & \underline{2.000} & 3.000 & 8.000 & \textbf{1.000} & 7.000 & 10.000 & 6.000 & 9.000 & 4.000 & 5.000 \\
bitbrains\_fast\_storage/H/short & 4.000 & 7.000 & 8.000 & \underline{2.000} & 9.000 & 6.000 & 3.000 & 11.000 & 5.000 & \textbf{1.000} \\
bitbrains\_rnd/5T/long & 6.000 & \underline{2.000} & 7.000 & \textbf{1.000} & 8.000 & 10.000 & 3.000 & 9.000 & 5.000 & 4.000 \\
bitbrains\_rnd/5T/medium & 5.000 & \textbf{1.000} & 8.000 & 6.000 & 3.000 & 10.000 & 7.000 & 9.000 & \underline{2.000} & 4.000 \\
bitbrains\_rnd/5T/short & 9.000 & \underline{2.000} & 7.000 & \textbf{1.000} & 5.000 & 10.000 & 4.000 & 8.000 & 3.000 & 6.000 \\
bitbrains\_rnd/H/short & 6.000 & 5.000 & 7.000 & 3.000 & 4.000 & 9.000 & 8.000 & 10.000 & \textbf{1.000} & \underline{2.000} \\
bizitobs\_application/10S/long & \textbf{2.000} & 4.000 & 6.000 & 5.000 & 10.000 & \underline{3.000} & 8.000 & 7.000 & 9.000 & 11.000 \\
bizitobs\_application/10S/medium & \textbf{1.000} & 3.000 & 5.000 & \underline{2.000} & 10.000 & 4.000 & 8.000 & 7.000 & 9.000 & 11.000 \\
bizitobs\_application/10S/short & \textbf{1.000} & 3.000 & 4.000 & \underline{2.000} & 11.000 & 5.000 & 6.000 & 7.000 & 10.000 & 8.000 \\
bizitobs\_l2c/5T/long & 8.000 & 9.000 & \textbf{1.000} & 5.000 & 11.000 & 3.000 & 6.000 & \underline{2.000} & 4.000 & 7.000 \\
bizitobs\_l2c/5T/medium & 5.000 & 6.000 & \textbf{1.000} & 4.000 & 10.000 & 3.000 & 7.000 & \underline{2.000} & 9.000 & 8.000 \\
bizitobs\_l2c/5T/short & 3.000 & 7.000 & 10.000 & \underline{2.000} & 4.000 & 9.000 & 6.000 & \textbf{1.000} & 8.000 & 5.000 \\
bizitobs\_l2c/H/long & 5.000 & 3.000 & \textbf{1.000} & 6.000 & \underline{2.000} & 4.000 & 7.000 & 10.000 & 9.000 & 8.000 \\
bizitobs\_l2c/H/medium & 5.000 & 4.000 & \textbf{1.000} & 6.000 & 3.000 & \underline{2.000} & 7.000 & 9.000 & 8.000 & 10.000 \\
bizitobs\_l2c/H/short & 3.000 & 6.000 & \textbf{1.000} & 4.000 & \underline{2.000} & 5.000 & 8.000 & 7.000 & 11.000 & 9.000 \\
bizitobs\_service/10S/long & 6.000 & 4.000 & 5.000 & \textbf{1.000} & 10.000 & \underline{2.000} & 8.000 & 7.000 & 9.000 & 11.000 \\
bizitobs\_service/10S/medium & \textbf{1.000} & 4.000 & \underline{2.000} & 3.000 & 11.000 & 5.000 & 7.000 & 6.000 & 9.000 & 10.000 \\
bizitobs\_service/10S/short & \underline{2.000} & 4.000 & 3.000 & \textbf{1.000} & 11.000 & 7.000 & 6.000 & 5.000 & 8.000 & 10.000 \\
car\_parts/M/short & 7.000 & 3.000 & 6.000 & \textbf{1.000} & 4.000 & \underline{2.000} & 10.000 & 8.000 & 9.000 & 5.000 \\
covid\_deaths/D/short & \underline{2.000} & 3.000 & 6.000 & \textbf{1.000} & 9.000 & 5.000 & 10.000 & 4.000 & 8.000 & 7.000 \\
electricity/15T/long & 10.000 & \textbf{1.000} & \underline{2.000} & 7.000 & 6.000 & 4.000 & 5.000 & 3.000 & 8.000 & 11.000 \\
electricity/15T/medium & 8.000 & \textbf{1.000} & \underline{2.000} & 7.000 & 5.000 & 4.000 & 6.000 & 3.000 & 9.000 & 10.000 \\
electricity/15T/short & 8.000 & 3.000 & 4.000 & 7.000 & \textbf{1.000} & 6.000 & 5.000 & \underline{2.000} & 10.000 & 9.000 \\
electricity/D/short & 9.000 & \textbf{1.000} & \underline{2.000} & 6.000 & 3.000 & 8.000 & 5.000 & 4.000 & 10.000 & 7.000 \\
electricity/H/long & 7.000 & 5.000 & 3.000 & \textbf{1.000} & 6.000 & 10.000 & 9.000 & 4.000 & 8.000 & \underline{2.000} \\
electricity/H/medium & 9.000 & 4.000 & \underline{2.000} & \textbf{1.000} & 5.000 & 10.000 & 7.000 & 3.000 & 8.000 & 6.000 \\
electricity/H/short & 8.000 & \textbf{1.000} & 3.000 & 5.000 & \underline{2.000} & 6.000 & 10.000 & 4.000 & 9.000 & 7.000 \\
electricity/W/short & 6.000 & \underline{2.000} & \textbf{1.000} & 9.000 & 3.000 & 4.000 & 8.000 & 5.000 & 7.000 & 10.000 \\
ett1/15T/long & 8.000 & 3.000 & \textbf{1.000} & 4.000 & 9.000 & 6.000 & \underline{2.000} & 5.000 & 11.000 & 7.000 \\
ett1/15T/medium & 7.000 & 3.000 & \textbf{1.000} & 5.000 & 8.000 & 4.000 & \underline{2.000} & 6.000 & 11.000 & 10.000 \\
ett1/15T/short & 8.000 & \underline{2.000} & 4.000 & 3.000 & \textbf{1.000} & 5.000 & 6.000 & 7.000 & 10.000 & 9.000 \\
ett1/D/short & \textbf{1.000} & 3.000 & 4.000 & 5.000 & 7.000 & 8.000 & \underline{2.000} & 10.000 & 6.000 & 9.000 \\
ett1/H/long & 6.000 & \underline{2.000} & 4.000 & 3.000 & 10.000 & 8.000 & \textbf{1.000} & 5.000 & 9.000 & 7.000 \\
ett1/H/medium & 7.000 & 3.000 & \textbf{1.000} & 4.000 & 10.000 & 9.000 & \underline{2.000} & 6.000 & 5.000 & 8.000 \\
ett1/H/short & 5.000 & 3.000 & \textbf{1.000} & 7.000 & 4.000 & 8.000 & \underline{2.000} & 9.000 & 6.000 & 10.000 \\
ett1/W/short & \underline{2.000} & 11.000 & \textbf{1.000} & 5.000 & 9.000 & 7.000 & 8.000 & 6.000 & 3.000 & 4.000 \\
ett2/15T/long & 7.000 & 5.000 & \underline{2.000} & \textbf{1.000} & 8.000 & 6.000 & 3.000 & 4.000 & 9.000 & 11.000 \\
ett2/15T/medium & 6.000 & 3.000 & \textbf{1.000} & 4.000 & 10.000 & 7.000 & \underline{2.000} & 5.000 & 8.000 & 9.000 \\
ett2/15T/short & 7.000 & 3.000 & \underline{2.000} & 6.000 & 4.000 & 8.000 & 5.000 & \textbf{1.000} & 10.000 & 9.000 \\
ett2/D/short & 9.000 & \underline{2.000} & 7.000 & 8.000 & 3.000 & 10.000 & 5.000 & \textbf{1.000} & 4.000 & 6.000 \\
ett2/H/long & \textbf{1.000} & 7.000 & 6.000 & 4.000 & 8.000 & 10.000 & 3.000 & \underline{2.000} & 9.000 & 5.000 \\
ett2/H/medium & 6.000 & 7.000 & 4.000 & \underline{2.000} & 8.000 & 10.000 & 3.000 & 5.000 & 9.000 & \textbf{1.000} \\
ett2/H/short & 5.000 & 7.000 & \textbf{1.000} & 3.000 & \underline{2.000} & 10.000 & 4.000 & 6.000 & 8.000 & 9.000 \\
ett2/W/short & 4.000 & \textbf{1.000} & 7.000 & 8.000 & 3.000 & 6.000 & 10.000 & 5.000 & 9.000 & \underline{2.000} \\
hierarchical\_sales/D/short & 7.000 & \textbf{1.000} & 6.000 & \underline{2.000} & 4.000 & 9.000 & 8.000 & 10.000 & 5.000 & 3.000 \\
hierarchical\_sales/W/short & \underline{2.000} & 4.000 & \textbf{1.000} & 6.000 & 5.000 & 3.000 & 10.000 & 9.000 & 8.000 & 7.000 \\
hospital/M/short & 6.000 & 4.000 & \textbf{1.000} & 5.000 & 9.000 & 8.000 & 10.000 & 7.000 & 3.000 & \underline{2.000} \\
jena\_weather/10T/long & 5.000 & 3.000 & \underline{2.000} & \textbf{1.000} & 7.000 & 4.000 & 6.000 & 8.000 & 10.000 & 9.000 \\
jena\_weather/10T/medium & 4.000 & 3.000 & \underline{2.000} & \textbf{1.000} & 6.000 & 5.000 & 7.000 & 9.000 & 10.000 & 8.000 \\
jena\_weather/10T/short & 4.000 & 3.000 & \underline{2.000} & \textbf{1.000} & 5.000 & 6.000 & 7.000 & 8.000 & 9.000 & 10.000 \\
jena\_weather/D/short & 7.000 & \underline{2.000} & 4.000 & 8.000 & \textbf{1.000} & 3.000 & 5.000 & 10.000 & 9.000 & 6.000 \\
jena\_weather/H/long & 5.000 & \textbf{1.000} & 8.000 & \underline{2.000} & 6.000 & 9.000 & 4.000 & 10.000 & 3.000 & 7.000 \\
jena\_weather/H/medium & 5.000 & \textbf{1.000} & 3.000 & \underline{2.000} & 4.000 & 9.000 & 6.000 & 10.000 & 8.000 & 7.000 \\
jena\_weather/H/short & 3.000 & \textbf{1.000} & \underline{2.000} & 6.000 & 5.000 & 4.000 & 7.000 & 10.000 & 9.000 & 8.000 \\
kdd\_cup\_2018/D/short & 5.000 & 9.000 & 7.000 & 8.000 & 3.000 & \textbf{1.000} & \underline{2.000} & 10.000 & 6.000 & 4.000 \\
kdd\_cup\_2018/H/long & 6.000 & \underline{2.000} & 8.000 & 7.000 & \textbf{1.000} & 10.000 & 5.000 & 9.000 & 3.000 & 4.000 \\
kdd\_cup\_2018/H/medium & 5.000 & \underline{2.000} & 6.000 & 8.000 & \textbf{1.000} & 9.000 & 4.000 & 10.000 & 3.000 & 7.000 \\
kdd\_cup\_2018/H/short & 6.000 & \underline{2.000} & 5.000 & 8.000 & \textbf{1.000} & 10.000 & 4.000 & 9.000 & 3.000 & 7.000 \\
loop\_seattle/5T/long & 9.000 & 6.000 & 3.000 & 4.000 & 11.000 & 7.000 & 8.000 & 5.000 & \textbf{1.000} & \underline{2.000} \\
loop\_seattle/5T/medium & 9.000 & 6.000 & 3.000 & 4.000 & 10.000 & 7.000 & 8.000 & 5.000 & \textbf{1.000} & \underline{2.000} \\
loop\_seattle/5T/short & 9.000 & 4.000 & 5.000 & 3.000 & 8.000 & 7.000 & 10.000 & 6.000 & \textbf{1.000} & \underline{2.000} \\
loop\_seattle/D/short & 3.000 & \underline{2.000} & \textbf{1.000} & 8.000 & 5.000 & 4.000 & 6.000 & 10.000 & 9.000 & 7.000 \\
loop\_seattle/H/long & 8.000 & \textbf{1.000} & 3.000 & 4.000 & 9.000 & \underline{2.000} & 6.000 & 5.000 & 7.000 & 10.000 \\
loop\_seattle/H/medium & 8.000 & \underline{2.000} & 4.000 & \textbf{1.000} & 9.000 & 3.000 & 6.000 & 5.000 & 7.000 & 10.000 \\
loop\_seattle/H/short & 9.000 & \textbf{1.000} & \underline{2.000} & 5.000 & 6.000 & 3.000 & 4.000 & 8.000 & 7.000 & 10.000 \\
m4\_daily/D/short & 9.000 & \textbf{1.000} & 7.000 & 3.000 & \underline{2.000} & 4.000 & 6.000 & 5.000 & 10.000 & 11.000 \\
m4\_hourly/H/short & 7.000 & \underline{2.000} & 3.000 & 10.000 & 6.000 & 8.000 & 5.000 & 9.000 & \textbf{1.000} & 4.000 \\
m4\_monthly/M/short & \textbf{1.000} & \underline{2.000} & 3.000 & 8.000 & 6.000 & 4.000 & 10.000 & 9.000 & 7.000 & 5.000 \\
m4\_quarterly/Q/short & 4.000 & \underline{3.000} & 5.000 & 7.000 & 6.000 & 8.000 & 10.000 & 9.000 & \textbf{1.500} & \textbf{1.500} \\
m4\_weekly/W/short & 6.000 & \textbf{1.000} & \underline{2.000} & 10.000 & 4.000 & 3.000 & 5.000 & 7.000 & 8.000 & 9.000 \\
m4\_yearly/A/short & 7.000 & 4.000 & 3.000 & 9.000 & 8.000 & 5.000 & 11.000 & 6.000 & \textbf{1.000} & \underline{2.000} \\
m\_dense/D/short & \underline{2.000} & 3.000 & 5.000 & 7.000 & 4.000 & \textbf{1.000} & 8.000 & 6.000 & 9.000 & 10.000 \\
m\_dense/H/long & 10.000 & 3.000 & \underline{2.000} & 6.000 & 8.000 & 7.000 & 9.000 & 5.000 & \textbf{1.000} & 4.000 \\
m\_dense/H/medium & 10.000 & 3.000 & \underline{2.000} & 4.000 & 8.000 & 9.000 & 7.000 & 5.000 & \textbf{1.000} & 6.000 \\
m\_dense/H/short & 10.000 & 3.000 & 4.000 & 7.000 & \textbf{1.000} & 8.000 & 9.000 & 6.000 & \underline{2.000} & 5.000 \\
restaurant/D/short & 3.000 & \textbf{1.000} & \underline{2.000} & 10.000 & 5.000 & 4.000 & 9.000 & 7.000 & 8.000 & 6.000 \\
saugeen/D/short & 7.000 & 8.000 & \textbf{1.000} & 3.000 & \underline{2.000} & 6.000 & 5.000 & 10.000 & 9.000 & 4.000 \\
saugeen/M/short & 5.000 & 7.000 & \underline{2.000} & 4.000 & 3.000 & \textbf{1.000} & 6.000 & 9.000 & 8.000 & 10.000 \\
saugeen/W/short & 7.000 & \textbf{1.000} & \underline{2.000} & 5.000 & 3.000 & 6.000 & 4.000 & 10.000 & 9.000 & 8.000 \\
solar/10T/long & 7.000 & \textbf{1.000} & 3.000 & 4.000 & 5.000 & \underline{2.000} & 8.000 & 6.000 & 10.000 & 11.000 \\
solar/10T/medium & 5.000 & 4.000 & 3.000 & \underline{2.000} & 6.000 & \textbf{1.000} & 8.000 & 7.000 & 10.000 & 11.000 \\
solar/10T/short & \underline{2.000} & 7.000 & 4.000 & 5.000 & 3.000 & \textbf{1.000} & 8.000 & 6.000 & 9.000 & 10.000 \\
solar/D/short & 3.000 & 4.000 & \underline{2.000} & 7.000 & 5.000 & \textbf{1.000} & 6.000 & 10.000 & 8.000 & 9.000 \\
solar/H/long & 8.000 & \textbf{1.000} & 3.000 & \underline{2.000} & 9.000 & 5.000 & 6.000 & 10.000 & 4.000 & 7.000 \\
solar/H/medium & 9.000 & \textbf{1.000} & 5.000 & 4.000 & 8.000 & \underline{2.000} & 7.000 & 10.000 & 6.000 & 3.000 \\
solar/H/short & 8.000 & \textbf{1.000} & 5.000 & 3.000 & \underline{2.000} & 6.000 & 9.000 & 10.000 & 4.000 & 7.000 \\
solar/W/short & 4.000 & 5.000 & \underline{2.000} & 7.000 & 3.000 & \textbf{1.000} & 11.000 & 6.000 & 9.000 & 10.000 \\
sz\_taxi/15T/long & 6.000 & \textbf{1.000} & 4.000 & \underline{2.000} & 10.000 & 9.000 & 3.000 & 8.000 & 7.000 & 5.000 \\
sz\_taxi/15T/medium & 6.000 & \textbf{1.000} & 3.000 & \underline{2.000} & 10.000 & 9.000 & 4.000 & 7.000 & 8.000 & 5.000 \\
sz\_taxi/15T/short & 6.000 & \textbf{1.000} & 3.000 & 4.000 & \underline{2.000} & 7.000 & 5.000 & 8.000 & 10.000 & 9.000 \\
sz\_taxi/H/short & 6.000 & \textbf{1.000} & 5.000 & 4.000 & \underline{2.000} & 7.000 & 3.000 & 8.000 & 10.000 & 9.000 \\
temperature\_rain/D/short & 8.000 & 5.000 & 4.000 & 6.000 & 3.000 & 7.000 & 9.000 & 10.000 & \textbf{1.000} & \underline{2.000} \\
us\_births/D/short & \underline{2.000} & 4.000 & 5.000 & 7.000 & 6.000 & \textbf{1.000} & 10.000 & 3.000 & 9.000 & 8.000 \\
us\_births/M/short & 4.000 & 7.000 & 6.000 & \underline{2.000} & 11.000 & 8.000 & 3.000 & \textbf{1.000} & 9.000 & 5.000 \\
us\_births/W/short & \underline{2.000} & 4.000 & 3.000 & 8.000 & 5.000 & \textbf{1.000} & 7.000 & 6.000 & 10.000 & 9.000 \\
\bottomrule
\end{tabular}
}
\end{table}

\begin{table}
\centering
\caption{MASE performance presented as average ranks for zero-shot models across the GiftEval benchmark. A lower average rank indicates consistently higher accuracy across datasets. The two models with the best overall average ranks are highlighted.}
\label{tab:mase_rank}
\footnotesize
\resizebox{\linewidth}{!}{
\begin{tabular}{l | r r r r r r r r r r}
\toprule
\multirow{2}{*}{Dataset} & \rotatebox{45}{ TempoPFN} & \rotatebox{45}{TiRex} & \rotatebox{45}{FlowState-9.1M} & \rotatebox{45}{Toto\_Open\_Base\_1.0} & \rotatebox{45}{Chronos Bolt B} & \rotatebox{45}{TTM-R2-Finetuned} & \rotatebox{45}{TabPFN-TS} & \rotatebox{45}{YingLong\_50m} & \rotatebox{45}{Moirai L 1.1} & \rotatebox{45}{Moirai B 1.1} \\
\midrule
bitbrains\_fast\_storage/5T/long & 6.000 & \underline{2.000} & 9.000 & \textbf{1.000} & 4.000 & 3.000 & 11.000 & 8.000 & 5.000 & 7.000 \\
bitbrains\_fast\_storage/5T/medium & 8.000 & \underline{2.000} & 10.000 & \textbf{1.000} & 5.000 & 6.000 & 11.000 & 7.000 & 3.000 & 4.000 \\
bitbrains\_fast\_storage/5T/short & 6.000 & \underline{2.000} & 10.000 & \textbf{1.000} & 4.000 & 3.000 & 9.000 & 7.000 & 8.000 & 5.000 \\
bitbrains\_fast\_storage/H/short & 8.000 & 3.000 & 5.000 & \textbf{1.000} & \underline{2.000} & 10.000 & 9.000 & 6.000 & 4.000 & 7.000 \\
bitbrains\_rnd/5T/long & 9.000 & \underline{2.000} & 10.000 & \textbf{1.000} & 3.000 & 7.000 & 11.000 & 8.000 & 4.000 & 5.000 \\
bitbrains\_rnd/5T/medium & 8.000 & \textbf{1.000} & 10.000 & \underline{2.000} & 3.000 & 5.000 & 11.000 & 9.000 & 4.000 & 6.000 \\
bitbrains\_rnd/5T/short & 8.000 & \underline{2.000} & 9.000 & \textbf{1.000} & 3.000 & 5.000 & 11.000 & 6.000 & 4.000 & 7.000 \\
bitbrains\_rnd/H/short & 7.000 & \underline{2.000} & 5.000 & \textbf{1.000} & 4.000 & 9.000 & 11.000 & 3.000 & 6.000 & 10.000 \\
bizitobs\_application/10S/long & 5.000 & 6.000 & 4.000 & \underline{3.000} & 10.000 & 7.000 & \textbf{1.000} & 8.000 & 9.000 & 11.000 \\
bizitobs\_application/10S/medium & \textbf{1.000} & 6.000 & 4.000 & \underline{2.000} & 10.000 & 7.000 & 3.000 & 8.000 & 9.000 & 11.000 \\
bizitobs\_application/10S/short & \textbf{1.000} & 4.000 & 5.000 & \underline{2.000} & 11.000 & 6.000 & 3.000 & 7.000 & 9.000 & 10.000 \\
bizitobs\_l2c/5T/long & 9.000 & 8.000 & \underline{2.000} & 6.000 & 10.000 & \textbf{1.000} & 3.000 & 7.000 & 4.500 & 4.500 \\
bizitobs\_l2c/5T/medium & 5.000 & 6.000 & \textbf{1.000} & 4.000 & 9.000 & \underline{2.000} & 3.000 & 7.000 & 10.000 & 8.000 \\
bizitobs\_l2c/5T/short & 3.000 & 8.000 & 10.000 & \underline{2.000} & 4.000 & \textbf{1.000} & 9.000 & 6.000 & 5.000 & 7.000 \\
bizitobs\_l2c/H/long & 5.000 & 3.000 & \textbf{1.000} & 7.000 & \underline{2.000} & 9.000 & 4.000 & 6.000 & 10.000 & 8.000 \\
bizitobs\_l2c/H/medium & 5.000 & 4.000 & \textbf{1.000} & 6.000 & 3.000 & 8.000 & \underline{2.000} & 7.000 & 9.000 & 10.000 \\
bizitobs\_l2c/H/short & 3.000 & 6.000 & \underline{2.000} & 4.000 & \textbf{1.000} & 7.000 & 5.000 & 8.000 & 10.000 & 9.000 \\
bizitobs\_service/10S/long & 7.000 & 6.000 & 4.000 & \textbf{1.000} & 10.000 & 5.000 & \underline{2.000} & 8.000 & 9.000 & 11.000 \\
bizitobs\_service/10S/medium & 7.000 & 4.000 & \underline{2.000} & \textbf{1.000} & 10.000 & 5.000 & 3.000 & 8.000 & 9.000 & 11.000 \\
bizitobs\_service/10S/short & 6.000 & 4.000 & 3.000 & \textbf{1.000} & 10.000 & \underline{2.000} & 5.000 & 7.000 & 9.000 & 11.000 \\
car\_parts/M/short & 5.000 & 4.000 & 8.000 & \textbf{1.000} & 7.000 & 3.000 & 6.000 & 11.000 & 9.000 & \underline{2.000} \\
covid\_deaths/D/short & 6.000 & 8.000 & 4.000 & \underline{2.000} & 7.000 & \textbf{1.000} & 9.000 & 10.000 & 5.000 & 3.000 \\
electricity/15T/long & 11.000 & \underline{2.000} & 3.000 & 7.000 & 4.000 & \textbf{1.000} & 5.000 & 6.000 & 9.000 & 10.000 \\
electricity/15T/medium & 9.000 & \underline{2.000} & 3.000 & 7.000 & 4.000 & \textbf{1.000} & 5.000 & 6.000 & 10.000 & 11.000 \\
electricity/15T/short & 8.000 & 3.000 & 5.000 & 6.000 & \underline{2.000} & \textbf{1.000} & 7.000 & 4.000 & 10.000 & 9.000 \\
electricity/D/short & 10.000 & \textbf{1.000} & \underline{2.000} & 6.000 & 3.000 & 4.000 & 7.000 & 5.000 & 9.000 & 8.000 \\
electricity/H/long & 10.000 & 3.000 & \underline{2.000} & 4.000 & 5.000 & \textbf{1.000} & 7.000 & 8.000 & 9.000 & 6.000 \\
electricity/H/medium & 10.000 & \underline{2.000} & 4.000 & 5.000 & \textbf{1.000} & 3.000 & 7.000 & 6.000 & 9.000 & 8.000 \\
electricity/H/short & 7.000 & \textbf{1.000} & 4.000 & 5.000 & \underline{2.000} & 3.000 & 6.000 & 10.000 & 8.000 & 9.000 \\
electricity/W/short & 6.000 & \underline{2.000} & \textbf{1.000} & 8.000 & 3.000 & 4.000 & 5.000 & 7.000 & 9.000 & 10.000 \\
ett1/15T/long & 10.000 & 4.000 & \textbf{1.000} & 5.000 & 8.000 & 3.000 & 6.000 & \underline{2.000} & 11.000 & 7.000 \\
ett1/15T/medium & 8.000 & 3.000 & \textbf{1.000} & 6.000 & 4.000 & 5.000 & 7.000 & \underline{2.000} & 11.000 & 10.000 \\
ett1/15T/short & 8.000 & 4.000 & \underline{2.000} & 3.000 & \textbf{1.000} & 5.000 & 7.000 & 6.000 & 10.000 & 9.000 \\
ett1/D/short & \underline{2.000} & 6.000 & \textbf{1.000} & 4.000 & 5.000 & 11.000 & 3.000 & 7.000 & 9.000 & 8.000 \\
ett1/H/long & 7.000 & \textbf{1.000} & \underline{2.000} & 5.000 & 4.000 & 8.000 & 10.000 & 3.000 & 9.000 & 6.000 \\
ett1/H/medium & 7.000 & \underline{2.000} & \textbf{1.000} & 4.000 & 9.000 & 5.000 & 10.000 & 3.000 & 6.000 & 8.000 \\
ett1/H/short & 6.000 & 3.000 & \textbf{1.000} & 8.000 & \underline{2.000} & 5.000 & 10.000 & 4.000 & 7.000 & 9.000 \\
ett1/W/short & \underline{2.000} & 11.000 & \textbf{1.000} & 5.000 & 8.000 & 6.000 & 7.000 & 9.000 & 3.000 & 4.000 \\
ett2/15T/long & 9.000 & 3.000 & \textbf{1.000} & \underline{2.000} & 6.000 & 5.000 & 7.000 & 4.000 & 10.000 & 11.000 \\
ett2/15T/medium & 7.000 & \underline{2.000} & \textbf{1.000} & 6.000 & 5.000 & 4.000 & 8.000 & 3.000 & 10.000 & 11.000 \\
ett2/15T/short & 6.000 & \underline{2.000} & 3.000 & 7.000 & 4.000 & \textbf{1.000} & 8.000 & 5.000 & 10.000 & 9.000 \\
ett2/D/short & 11.000 & \underline{2.000} & 9.000 & 10.000 & 5.000 & \textbf{1.000} & 7.000 & 4.000 & 8.000 & 3.000 \\
ett2/H/long & \textbf{1.000} & 9.000 & 6.000 & 5.000 & \underline{2.000} & 3.000 & 11.000 & 4.000 & 10.000 & 7.000 \\
ett2/H/medium & 7.000 & 8.000 & 6.000 & \underline{2.000} & 3.000 & 4.000 & 11.000 & \textbf{1.000} & 9.000 & 5.000 \\
ett2/H/short & 5.000 & 4.000 & \textbf{1.000} & 3.000 & \underline{2.000} & 6.000 & 10.000 & 7.000 & 8.000 & 9.000 \\
ett2/W/short & 7.000 & 5.000 & 9.000 & 10.000 & \textbf{1.000} & 4.000 & \underline{2.000} & 8.000 & 11.000 & 6.000 \\
hierarchical\_sales/D/short & 7.000 & \underline{2.000} & 6.000 & \textbf{1.000} & 3.000 & 8.000 & 9.000 & 10.000 & 4.000 & 5.000 \\
hierarchical\_sales/W/short & 4.000 & \underline{2.000} & \textbf{1.000} & 7.000 & 6.000 & 3.000 & 5.000 & 10.000 & 9.000 & 8.000 \\
hospital/M/short & 6.000 & 3.000 & \underline{2.000} & 8.000 & 9.000 & \textbf{1.000} & 4.000 & 10.000 & 5.000 & 7.000 \\
jena\_weather/10T/long & 4.000 & \underline{2.000} & 6.000 & \textbf{1.000} & 5.000 & 10.000 & 7.000 & 3.000 & 11.000 & 9.000 \\
jena\_weather/10T/medium & 4.000 & \underline{2.000} & 5.000 & \textbf{1.000} & 3.000 & 11.000 & 6.000 & 7.000 & 8.000 & 9.000 \\
jena\_weather/10T/short & 3.000 & 4.000 & \underline{2.000} & \textbf{1.000} & 5.000 & 7.000 & 6.000 & 8.000 & 9.000 & 10.000 \\
jena\_weather/D/short & 9.000 & \textbf{1.000} & \underline{2.000} & 7.000 & 3.000 & 10.000 & 8.000 & 4.000 & 5.000 & 6.000 \\
jena\_weather/H/long & 8.000 & \underline{2.000} & 7.000 & 3.000 & 4.000 & 11.000 & 10.000 & 5.000 & \textbf{1.000} & 6.000 \\
jena\_weather/H/medium & 4.000 & 5.000 & 6.000 & \underline{2.000} & \textbf{1.000} & 7.000 & 11.000 & 8.000 & 10.000 & 3.000 \\
jena\_weather/H/short & 3.000 & \textbf{1.000} & \underline{2.000} & 6.000 & 5.000 & 9.000 & 7.000 & 4.000 & 10.000 & 8.000 \\
kdd\_cup\_2018/D/short & 7.000 & 10.000 & 9.000 & 8.000 & 3.000 & 4.000 & \underline{2.000} & \textbf{1.000} & 5.500 & 5.500 \\
kdd\_cup\_2018/H/long & 7.000 & \underline{2.000} & 9.000 & 8.000 & \textbf{1.000} & 5.000 & 10.000 & 6.000 & 3.000 & 4.000 \\
kdd\_cup\_2018/H/medium & 7.000 & \underline{2.000} & 9.000 & 8.000 & \textbf{1.000} & 4.000 & 10.000 & 5.000 & 3.000 & 6.000 \\
kdd\_cup\_2018/H/short & 8.000 & \underline{2.000} & 7.000 & 9.000 & \textbf{1.000} & 6.000 & 10.000 & 4.000 & 3.000 & 5.000 \\
loop\_seattle/5T/long & 9.000 & 6.000 & 3.000 & 4.000 & 10.000 & 5.000 & 7.000 & 8.000 & \textbf{1.000} & \underline{2.000} \\
loop\_seattle/5T/medium & 11.000 & 6.000 & 3.000 & 4.000 & 9.000 & 5.000 & 7.000 & 8.000 & \textbf{1.000} & \underline{2.000} \\
loop\_seattle/5T/short & 9.000 & 5.000 & 6.000 & 4.000 & 8.000 & 3.000 & 7.000 & 10.000 & \textbf{1.000} & \underline{2.000} \\
loop\_seattle/D/short & 3.000 & \textbf{1.000} & \underline{2.000} & 9.000 & 5.000 & 4.000 & 7.000 & 10.000 & 8.000 & 6.000 \\
loop\_seattle/H/long & 9.000 & \textbf{1.000} & 3.000 & 5.000 & 7.000 & \underline{2.000} & 4.000 & 6.000 & 8.000 & 10.000 \\
loop\_seattle/H/medium & 9.000 & 3.000 & 5.000 & \underline{2.000} & 8.000 & \textbf{1.000} & 4.000 & 7.000 & 6.000 & 10.000 \\
loop\_seattle/H/short & 9.000 & \textbf{1.000} & \underline{2.000} & 5.000 & 6.000 & 3.000 & 7.000 & 4.000 & 8.000 & 10.000 \\
m4\_daily/D/short & 10.000 & \textbf{1.000} & 6.000 & 5.000 & \underline{2.000} & 4.000 & 9.000 & 7.000 & 8.000 & 11.000 \\
m4\_hourly/H/short & 4.000 & \textbf{1.000} & \underline{2.000} & 6.000 & 5.000 & 10.000 & 3.000 & 8.000 & 7.000 & 9.000 \\
m4\_monthly/M/short & 3.000 & \underline{2.000} & \textbf{1.000} & 9.000 & 5.000 & 4.000 & 7.000 & 10.000 & 8.000 & 6.000 \\
m4\_quarterly/Q/short & 6.000 & 4.000 & \underline{3.000} & 9.000 & 7.000 & 5.000 & 8.000 & 10.000 & \textbf{1.500} & \textbf{1.500} \\
m4\_weekly/W/short & 8.000 & \textbf{1.000} & 3.000 & 7.000 & 5.000 & \underline{2.000} & 4.000 & 6.000 & 9.000 & 11.000 \\
m4\_yearly/A/short & 7.000 & 8.000 & \textbf{1.000} & 6.000 & 9.000 & 4.000 & 5.000 & 11.000 & \underline{2.000} & 3.000 \\
m\_dense/D/short & \underline{2.000} & 3.000 & 6.000 & 7.000 & 4.000 & 5.000 & \textbf{1.000} & 8.000 & 9.000 & 10.000 \\
m\_dense/H/long & 10.000 & \underline{2.000} & 5.000 & 6.000 & 7.000 & 4.000 & 9.000 & 8.000 & \textbf{1.000} & 3.000 \\
m\_dense/H/medium & 10.000 & 4.000 & \underline{2.000} & 5.000 & 7.000 & 3.000 & 9.000 & 8.000 & \textbf{1.000} & 6.000 \\
m\_dense/H/short & 10.000 & 3.000 & 4.000 & 7.000 & \textbf{1.000} & 5.000 & 8.000 & 9.000 & \underline{2.000} & 6.000 \\
restaurant/D/short & 3.000 & \textbf{1.000} & \underline{2.000} & 10.000 & 6.000 & 5.000 & 4.000 & 9.000 & 8.000 & 7.000 \\
saugeen/D/short & 7.000 & 9.000 & \textbf{1.000} & 4.000 & \underline{2.000} & 5.000 & 8.000 & 6.000 & 10.000 & 3.000 \\
saugeen/M/short & 9.000 & 8.000 & 4.000 & 7.000 & \underline{2.000} & 5.000 & \textbf{1.000} & 3.000 & 6.000 & 10.000 \\
saugeen/W/short & 7.000 & \underline{2.000} & \textbf{1.000} & 5.000 & 3.000 & 8.000 & 6.000 & 4.000 & 9.000 & 10.000 \\
solar/10T/long & 8.000 & \textbf{1.000} & \underline{2.000} & 5.000 & 7.000 & 6.000 & 3.000 & 9.000 & 10.000 & 11.000 \\
solar/10T/medium & 8.000 & 3.000 & 4.000 & \underline{2.000} & 7.000 & 6.000 & \textbf{1.000} & 9.000 & 10.000 & 11.000 \\
solar/10T/short & \underline{2.000} & 7.000 & 5.000 & 6.000 & 4.000 & \textbf{1.000} & 3.000 & 11.000 & 10.000 & 8.000 \\
solar/D/short & \underline{2.000} & \textbf{1.000} & 8.000 & 9.000 & 3.000 & 4.000 & 6.000 & 7.000 & 5.000 & 10.000 \\
solar/H/long & 5.000 & \textbf{1.000} & \underline{2.000} & 3.000 & 7.000 & 11.000 & 10.000 & 4.000 & 6.000 & 8.000 \\
solar/H/medium & 10.000 & \textbf{1.000} & 4.000 & 3.000 & 8.000 & 11.000 & \underline{2.000} & 7.000 & 6.000 & 5.000 \\
solar/H/short & 10.000 & \textbf{1.000} & 4.000 & 3.000 & \underline{2.000} & 7.000 & 9.000 & 6.000 & 5.000 & 8.000 \\
solar/W/short & 5.000 & 6.000 & \underline{2.000} & 7.000 & 3.000 & 4.000 & \textbf{1.000} & 11.000 & 9.000 & 10.000 \\
sz\_taxi/15T/long & 6.000 & \textbf{1.000} & 4.000 & \underline{2.000} & 7.000 & 10.000 & 9.000 & 3.000 & 8.000 & 5.000 \\
sz\_taxi/15T/medium & 8.000 & \textbf{1.000} & 4.000 & 3.000 & 7.000 & \underline{2.000} & 9.000 & 5.000 & 10.000 & 6.000 \\
sz\_taxi/15T/short & 7.000 & \textbf{1.000} & 5.000 & 3.000 & \underline{2.000} & 8.000 & 6.000 & 4.000 & 10.000 & 9.000 \\
sz\_taxi/H/short & 8.000 & \underline{2.000} & 6.000 & 4.000 & \textbf{1.000} & 7.000 & 5.000 & 3.000 & 10.000 & 9.000 \\
temperature\_rain/D/short & 8.000 & 4.000 & 5.000 & 6.000 & \underline{2.000} & 9.000 & 7.000 & 10.000 & \textbf{1.000} & 3.000 \\
us\_births/D/short & 3.000 & 4.000 & 5.000 & 7.000 & 6.000 & \underline{2.000} & \textbf{1.000} & 10.000 & 8.000 & 9.000 \\
us\_births/M/short & 4.000 & 7.000 & 5.000 & \underline{2.000} & 11.000 & \textbf{1.000} & 6.000 & 3.000 & 10.000 & 8.000 \\
us\_births/W/short & \underline{2.000} & 4.000 & 3.000 & 8.000 & 5.000 & 6.000 & \textbf{1.000} & 7.000 & 10.000 & 9.000 \\
\bottomrule
\end{tabular}
}
\end{table}

\newpage
\section*{NeurIPS Paper Checklist}

\begin{enumerate}

\item {\bf Claims}
    \item[] Question: Do the main claims made in the abstract and introduction accurately reflect the paper's contributions and scope?
    \item[] Answer: \answerYes{} %
    \item[] Justification: The abstract and introduction state the scope (univariate, zero-shot, synthetic-only pre-training) and claims that are supported by results on Gift-Eval, \texttt{fev-bench} and Chronos-ZS (\Cref{subsec:gifteval_results,subsec:fev_results}), the backbone swap (\Cref{subsec:backbone_vs_prior}), and the ablations (\Cref{subsec:ablations}).

\item {\bf Limitations}
    \item[] Question: Does the paper discuss the limitations of the work performed by the authors?
    \item[] Answer: \answerYes{} %
    \item[] Justification: \Cref{sec:conclusion} discusses the univariate scope and extensions to covariates and real data; per-dataset failure modes are analyzed in \Cref{subsec:gifteval_results}, wall-clock runtime vs.\ TiRex in \Cref{subsec:fev_results}, and single-seed ablations in \Cref{subsec:ablations}.

\item {\bf Theory assumptions and proofs}
    \item[] Question: For each theoretical result, does the paper provide the full set of assumptions and a complete (and correct) proof?
    \item[] Answer: \answerNA{} %
    \item[] Justification: The paper does not contain theoretical results.

    \item {\bf Experimental result reproducibility}
    \item[] Question: Does the paper fully disclose all the information needed to reproduce the main experimental results of the paper to the extent that it affects the main claims and/or conclusions of the paper (regardless of whether the code and data are provided or not)?
    \item[] Answer: \answerYes{} %
    \item[] Justification: The full architecture, synthetic generators, augmentations, and training hyperparameters are described in \Cref{sec:architecture}, \Cref{app:synthetic_details}, and \Cref{app:training-details}, and we release code to generate the data, train the model, and reproduce the evaluations.

\item {\bf Open access to data and code}
    \item[] Question: Does the paper provide open access to the data and code, with sufficient instructions to faithfully reproduce the main experimental results, as described in supplemental material?
    \item[] Answer: \answerYes{} %
    \item[] Justification: The code for synthetic data generation, training, and evaluation is available at \url{https://github.com/automl/TempoPFN}. We do not train on real time series; the evaluation benchmarks are publicly available.

\item {\bf Experimental setting/details}
    \item[] Question: Does the paper specify all the training and test details (e.g., data splits, hyperparameters, how they were chosen, type of optimizer) necessary to understand the results?
    \item[] Answer: \answerYes{} %
    \item[] Justification: Training details and hyperparameters are given in \Cref{app:training-details} and \Cref{tab:training-hyperparams}; we follow the standard evaluation protocols of Gift-Eval, \texttt{fev-bench} and Chronos-ZS.

\item {\bf Experiment statistical significance}
    \item[] Question: Does the paper report error bars suitably and correctly defined or other appropriate information about the statistical significance of the experiments?
    \item[] Answer: \answerNo{} %
    \item[] Justification: Due to the high cost of pre-training, we did not repeat pre-training runs with multiple seeds, and the ablations report single runs; we state this explicitly when interpreting small differences (\Cref{subsec:ablations}). For \texttt{fev-bench}, pairwise comparisons include 95\% bootstrap confidence intervals (\Cref{fig:pairwise_heatmaps_sql}).

\item {\bf Experiments compute resources}
    \item[] Question: For each experiment, does the paper provide sufficient information on the computer resources (type of compute workers, memory, time of execution) needed to reproduce the experiments?
    \item[] Answer: \answerYes{} %
    \item[] Justification: Pre-training uses 8--16 NVIDIA A100 or H100 GPUs with \texttt{bfloat16} (\Cref{app:training-details}); synthetic data generation throughput and hardware are reported in \Cref{app:throughput}. The project also required compute for preliminary and ablation runs.
    
\item {\bf Code of ethics}
    \item[] Question: Does the research conducted in the paper conform, in every respect, with the NeurIPS Code of Ethics \url{https://neurips.cc/public/EthicsGuidelines}?
    \item[] Answer: \answerYes{} %
    \item[] Justification: The research conforms to the NeurIPS Code of Ethics; it uses only synthetic training data and public benchmarks.

\item {\bf Broader impacts}
    \item[] Question: Does the paper discuss both potential positive societal impacts and negative societal impacts of the work performed?
    \item[] Answer: \answerNA{} %
    \item[] Justification: This is foundational research on forecasting models trained only on synthetic data; we do not foresee societal impacts beyond those of general-purpose forecasting tools.
    
\item {\bf Safeguards}
    \item[] Question: Does the paper describe safeguards that have been put in place for responsible release of data or models that have a high risk for misuse (e.g., pre-trained language models, image generators, or scraped datasets)?
    \item[] Answer: \answerNA{} %
    \item[] Justification: The model and data generators pose no particular risk of misuse.

\item {\bf Licenses for existing assets}
    \item[] Question: Are the creators or original owners of assets (e.g., code, data, models), used in the paper, properly credited and are the license and terms of use explicitly mentioned and properly respected?
    \item[] Answer: \answerYes{} %
    \item[] Justification: Yes, we cite all works which proposed synthetic data, augmentations and models we use in our model and comparisons.

\item {\bf New assets}
    \item[] Question: Are new assets introduced in the paper well documented and is the documentation provided alongside the assets?
    \item[] Answer: \answerYes{} %
    \item[] Justification: We release the synthetic data generators and training code with documentation at \url{https://github.com/automl/TempoPFN}.

\item {\bf Crowdsourcing and research with human subjects}
    \item[] Question: For crowdsourcing experiments and research with human subjects, does the paper include the full text of instructions given to participants and screenshots, if applicable, as well as details about compensation (if any)? 
    \item[] Answer: \answerNA{} %
    \item[] Justification: The paper does not involve crowdsourcing or human subjects.

\item {\bf Institutional review board (IRB) approvals or equivalent for research with human subjects}
    \item[] Question: Does the paper describe potential risks incurred by study participants, whether such risks were disclosed to the subjects, and whether Institutional Review Board (IRB) approvals (or an equivalent approval/review based on the requirements of your country or institution) were obtained?
    \item[] Answer: \answerNA{} %
    \item[] Justification: The paper does not involve human subjects.

\item {\bf Declaration of LLM usage}
    \item[] Question: Does the paper describe the usage of LLMs if it is an important, original, or non-standard component of the core methods in this research? Note that if the LLM is used only for writing, editing, or formatting purposes and does \emph{not} impact the core methodology, scientific rigor, or originality of the research, declaration is not required.
    \item[] Answer: \answerNA{} %
    \item[] Justification: LLMs were used only for editing; they are not part of the core methodology.

\end{enumerate}

\end{document}